\documentclass[10pt,twocolumn,letterpaper]{article}

\usepackage{arxivstyle}   %
\usepackage{pifont}
\usepackage{array}
\usepackage{ragged2e}
\usepackage{float}
\usepackage{inconsolata}
\usepackage{listings}
\usepackage{supertabular}
\usepackage{capt-of}
\usepackage{xurl}

\newcommand{\catname}[1]{\texttt{#1}}
\newcommand{\defn}{\textbf{Definition.}}
\newcommand{\incl}{\textbf{Include}}
\newcommand{\excl}{\textbf{Exclude}}
\newcommand{\bnd}{\textbf{Boundary}}
\newcommand{\pex}{\textbf{Positive Example}}
\newcommand{\cex}{\textbf{Counterexample (Reject)}}
\newcommand{\realex}{\textbf{Real Sample}}

\definecolor{ideaBrick}{HTML}{AB240E}
\definecolor{ideaOlive}{HTML}{6F7D36}

\newcommand{\ourbench}{%
  \textcolor{ideaBrick}{Idea}\textsc{\textcolor{ideaOlive}{Ambig}}\xspace%
}

\definecolor{rqtitle}{HTML}{F8E5AC}
\definecolor{rqborder}{HTML}{F4AB5A}
\definecolor{rqbackground}{HTML}{FFFFFF}
\definecolor{rqtitletext}{HTML}{AB240E}

\newtcolorbox{rqbox}[1]{
    enhanced,
    frame hidden,
    sharp corners,
    colback=rqtitle!25!white,
    borderline west={2pt}{0pt}{rqtitletext},
    fontupper=\normalsize\bfseries\itshape,
    before upper={\textup{\textcolor{rqtitletext}{[#1]}}\enspace},
    boxrule=0pt,
    boxsep=0pt,
    left=8pt,
    right=6pt,
    top=5pt,
    bottom=5pt,
    before skip=7pt,
    after skip=5pt
}

\newcommand{\numModels}{13}
\newcommand{\numTasks}{3}

\newcommand{\numLevelOneCategories}{3}
\newcommand{\numLevelTwoCategories}{10}
\newcommand{\numInstances}{660}
\newcommand{\numRealInstances}{163}
\newcommand{\numSyntheticInstances}{497}
\newcommand{\numMultiDefectInstances}{50}
\newcommand{\numGitHubInstances}{121}
\newcommand{\numRealReproInstances}{42}
\newcommand{\numSyntheticReproInstances}{358}
\newcommand{\numIdeationInstances}{139}
\newcommand{\numSyntheticEvalInstances}{200}

\newcommand{\casecell}[3]{%
  \cellcolor{#1!14}%
  \textbf{\textcolor{#1}{#2}}\par
  \vspace{0.10em}
  \textbf{\hspace{0pt}#3}%
}

\definecolor{headerbg}{HTML}{FCF2D6}
\definecolor{inputbg}{HTML}{FEF7EF}
\definecolor{goldbg}{HTML}{EDF0DF}
\definecolor{modelbg}{HTML}{FCEAD6}
\definecolor{interpbg}{HTML}{FDF6E3}

\definecolor{caseA}{HTML}{AB240E}
\definecolor{caseB}{HTML}{B85A0F}
\definecolor{caseC}{HTML}{6F7D36}

\lstdefinestyle{jsonschema}{
    basicstyle=\ttfamily\scriptsize,
    breaklines=true,
    breakatwhitespace=false,
    columns=fullflexible,
    keepspaces=true,
    showstringspaces=false,
    frame=none,
    tabsize=2
}
\newcommand{\code}[1]{\nolinkurl{#1}}

\newcolumntype{Y}{>{\RaggedRight\arraybackslash}X}
\newcolumntype{L}[1]{>{\RaggedRight\arraybackslash}p{#1}}

\definecolor{GuidelineBlue}{HTML}{AB240E}
\definecolor{GuidelineLightBlue}{HTML}{FEF7EF}
\definecolor{ReadyGreen}{HTML}{6F7D36}
\definecolor{ReadyLightGreen}{HTML}{EDF0DF}
\definecolor{NotReadyRed}{HTML}{AB240E}
\definecolor{NotReadyLightRed}{HTML}{FDE9E3}
\definecolor{UnsureOrange}{HTML}{B85A0F}
\definecolor{UnsureLightOrange}{HTML}{FEF0E6}
\definecolor{NeutralGray}{HTML}{3F7288}
\definecolor{NeutralLightGray}{HTML}{EEF6F9}

\newcommand{\READY}{
  \textcolor{ReadyGreen}{\textbf{\texttt{READY}}}
}

\newcommand{\NOTREADY}{
  \textcolor{NotReadyRed}{\textbf{\texttt{NOT\_READY}}}
}

\newcommand{\UNSURE}{
  \textcolor{UnsureOrange}{\textbf{\texttt{UNSURE}}}
}

\newtcolorbox{guidelinebox}[2][]{
  enhanced,
  breakable,
  colback=GuidelineLightBlue,
  colframe=GuidelineBlue,
  coltitle=white,
  fonttitle=\bfseries,
  title=#2,
  boxrule=0.7pt,
  arc=2mm,
  left=2.5mm,
  right=2.5mm,
  top=2mm,
  bottom=2mm,
  before skip=7pt,
  after skip=7pt,
  #1
}

\newtcolorbox{readybox}[1][]{
  enhanced,
  breakable,
  colback=ReadyLightGreen,
  colframe=ReadyGreen,
  boxrule=0.7pt,
  arc=2mm,
  left=2.5mm,
  right=2.5mm,
  top=2mm,
  bottom=2mm,
  before skip=6pt,
  after skip=6pt,
  #1
}

\newtcolorbox{notreadybox}[1][]{
  enhanced,
  breakable,
  colback=NotReadyLightRed,
  colframe=NotReadyRed,
  boxrule=0.7pt,
  arc=2mm,
  left=2.5mm,
  right=2.5mm,
  top=2mm,
  bottom=2mm,
  before skip=6pt,
  after skip=6pt,
  #1
}

\newtcolorbox{unsurebox}[1][]{
  enhanced,
  breakable,
  colback=UnsureLightOrange,
  colframe=UnsureOrange,
  boxrule=0.7pt,
  arc=2mm,
  left=2.5mm,
  right=2.5mm,
  top=2mm,
  bottom=2mm,
  before skip=6pt,
  after skip=6pt,
  #1
}

\newtcolorbox{summarybox}[1][]{
  enhanced,
  breakable,
  colback=NeutralLightGray,
  colframe=NeutralGray,
  coltitle=white,
  fonttitle=\bfseries,
  title=Annotation Summary for a \texttt{NOT\_READY} Decision,
  boxrule=0.9pt,
  arc=2mm,
  left=3mm,
  right=3mm,
  top=2.5mm,
  bottom=2.5mm,
  before skip=9pt,
  after skip=7pt,
  #1
}

\newcommand{\papercode}{\href{https://github.com/Yiling-Ma/IdeaAMBIG}{github.com/Yiling-Ma/IdeaAMBIG}}
\newcommand{\paperwebsite}{\href{https://yiling-ma.github.io/IdeaAMBIG-website/}{yiling-ma.github.io/IdeaAMBIG-website/}}
\newcommand{\papercorrespondence}{\href{mailto:yilun.zhao@yale.edu}{\texttt{yilun.zhao@yale.edu}}}

\newcommand{\paperlogo}{%
  \raisebox{-0.22em}{\includegraphics[height=1.2em]{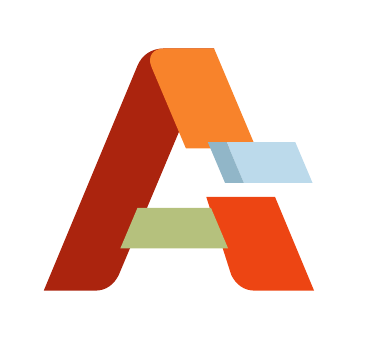}}\hspace{0.2em}%
}
\newcommand{\papertitle}{\mbox{\paperlogo\ourbench:} Benchmarking Implementation-Critical Gaps in Research-Idea Specifications}
\newcommand{\paperauthors}{Yiling Ma$^{1}$, Yilun Zhao$^{1}$, Sihong Wu$^{2}$, Manasi Patwardhan$^{3}$, Arman Cohan$^{1}$}
\newcommand{\paperaffiliations}{$^{1}$Yale University, \ $^{2}$University of Maryland, College Park, \ $^{3}$TCS Research}

\hypersetup{pdftitle={IdeaAMBIG: Benchmarking Implementation-Critical Gaps in Research-Idea Specifications},
            pdfauthor={Yiling Ma, Yilun Zhao, Sihong Wu, Manasi Patwardhan, Arman Cohan}}

\long\def\paperabstract{%
A research idea may be novel and scientifically plausible, yet its proposed method may remain insufficiently specified for faithful implementation.
We study codification readiness: whether a research-method
specification provides sufficient information for faithful
implementation without unsupported assumptions.
We introduce \ourbench{}, a benchmark of
660 evidence-grounded instances comprising 163 real-world gaps from
reproducibility reports and GitHub issues and 497 controlled synthetic
gaps injected into codification-ready references.
\ourbench{} evaluates three capabilities: codification-readiness assessment, defect localization, and clarification action generation, separating defect discovery from clarification given the defect.
Across \numModels{} LLMs, the best model achieves 9.6\% Macro Defect Recovery Rate on real-world instances.
When given the annotated defect, it achieves 80.6\% Macro Clarification Action Success Rate.
These results identify defect discovery as a key bottleneck for research automation, with models better able to formulate clarification actions once the defect is known.
}

\begin{document}
\raggedbottom %
\twocolumn[{%
  \arxivtitlecard{\papertitle}{\paperauthors}{\paperaffiliations}{\paperabstract}{%
    \metaline{\faGithub}{Code:}{\papercode}%
    \metaline{\faGlobe}{Website:}{\paperwebsite}%
    \metaline{\faEnvelope[regular]}{Correspondence:}{\papercorrespondence}}
}]

\section{Introduction}

\label{sec:intro}
\begin{figure*}[!t]
    \centering
    \includegraphics[width=\textwidth]{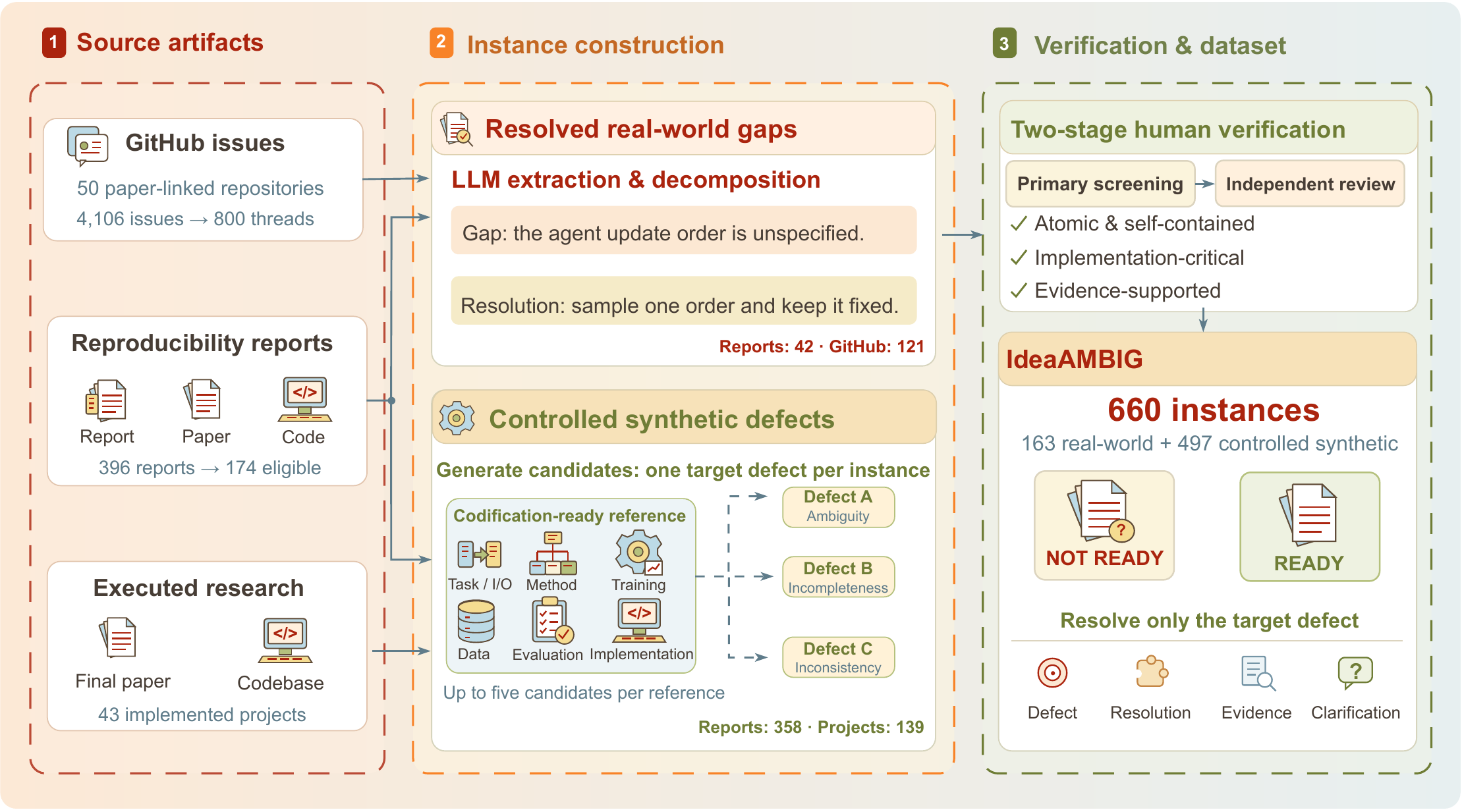}
\caption{
\textbf{Construction pipeline of \ourbench{}.}
\ding{182} Collect source artifacts from GitHub issues,
reproducibility reports, and executed research.
\ding{183} Reconstruct real-world gaps and inject single defects
into codification-ready references.
\ding{184} Two-stage human verification yields
\numInstances{} evidence-grounded instances.
}
    \label{fig:data_construction_pipeline}
\end{figure*}

LLMs increasingly support scientific workflows from research ideation to experimental execution and code generation~\citep{lu2024ai,weng2025deepscientist,schmidgall2025agent,si2025ideation}.
As these stages become coupled in research agents, reliable execution assumes that a generated idea specifies its intended method well enough for faithful implementation. 
When a method-defining choice is omitted, ambiguous, or
internally inconsistent, a downstream agent must either seek clarification or silently introduce an unsupported assumption, potentially producing working code that implements a different method. 
Reliable research automation therefore requires detecting and resolving specification gaps before codification.

\flushbottom
We ask whether the proposed methodological mechanism within a research idea is ready to be implemented as intended, rather than whether the idea is novel, scientifically valuable, or likely to succeed. We call this \emph{implementation readiness}. 
An implementation-facing specification of the idea is \emph{codification-ready} when a competent implementer can construct a faithful initial implementation or experimental prototype without unsupported assumptions about the core method. 
It is not ready if two competent implementers could make materially different method-defining choices with no evidence for which is intended. 
A plausible implementation merely hides this unresolved choice; a reliable model should instead locate it and seek the information needed to resolve it, avoiding implementation of the wrong method and resulting reproducibility failures~\citep{zhu2025ai,dobbins2025large}.

Existing evaluations largely target upstream idea quality or
downstream plans and artifacts~\citep{qiu2025ai,si2025can,
zhao2025abgen,baumgartner2026scicoqa}.
We complement these evaluations by assessing whether an
implementation-facing specification sufficiently determines
the intended method before codification, separating readiness
assessment, defect localization, and targeted clarification.

We introduce \ourbench{}, a benchmark of 660 evidence-grounded,
single-defect instances for evaluating codification readiness.
The \numRealInstances{} real-world instances come from GitHub issues\footnote{\url{https://docs.github.com/en/rest/search}}
and reproducibility reports \footnote{\url{https://jmlr.org/tmlr/papers/}}
in which implementers encountered genuine gaps with evidence-supported
resolutions~\citep{
joelle_pineau_2019_3158244,
koustuv_sinha_2021_4833117,
koustuv_sinha_2022_6574723,
koustuv_sinha_2023_8200058,
Jose2020-wb,
Hiemstra2021-hj,
Hagen2022-gj,
Kamps2023-ny,
Goharian2024-yw,
Hauff2025-ac}.
From these sources, we reconstruct self-contained,
implementation-facing specifications, using the documented
resolutions to establish the intended methodological choices.
The \numSyntheticInstances{} controlled synthetic instances alter exactly one implementation-critical detail in a codification-ready reference, providing a precise counterfactual target \footnote{A reference is considered codification-ready only if it is derived
from a paper with an independently verified successful reproduction or
execution, including working code and reported results; see
\S\ref{sec:data_collection}.}.
Each instance includes the target defect, its supported resolution, and labels from \numLevelOneCategories{} Level-1 and \numLevelTwoCategories{} Level-2 categories.

We evaluate \numTasks{} successive capabilities: readiness assessment, defect localization, and clarification action generation, separating blocker discovery from action once the blocker is known. 
Human studies, blind review, and independent reannotation show that both subsets largely contain resolvable, implementation-critical gaps with reliable labels. 
They further confirm that \ourbench{} measures implementation-oriented specification clarifications and that controlled instances preserve the validity and workflow plausibility of naturally occurring gaps (\S\ref{sec:benchmark-validation};
Appendix~\ref{app:construct_validity}; Appendix~\ref{app:synthetic_validation}).

Across \numModels{} LLMs, the strongest model reaches only 9.6\% Macro Defect Recovery Rate on real-world instances but 80.6\% Macro Clarification Action Success Rate when given the annotated defect. 
Removing defect-category prediction raises GPT-5.6-Sol's localization accuracy from 16.0\% to 40.5\% on the full real-world set, showing that taxonomy prediction adds difficulty while defect discovery remains challenging.
Models often request useful information once directed to the blocker but struggle to identify it independently. 
In an oracle study, supplying the missing information raises
the downstream codification-ready rate from 14\% to 98\%.

Our main contributions are summarized below:
\begin{itemize}[leftmargin=*,itemsep=2pt,topsep=2pt]
    \item We formalize the codification readiness of
implementation-facing idea specifications as a missing link
between scientific ideation and execution, and define \numTasks{} evaluation tasks
    (\S\ref{sec:task_formulation}).
    \item We introduce the \numInstances{}-instance \ourbench benchmark, combining
    real-world and controlled synthetic defects with supported resolutions
    (\S\ref{sec:benchmark}).
    \item We evaluate \numModels{} LLMs, identify defect localization as the principal
    bottleneck, and demonstrate the downstream value of clarification
    (\S\ref{sec:experiments}).
\end{itemize}

\section{Related Work}
\label{sec:related_work}

\paragraph{Scientific Ideation and Automated Research.}
Research-ideation benchmarks assess novelty, feasibility, diversity, and
alignment~\citep{qiu2025ai,ruan2024liveideabench,si2025can}, while research
agents integrate ideation with literature review, experimentation, coding, and
writing~\citep{lu2024ai,schmidgall2025agent,lu2026towards,weng2025deepscientist}.
The performance gap between ideas evaluated before and after execution shows that a promising idea need not yield a strong executed project~\citep{si2025ideation}.
Existing evaluations focus on proposed ideas or resulting artifacts, rather than the specification connecting these stages.
\ourbench{} asks whether this specification defines the core method sufficiently for faithful implementation.

\suppressfloats[t]
\begin{figure}[t]
    \centering
    \includegraphics[width=\columnwidth]{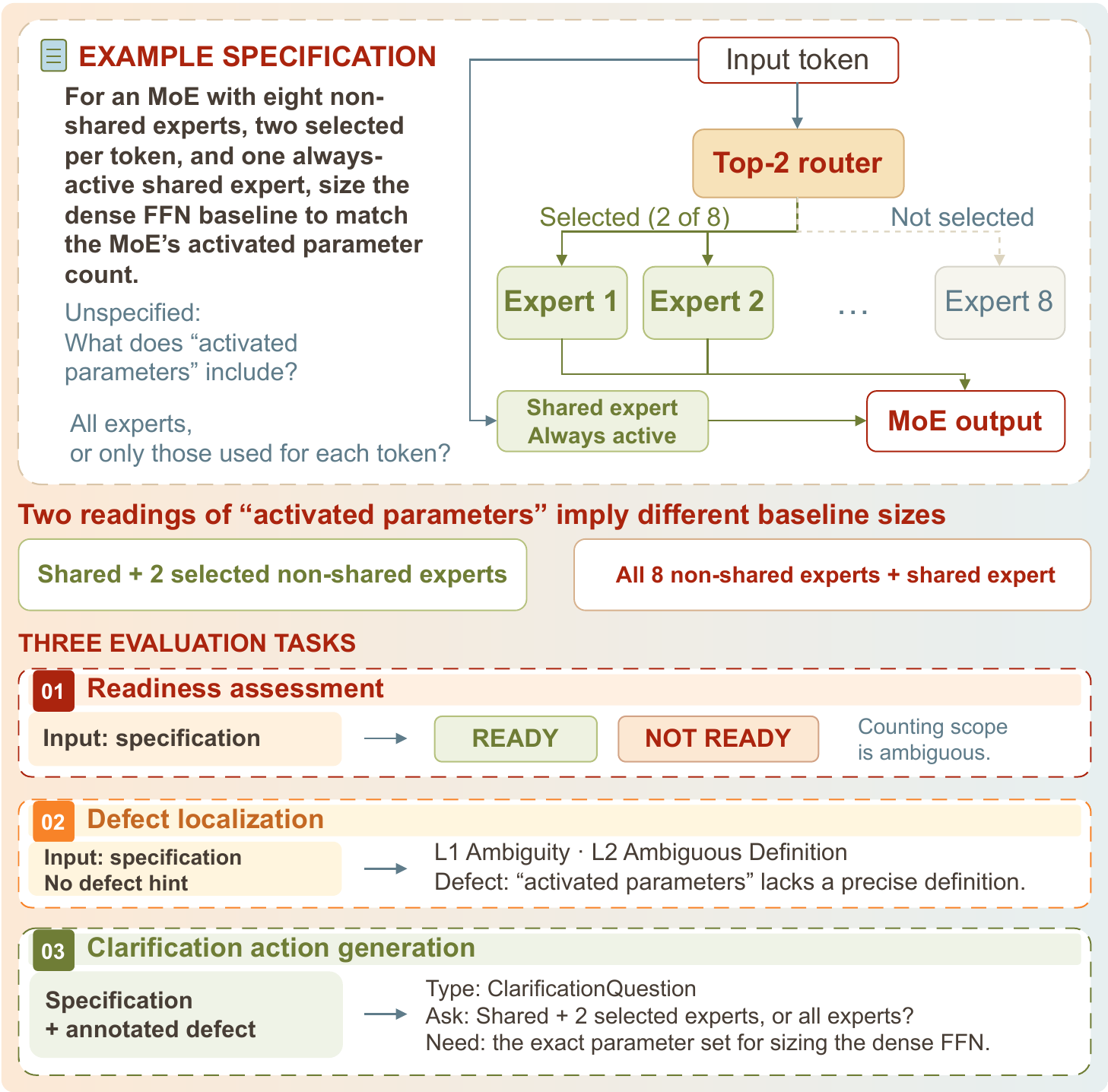}
\caption{
\textbf{(Top)} An ambiguous definition of ``activated parameters''
leaves the FFN baseline size undetermined.
    \textbf{(Bottom)} \ourbench{} evaluates readiness assessment,
    defect localization, and clarification action generation.
    Only clarification receives the annotated defect.
}

    \label{fig:teaser}
\end{figure}

\paragraph{Scientific Execution and Artifact Verification.}
Scientific benchmarks evaluate capabilities such as inspiration-based reasoning and experiment design from paper context~\citep{liu2025researchbench,zhao2025abgen}.
Execution benchmarks assess research replication (PaperBench),
machine learning engineering (MLE-bench), and scientific tasks
through generated programs and execution results
(ScienceAgentBench)~\citep{starace2025paperbench,
chan2025mlebenchevaluatingmachinelearning,
chen2025scienceagentbenchrigorousassessmentlanguage}.
Beyond task completion, work examines whether execution
faithfully reflects the described research.
An analysis of agent-generated research in
ResearchArena~\citep{zhang2026fartrueautoresearch}
identifies mismatches between research plans and their execution, while paper--code consistency benchmarks detect discrepancies between completed artifacts~\citep{baumgartner2026scicoqa,xu2026papers}.
These evaluations examine execution outcomes and artifact alignment, leaving open whether the research specification itself sufficiently determines the intended method.
\ourbench{} addresses this question before implementation, separating codification-readiness assessment, defect localization, and clarification action generation.
Our executable validation connects specification gaps to implementation fidelity, showing that resolving the target gap can improve faithful implementation of the corresponding method detail (Appendix~\ref{app:executable_validation}).

\paragraph{Specification Defects and Clarification.}
Requirements engineering treats ambiguity, incompleteness, and
inconsistency as threats to reliable
implementation~\citep{sommerville1997requirements,
berry2004ambiguity,zave1997classification}.
Recent LLM work studies unclear instructions in dialogue,
tool use, and software
development~\citep{wang2025learning,zhang2025modeling,
larbi2025prompts,vijayvargiya2025interactive}.
SciConvBench~\citep{somasekharan2026sciconvbench} evaluates
multi-turn clarification of missing information and conflicting
requirements in computational-science task formulation.
SpecBench~\citep{hamblin2026specbench} evaluates defect
identification in software RFCs, while
ClarifyCodeBench~\citep{fang2026clarifycodebenchevaluatingllmsclarifying}
evaluates clarification of ambiguous code-generation requirements.
\ourbench{} focuses on research-method specifications, where
runnable implementations may instantiate different methodological
choices. It grounds target defects in evidence-supported resolutions
and separates defect discovery from clarification given the defect.

\section{\ourbench{}}
\label{sec:benchmark}
\ourbench{} evaluates specification readiness before codification through three diagnostic tasks over evidence-grounded, single-defect research-method specifications.

\subsection{Task Formulation}
\label{sec:task_formulation}
We view a research idea as comprising both a scientific objective and a proposed methodological mechanism. \ourbench{} evaluates whether the proposed methodological mechanism contains sufficient information for faithful
codification by a competent implementer or coding agent, rather than assessing
novelty or revising the scientific direction. We define an
\textit{idea specification} as a description of how a research idea or method
is intended to be implemented. It is \textit{codification-ready} when a
competent implementer can construct a faithful initial implementation or
experimental prototype without unsupported assumptions about the core method.
Routine hyperparameters, engineering details, and explicitly open design
choices need not be fixed. A \textit{specification defect} is an omission,
ambiguity, or internal inconsistency that leaves a method-defining decision
underdetermined. Appendix~\ref{app:scope} further clarifies the scope of
\textit{research-idea specification} and the retrospective reconstruction
setting.

We formalize each instance as a \textsc{NotReady} specification
$x_i^{-}\in\mathcal{X}$ containing exactly one target defect
$d_i=(z_i^{(1)},z_i^{(2)},e_i)$, together with a corresponding
\textsc{Ready} specification $x_i^{+}$ obtained by resolving that defect.
Here, $z_i^{(1)}$ and $z_i^{(2)}$ are the Level-1 and Level-2 taxonomy labels,
and $e_i$ describes the unresolved implementation decision. The
\numLevelOneCategories{} Level-1 types are \textit{Ambiguity},
\textit{Incompleteness}, and \textit{Inconsistency}. Their
\numLevelTwoCategories{} Level-2 categories are defined in
Table~\ref{tab:final-taxonomy}. Each instance also includes a gold
clarification action $a_i$. If a source contains multiple independent defects,
we split it into single-target instances. This design isolates recovery of a specific implementation-critical
defect for controlled evaluation. 
We separately evaluate a controlled two-defect setting
(Appendix~\ref{app:multi_defect_ablation}).

\paragraph{Readiness Assessment.}
Given one specification $x_i\in\{x_i^{-},x_i^{+}\}$, Task~1 predicts
$\hat{y}_i\in\{\textsc{Ready},\textsc{NotReady}\}$ and generates
a rationale. Each specification is presented independently,
so the model never observes a \textsc{Ready} specification together
with its underspecified counterpart.

\paragraph{Defect Localization.}
Given an underspecified specification $x_i$, Task~2 returns one diagnosis
$\hat{d}_i=(\hat{z}_i^{(1)},\hat{z}_i^{(2)},\hat{e}_i)$: a Level-1 label, a
Level-2 label, and a natural-language description of the target defect.

\paragraph{Clarification Action Generation.}
\label{sec:task3}
Given $x_i$ and the gold target-defect description $e_i$, Task~3 returns one
clarification action
$\hat{a}_i=(\hat{\tau}_i,\hat{q}_i,\hat{r}_i)$, where $\hat{\tau}_i$ is the
action type,
$\hat{q}_i$
is the concrete natural-language action, and
$\hat{r}_i$ specifies the information expected from carrying out the action.
The action type $\hat{\tau}_i$ specifies how the missing information is to be
obtained. We use two types: \textsc{ClarificationQuestion}, which asks a
targeted question to an author or implementer, and \textsc{EvidenceSeeking},
which directs the model to inspect an artifact such as paper text, source code,
configuration files, data documentation, or experiment logs. The
natural-language action $\hat{q}_i$ is the actual question or inspection
instruction, such as asking for a missing hyperparameter, a preprocessing rule,
an algorithmic choice, or the artifact evidence needed to determine such a
detail.

\begin{table}[!t]
\centering
\small
\setlength{\tabcolsep}{0pt}
\renewcommand{\arraystretch}{1.10}

\definecolor{datasetgroupbg}{HTML}{FCF2D6}
\definecolor{datasettotalbg}{HTML}{EDF0DF}

\begin{tabularx}{\columnwidth}{@{}Xr@{}}
\toprule
\textbf{Source} & \textbf{\# Instances} \\
\midrule

\rowcolor{datasetgroupbg}[0pt][0pt]
\multicolumn{2}{@{}l@{}}{\textit{Real-world instances}} \\
GitHub issues & \numGitHubInstances{} \\
Reproducibility reports & \numRealReproInstances{} \\
\textbf{Real-world subtotal} & \textbf{\numRealInstances{}} \\

\addlinespace[3pt]
\rowcolor{datasetgroupbg}[0pt][0pt]
\multicolumn{2}{@{}l@{}}{\textit{Controlled synthetic instances}} \\
Reproducibility reports & \numSyntheticReproInstances{} \\
Ideation--execution trajectories & \numIdeationInstances{} \\
\textbf{Synthetic subtotal} & \textbf{\numSyntheticInstances{}} \\

\midrule
\rowcolor{datasettotalbg}[0pt][0pt]
\textbf{Total} & \textbf{\numInstances{}} \\
\bottomrule
\end{tabularx}

\caption{
\textbf{Composition of \ourbench{}.}
Real-world instances capture evidence-supported specification gaps.
Controlled synthetic instances alter one implementation-critical
detail in codification-ready references.
}
\label{tab:dataset_snapshot}
\end{table}

\subsection{Data Collection}
\label{sec:data_collection}
\paragraph{Data Sources.}
We combine resolved real-world gaps with controlled synthetic defects.
Papers, codebases, reproducibility reports, and issue discussions provide
retrospective evidence for either an implementation-blocking gap and its
resolution or a codification-ready reference for controlled construction.
From these materials, we reconstruct self-contained, implementation-facing
specifications containing the information required for faithful codification.

These are not verbatim handoff records.
In the main benchmark tasks, evaluated models receive neither
the downstream artifacts nor the resolution evidence. Thus, \ourbench{} uses
retrospectively grounded instances to evaluate readiness, blocker
localization, and clarification rather than reconstructing original handoff
transcripts.

Real-world instances come from reproducibility reports and GitHub issues,
while synthetic instances modify codification-ready references from successful
reproductions and executed research projects~\citep{si2025ideation}. We retain
only atomic, implementation-relevant, evidence-supported defects and exclude
issues involving environments, resources, runtime, or credentials. Each \textsc{NotReady} specification has a \textsc{Ready} counterpart resolving only the target defect (Appendix~\ref{app:ready_counterparts}).

\paragraph{Reproducibility Reports.}
We collect 396 reports from the ML Reproducibility
Challenge~\citep{joelle_pineau_2019_3158244,koustuv_sinha_2021_4833117,koustuv_sinha_2022_6574723,koustuv_sinha_2023_8200058},
ECIR~\citep{Jose2020-wb,Hiemstra2021-hj,Hagen2022-gj,Kamps2023-ny,Goharian2024-yw,Hauff2025-ac},
and TMLR\footnote{\url{https://jmlr.org/tmlr/papers/}} (2022--2025).
After converting them to Markdown with MinerU, we retain 174 reports covering a single open-access paper with an open-source implementation.
DeepSeek-V4-Pro routes each paper--report pair to \textit{resolved real gap},
\textit{synthetic controlled}, or \textit{unusable}, and extracts 210
candidate gap mentions
(Figure~\ref{fig:reproducibility-report-routing-prompt}). Extraction and decomposition yield 233 atomic candidates.
Human verification retains \numRealReproInstances{} real-world
instances with self-contained underspecified inputs and
evidence-supported clarifications.

For controlled construction, we use pairs from the same collections routed
to the \textit{synthetic-controlled} track. These pairs document a successful
reproduction or evaluation but contain no resolved method-core gap suitable
for the real-world subset. We reconstruct a codification-ready reference only
when the paper, report, and implementation jointly specify the
method-defining choices needed to recover the reproduced method. 
Independently altering one implementation-critical detail per
instance yields up to five candidates per reference, with
\numSyntheticReproInstances{} instances retained
(Figure~\ref{fig:reproducibility-defect-injection-prompt}).

\paragraph{Real Gaps from GitHub Issues.}
From 1,000 paper-linked repositories published between 2017 and 2025, we
crawl 4,106 closed or answered issues across 50 top-ranked repositories and
prefilter 800 threads containing gap and resolution signals.
DeepSeek-V4-Pro classifies and decomposes these threads
(Figure~\ref{fig:github-real-gap-annotation-prompt}); GPT-5.5 validation,
paper retrieval, and evidence checks then reduce 152 resolved candidates to
\numGitHubInstances{} atomic benchmark instances. The paper or repository
description provides the underspecified surface form, and the issue thread
provides the supported clarification.

\paragraph{Synthetic-Controlled Instances from Ideation--Execution
Trajectories.}
We use 43 executed projects from the AI-Researcher execution
study~\citep{si2025ideation}, including 19 human- and 24 LLM-generated ideas.
Each assigned idea was implemented by an expert researcher and documented
through a final paper and codebase. Because the protocol prohibited
substantial changes to the proposed method and reported modifications mainly
concerned experimental details rather than the core algorithm, we use the
final paper--code pair, rather than the original idea, as an
execution-grounded source artifact. From each pair, we reconstruct a structured reference specification covering
the task, inputs and outputs, core method, model or algorithm, training
procedure, data and preprocessing, evaluation protocol, and implementation
details needed to recover the executed method. We then generate up to five
controlled candidates per project by removing or abstracting exactly one
implementation-critical detail while preserving all remaining content,
yielding \numIdeationInstances{} controlled synthetic instances.

\begin{figure*}[!t]
    \centering
    \includegraphics[width=\textwidth]{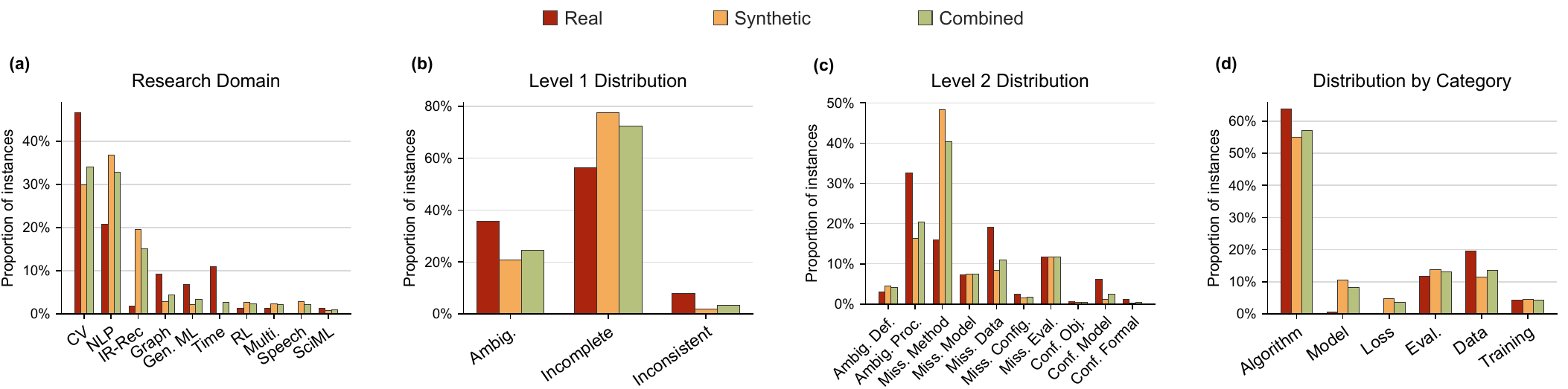}
    \caption{
    Dataset analysis of \ourbench{} across real-world, controlled synthetic, and
    combined subsets.
    (a) Distribution across research domains.
    (b,c) Level-1 defect type and Level-2 defect category distributions.
    (d) Distribution of specification defects across methodological components.
    }
    \label{fig:data_analysis}
\end{figure*}

\paragraph{Human Verification.}
All three construction paths use the same two-stage review protocol. A
primary annotator screens every candidate, and a second independently reviews
all provisionally retained and uncertain cases. For real-world candidates,
reviewers verify that each instance captures a genuine method-core gap,
isolates one atomic and self-contained target defect, and provides sufficient
evidence for a concrete clarification without unsupported inference. For
controlled synthetic candidates, they additionally verify that the reference
is codification-ready, exactly one method-defining detail is altered, all
non-target information is preserved, and the hidden detail is directly
recoverable from the source artifacts. Disagreements are resolved against the
source evidence and predefined inclusion criteria; cases that remain
non-atomic, inconsistent, or insufficiently supported are discarded. Details appear in Appendix~\ref{app:source-verification}.

\subsection{Dataset Statistics}

As shown in Table~\ref{tab:dataset_snapshot}, \ourbench{} contains
\numRealInstances{} real-world and \numSyntheticInstances{} controlled
synthetic instances. It spans ten research domains, including computer vision,
natural language processing, information retrieval, graph learning, and
time-series modeling (Figure~\ref{fig:data_analysis}(a)). Real-world instances
concentrate in computer vision and NLP, where open implementations and issue
discussions are more available; synthetic instances broaden the benchmark's
domain coverage. Incompleteness is the most frequent Level-1 type in both subsets,
while ambiguity accounts for a larger proportion of real-world
than synthetic instances (Figure~\ref{fig:data_analysis}(b)).
At Level~2, Ambiguous Procedure is the most frequent category
in the real-world subset, whereas Missing Method Procedure
dominates the synthetic subset
(Figure~\ref{fig:data_analysis}(c)). Algorithm-related defects dominate both subsets, while the remaining
component distributions differ: data-related defects are more frequent
in real-world instances, whereas model- and loss-related defects are
more frequent in synthetic instances
(Figure~\ref{fig:data_analysis}(d)).
Evaluation-related defects occur in both subsets, while training-related
defects remain relatively infrequent.

\subsection{Benchmark Validation}
\label{sec:benchmark-validation}

We conduct two human studies to validate the benchmark. First, two
machine-learning researchers who are blinded to source type evaluate 50
real-world and 50 controlled-synthetic instances. Positive judgments for gap
validity, implementation criticality, clarification sufficiency, and realism
are 92/92/96/93\% for real-world instances and 96/94/98/91\% for synthetic
instances. These results support the controlled construction procedure and
show that synthetic instances preserve the key properties of naturally
occurring gaps. Second, two graduate-level annotators independently label all
\numRealInstances{} real-world instances and a fixed sample of
\numSyntheticEvalInstances{} synthetic instances. Cohen's $\kappa$ reaches
0.84/0.92 for readiness, 0.89/0.89 for Level-1 labels, and 0.92/0.85 for
Level-2 labels on the real and synthetic subsets, respectively. Target-defect
agreement is 87.7/84.0\%. A target-uniqueness audit further retains only
instances whose annotated blocker is valid, primary, and unique. Before
adjudication, 94.5\% of the original targets satisfy these criteria. Overall,
the results support annotation reliability, although target-defect boundaries
are slightly less stable for synthetic instances. Full protocols appear in
Appendices~\ref{app:synthetic_validation},
\ref{app:target_uniqueness}, and~\ref{app:annotation-quality}.

\paragraph{Validation on Original Proposals.}
We additionally examine 24 original LLM-generated proposals
selected for implementation in the AI-Researcher execution
study~\citep{si2025ideation}.
We identify candidate gaps from each proposal without access
to its final paper or code, then check whether the same decision
was subsequently specified during execution.
Human verification confirms such gaps in seven proposals
(29.2\%), showing that implementation-critical gaps also occur
in original proposals selected for execution.
Full procedures and confirmed cases appear in
Appendix~\ref{app:original_proposal_validation}.

\section{Experiments}
\label{sec:experiments}

\begin{table*}[t]
\centering

\small
\setlength{\tabcolsep}{3pt}
\renewcommand{\arraystretch}{1.08}
\definecolor{modelgroupbg}{HTML}{FCF2D6}

\begin{tabular*}{\textwidth}
{@{\extracolsep{\fill}}l*{6}{c}@{}}
\toprule
\multirow{2}{*}{\textbf{Model}}
& \multicolumn{2}{c}{\textbf{Task~1: Macro-F1 $\uparrow$}}
& \multicolumn{2}{c}{\textbf{Task~2: Macro DRR $\uparrow$}}
& \multicolumn{2}{c}{\textbf{Task~3: Macro-CAS $\uparrow$}} \\
\cmidrule(lr){2-3}
\cmidrule(lr){4-5}
\cmidrule(lr){6-7}
& \textbf{Real} & \textbf{Synth.}
& \textbf{Real} & \textbf{Synth.}
& \textbf{Real} & \textbf{Synth.} \\
\midrule

\rowcolor{modelgroupbg}
\multicolumn{7}{@{}l}{\textit{Frontier models}} \\
GPT-5.6-Sol
& \textbf{67.5} & \textbf{86.4}
& \textbf{9.6} & \textbf{12.2}
& \textbf{80.6} & \textbf{96.2} \\
Claude Sonnet 5
& \underline{59.3} & \underline{76.2}
& \underline{6.8} & \underline{11.4}
& \underline{76.8} & \underline{94.5} \\
Gemini 3.1 Pro Preview
& 45.8 & 67.3 & 3.4 & 7.6 & 68.9 & 91.6 \\
DeepSeek-V3.2
& 44.5 & 67.2 & 5.9 & 8.0 & 62.9 & 93.0 \\

\addlinespace[3pt]
\rowcolor{modelgroupbg}
\multicolumn{7}{@{}l}{\textit{Open-weight reasoning models}} \\
Qwen3.5-397B-A17B
& 57.5 & 55.9 & 3.3 & 4.6 & 72.0 & 89.0 \\
GLM-5.2
& 45.1 & 65.9 & 5.7 & 7.0 & 63.6 & 77.0 \\
DeepSeek-R1-0528
& 44.3 & 59.9 & 1.8 & 2.2 & 68.0 & 68.5 \\
Kimi-K3
& 32.7 & 43.1 & 6.6 & 8.2 & 60.6 & 92.5 \\

\addlinespace[3pt]
\rowcolor{modelgroupbg}
\multicolumn{7}{@{}l}{\textit{Open-weight general models}} \\
Gemma-4-31B-IT
& 47.1 & 56.9 & 4.7 & 7.0 & 67.7 & 73.0 \\
Qwen3-8B
& 40.1 & 48.7 & 3.2 & 4.6 & 53.8 & 82.6 \\
Qwen3-32B
& 34.5 & 37.0 & 3.6 & 5.8 & 61.9 & 76.3 \\
Qwen3.5-9B
& 33.3 & 54.3 & 5.3 & 5.5 & 64.6 & 81.2 \\
GPT-OSS-120B
& 30.0 & 44.4 & 2.5 & 3.3 & 53.4 & 65.3 \\

\bottomrule
\end{tabular*}
\caption{
\textbf{Main results on \ourbench{}.}
Real and Synth.\ denote real-world and controlled synthetic subsets.
Task~1 uses 100 \textsc{Ready} and 100 \textsc{NotReady}
specifications per subset.
Tasks~2 and~3 use all \numRealInstances{} real-world instances
and \numSyntheticEvalInstances{} synthetic instances.
The primary metrics are Macro-F1 for readiness assessment,
Macro Defect Recovery Rate (Macro DRR) for defect localization,
and Macro Clarification Action Success Rate (Macro-CAS)
for clarification action generation.
All values are percentages.
\textbf{Bold}: best; \underline{underline}: second best.
Within each group, models are ordered by real-world Task~1 performance.
Full component metrics and sample definitions appear in
Appendix~\ref{eval}.
}
\label{tab:main_results}
\end{table*}

\subsection{Experimental Setup}
\label{sec:evaluated-models}

We evaluate 13 proprietary and open-weight LLMs listed in
Table~\ref{tab:main_results} with fixed task-specific prompts
and structured outputs. Decoding, deployment, and model-access
details are provided in Appendix~\ref{app:inference}.
For the three benchmark tasks, semantic and rubric-based
evaluation uses Claude Opus~4.8, validated against adjudicated
human annotations (Appendix~\ref{app:judge_validation}).

We report Macro-F1 for readiness assessment, Macro Defect Recovery Rate
(Macro DRR) for defect localization, and Macro Clarification Action Success
Rate (Macro-CAS) for clarification generation. Macro DRR requires recovering
the annotated blocker and both taxonomy labels. Macro-CAS requires the action to address the blocker,
request sufficient information for resolution, and avoid
unsupported assumptions. RGS evaluates generated Task~1 rationales against
the gold readiness rationale in terms of label support, target-defect
match, and faithfulness. Formal definitions and component metrics appear in
Appendix~\ref{eval}.

Because instances may share a repository, source paper, or executed project,
Appendix~\ref{app:source_uncertainty} reports source-clustered bootstrap
confidence intervals, source-balanced estimates, and paired source-level
comparisons. To separate blocker discovery from clarification formulation,
we also evaluate an end-to-end variant on the real-world Task~3 set. It uses
the same GPT-5.6-Sol model, output schema, decoding configuration, and
evaluation protocol as the defect-guided setting, but does not receive the
annotated blocker.

\subsection{Main Results}
\label{sec:results}

We organize the results around three questions: whether models can assess
codification readiness, identify the unresolved implementation blocker,
and formulate an effective clarification once that blocker is known.
Table~\ref{tab:main_results} summarizes the three benchmark tasks.

\begin{rqbox}{RQ1: Readiness Assessment}
Can models reliably assess codification readiness?
\end{rqbox}
\label{sec:task1_results}

Readiness assessment remains unreliable, especially on real-world gaps.
GPT-5.6-Sol achieves 67.5 Macro-F1 on real-world instances and 86.4 on
controlled synthetic instances. On the real-world subset, it accepts 31\%
of underspecified specifications and rejects 34\% of codification-ready
ones. Moreover, the highest Reason Grounding Score is only 0.36 on
real-world instances and 0.49 on synthetic instances, showing that
correct labels are often not accompanied by rationales that support the
decision, identify the relevant blocker, and remain faithful to the
specification.

\begin{rqbox}{RQ2: Defect Localization}
Can models identify the implementation-critical blocker?
\end{rqbox}
\label{sec:task2_results}

Blocker localization is the most difficult capability in \ourbench{}.
On real-world instances, GPT-5.6-Sol reaches 60.1 Level-1 accuracy and 25.2 Level-2 accuracy, but only 16.0 Loc-Acc and 9.6 Macro DRR.
Models therefore often predict plausible defect categories without recovering the unresolved method-defining decision. 
Performance is higher on controlled synthetic instances, whose target defects are generally more
explicit. 

To separate defect localization from taxonomy prediction, we evaluate GPT-5.6-Sol on all 163 real-world instances under two output requirements.
The original Task~2 setting requests a defect description and both taxonomy labels, whereas the taxonomy-free setting requests only the description.
Both conditions use the same evaluator and same-blocker criterion, scoring
the description without requiring correct taxonomy labels.
Removing category prediction raises localization accuracy from 16.0\% to 40.5\%.
Taxonomy prediction therefore adds difficulty, while defect discovery remains
challenging without it. 
Full protocols appear in
Appendix~\ref{app:taxonomy_free_localization}.

A separate human baseline shows stronger localization performance than model predictions on matched sampled records (Appendix~\ref{app:human_baseline}). 
An inference-time search ablation further shows that broader
candidate generation improves candidate recall but does not
improve final localization after self-selection
(Appendix~\ref{app:inference_time_localization}).

\begin{rqbox}{RQ3: Clarification}
Once the blocker is known, can models elicit the information
needed to resolve it?
\end{rqbox}
\label{sec:task3_results}
When supplied with the annotated blocker, GPT-5.6-Sol reaches 80.6 Macro-CAS on real-world instances and 96.2 on controlled synthetic
instances. 
On the real-world subset, its No-Assumption score is 95.7, compared with 80.4 Sufficiency, indicating that remaining failures
mainly reflect incomplete clarification rather than unsupported assumptions. 

To isolate the effect of blocker availability, we compare two clarification
settings on all 163 real-world Task~3 instances using GPT-5.6-Sol with the
same output schema, decoding settings, and evaluation metrics.
\textsc{End-to-End} receives only the underspecified specification,
whereas \textsc{Defect-Guided} additionally receives the annotated blocker.
As shown in Table~\ref{tab:information_bottleneck}, providing the blocker raises Macro-CAS from 13.6 to 80.6 and Sufficiency from 8.6 to 80.4.
No-Assumption remains nearly unchanged (96.3 vs.\ 95.7).
For GPT-5.6-Sol, this controlled comparison identifies blocker availability as the primary bottleneck in generating effective clarification actions.
Together with the consistent cross-task pattern across all 13 models,
the results suggest that current LLMs are substantially better at acting
on a known blocker than at discovering it from the specification alone.

\begin{table}[t]
\centering
\footnotesize
\setlength{\tabcolsep}{3pt}
\renewcommand{\arraystretch}{1.15}

\begin{tabular*}{\columnwidth}
{@{\extracolsep{\fill}}lccc@{}}
\toprule
\textbf{Metric}
& \textbf{End-to-End}
& \textbf{Defect-Guided}
& \textbf{$\Delta$} \\
\midrule

Macro-CAS $\uparrow$
& 13.6 & \textbf{80.6} & +67.0 \\

Sufficiency $\uparrow$
& 8.6 & \textbf{80.4} & +71.8 \\

No-Assumption $\uparrow$
& 96.3 & 95.7 & $-0.6$ \\

\bottomrule
\end{tabular*}

\caption{
\textbf{Effect of blocker availability.}
GPT-5.6-Sol on 163 real-world instances.
Scores are percentages; $\Delta$ is Defect-Guided minus
End-to-End, in percentage points.
}
\label{tab:information_bottleneck}
\end{table}

Full protocols appear in Appendix~\ref{app:information_bottleneck}.

A complementary study executes generated evidence-seeking
actions over paper artifacts. Given the annotated defect,
the model achieves evidence-supported resolution in 25\%
of 20 selected cases
(Appendix~\ref{app:artifact_execution}).

\paragraph{Source-Level Robustness.}
Source-clustered analyses preserve the main findings
(Appendix~\ref{app:source_uncertainty}). Source-balanced GPT-5.6-Sol
estimates differ from the instance-level results by at most 0.6 points,
although five of six paired comparisons with Claude Sonnet~5 do not
yield reliable differences. In contrast, the
\textsc{Defect-Guided} advantage remains large and reliable
($\Delta=67.0$, $95\%~\mathrm{CI}=[59.0,75.2]$, $p<0.001$).

\subsection{Error Analysis and Representative Cases}
\label{sec:error_analysis}

We analyze GPT-5.6-Sol to characterize these failure patterns. Full
definitions and statistics appear in Appendix~\ref{app:error_analysis}.
On real-world Task~1 instances, 67.5\% of predictions receive the correct
label, but only 1.5\% are both correct and fully grounded. For Task~2, the
model recovers the annotated blocker in only 16.0\% of instances. Another
17.8\% identify a neighboring decision, while 66.3\% identify a different
blocker. Once the blocker is provided, the main remaining error is
insufficient clarification, which occurs in 19.6\% of instances, compared
with 4.3\% of actions that impose unsupported assumptions.
Table~\ref{tab:qualitative_cases} illustrates these patterns. Overall, the
dominant failure is identifying the correct unresolved method-defining
decision.

\begin{table}[t]
\centering
\footnotesize
\setlength{\tabcolsep}{1.5pt}
\renewcommand{\arraystretch}{1.15}
\renewcommand{\tabularxcolumn}[1]{m{#1}}

\begin{tabularx}{\columnwidth}
{@{}>{\raggedright\arraybackslash}Xccccc@{}}
\toprule
\textbf{Metric}
& \textbf{Direct}
& \textbf{Assist.}
& \textbf{$\Delta$}
& \textbf{95\% CI}
& \textbf{$p_{\mathrm{adj}}$} \\
\midrule

\textsc{Ready} Rate $\uparrow$
& 14 & 98 & +84
& [72.0, 92.0] & $<0.001$ \\

Completeness $\uparrow$
& 30 & 98 & +68
& [58.0, 75.0] & $<0.001$ \\

Missing Detail Recovery $\uparrow$
& 14 & 98 & +84
& [72.0, 92.0] & $<0.001$ \\

Unsupported Assumption Rate $\downarrow$
& 6 & 0 & $-6$
& [-12.0, 0.0] & 0.250 \\

\bottomrule
\end{tabularx}

\caption{
\textbf{Oracle clarification on 50 real-world instances.}
Scores are percentages; $\Delta$ is Assisted minus Direct,
in percentage points.
We report paired-bootstrap 95\% CIs and Holm-adjusted
$p$-values. Test details appear in
Appendix~\ref{app:clarification_utility}.
}
\label{tab:clarification_effect_significance}
\end{table}

\subsection{Oracle Clarification Utility}
\label{sec:oracle-clarification-utility}
To test whether resolving a specification gap improves downstream
codification, we conduct an oracle study on a taxonomy-stratified sample of
50 real-world instances. Using GPT-5.6-Sol with identical decoding settings,
\textsc{Direct Generation} receives only the underspecified specification.
\textsc{Clarification-Assisted Generation} additionally receives the
annotated defect, clarification action, and oracle resolution. Two human
annotators who are blinded to the generation condition evaluate the resulting
specifications. Full definitions and protocols appear in
Appendix~\ref{app:clarification_utility}. As shown in Table~\ref{tab:clarification_effect_significance}, oracle
clarification raises the \textsc{Ready} rate from 14\% to 98\%,
completeness from 30\% to 98\%, and missing-detail recovery from 14\% to
98\%. All three improvements are statistically significant under paired
testing. The unsupported-assumption rate decreases from 6\% to 0\%.
However, this difference is based on only three discordant instances and is
not statistically significant under a two-sided exact McNemar test
($p_{\mathrm{adj}}=0.250$). It should therefore be interpreted
descriptively. These results show that the model can produce a
codification-ready specification once the missing information is supplied.
The remaining challenge is to identify the gap and obtain the information
needed to resolve it.

We further test whether resolving the target gap improves executable code
on 20 controlled instances from ideation--execution trajectories.
Each instance concerns a bounded implementation component with an available
reference codebase. We define instance-specific tests and target-fidelity
criteria from the final paper and code before inspecting generated
implementations. GPT-5.6-Sol implements the \textsc{Direct Generation} and
\textsc{Clarification-Assisted Generation} specifications using the same
implementation prompt and decoding settings. The code-generation stage
receives the resulting specification, with the reference paper and code
withheld.

Table~\ref{tab:executable_validation} shows that oracle clarification raises
the all-tests-pass rate from 45\% to 85\% and target fidelity from 30\% to
90\%. Target fidelity checks whether the code implements the
evidence-supported method detail through output equivalence, targeted tests,
or blinded inspection against a predefined criterion.
Although 70\% of directly generated implementations run successfully,
30\% implement the intended target detail.
Oracle clarification thus improves both code executability and fidelity
to the reference method. Full protocols appear in
Appendix~\ref{app:executable_validation}.
Because both utility studies provide the gold resolution, they measure the
utility of successful clarification under oracle conditions rather than
end-to-end agent performance.

\begin{table}[t]
\centering
\footnotesize
\setlength{\tabcolsep}{4pt}
\renewcommand{\arraystretch}{1.15}

\begin{tabular*}{\columnwidth}
{@{\extracolsep{\fill}}lcc@{}}
\toprule
\textbf{Metric}
& \textbf{Direct}
& \textbf{Assisted} \\
\midrule

Runnable $\uparrow$
& 70.0 & \textbf{95.0} \\

All Tests Pass $\uparrow$
& 45.0 & \textbf{85.0} \\

Target Fidelity $\uparrow$
& 30.0 & \textbf{90.0} \\

Further Clarification $\downarrow$
& 35.0 & \textbf{5.0} \\

\bottomrule
\end{tabular*}

\caption{
\textbf{Executable validation on 20 controlled instances.}
Direct and Assisted denote Direct Generation and
Clarification-Assisted Generation.
We evaluate bounded components using predefined tests and fidelity
criteria. All values are percentages.
}
\label{tab:executable_validation}
\end{table}

\section{Conclusion}
\label{sec:conclusion}
We introduced \ourbench{}, a benchmark of \numInstances{}
evidence-grounded, single-defect instances for evaluating the
codification readiness of research-method specifications through
readiness assessment, defect localization, and clarification
action generation.
Across \numModels{} LLMs, the strongest model achieves 9.6
Macro DRR on real-world localization and 80.6 Macro-CAS when
given the annotated defect, while readiness judgments remain
frequently incorrect or weakly grounded.
Taxonomy prediction adds difficulty, but defect discovery
remains challenging even without category prediction.
Together with the defect-guided clarification comparison,
these results identify defect discovery as a key bottleneck
for research automation.
Oracle clarification raises the codification-ready rate from
14\% to 98\%, demonstrating the value of supplying the correct
missing information.
These findings motivate readiness checks and evidence-supported
clarification before implementation.

\section*{Limitations}
\ourbench focuses on assessing a specification before implementation. This
scope leaves several questions open. Its evidence-resolved, single-defect
instances are concentrated in AI, NLP, and machine learning;
extending collection to other computational sciences, naturally occurring
early-stage ideas, and specifications with interacting defects would test how
codification readiness and the taxonomy transfer. Clarification is currently
evaluated as a single action, while the oracle utility study supplies the gold
resolution. Future systems could instead conduct multi-turn clarification,
choose between asking a person and inspecting an artifact, reconcile conflicting
evidence, and determine when enough information has been obtained. Finally,
codification readiness is evaluated before execution rather than through a full
implementation of every specification. Integrating this diagnostic as a gate in
research agents would enable direct measurement of whether clarification reduces
implementation divergence, unsupported assumptions, and human correction cost,
and whether these gains improve final scientific outcomes.
\section*{Ethical Considerations}

\ourbench{} is constructed from publicly available research artifacts,
including papers, codebases, reproducibility reports, GitHub repository
metadata, and issue discussions. We use these materials only to identify
methodological specification gaps and evidence-supported resolutions. We
exclude issues involving credentials, private access, runtime environments,
or other non-methodological concerns. The released dataset will preserve
source attribution, follow applicable licenses and platform policies, and
exclude personal identifiers unnecessary for understanding the gaps.

The benchmark evaluates whether LLMs recognize when a methodological
specification requires clarification. It should not be used to assess the
scientific value of an idea, replace expert review, or justify autonomous
implementation without human oversight. A high score does not demonstrate
reliable end-to-end research ability, and the oracle clarification experiments
represent an upper bound in which the correct missing information is supplied.
Incorrect readiness judgments may introduce unsupported methodological
choices or unnecessary clarification requests.

Some evaluation components rely on LLM-based judgments. Although these
judgments are validated against human annotations, evaluator models may still
exhibit systematic or model-family biases. We release the evaluation prompts
and criteria to support auditing and alternative evaluation. In practical
research agents, specification-readiness decisions should remain transparent,
evidence-grounded, and subject to human review, particularly in high-stakes
applications.

{
  \small
  \renewcommand{\bibsection}{\section*{References}}
  \bibliographystyle{ieeenat_fullname}
  \bibliography{main}
}

\clearpage
\onecolumn
\raggedbottom
\pagestyle{supp}
\titleformat{\section}{\bfseries\fontsize{12.5}{15}\selectfont\color{Accent}}
  {\thesection}{0.8em}{}

{\bfseries\fontsize{12.5}{15}\selectfont\color{Accent}
\phantomsection\addcontentsline{toc}{section}{Appendix}
Appendix\par}
\vspace{4pt}

\begin{tcolorbox}[enhanced,colback=AccentLight,colframe=AccentLight,
  boxrule=0pt,arc=0pt,borderline west={2.2pt}{0pt}{Accent},
  left=12pt,right=10pt,top=6pt,bottom=6pt,before skip=2pt,after skip=10pt]
{\bfseries\small\color{Accent}Contents}\par\vspace{3pt}
\begingroup
\small
\makeatletter
\newcommand{\suppoutlineentry}[2]{%
  \par\begingroup
  \let\texorpdfstring\@secondoftwo
  \leftskip=\dimexpr#1\dimexpr1em\relax\relax
  \rightskip=1.6em
  \parfillskip=-\rightskip
  \parindent=0pt
  \ifnum#1=0\relax\bfseries\fi
  \hyperref[#2]{%
    \makebox[2.4em][l]{\ref*{#2}}\nameref*{#2}%
    \nobreak\leaders\hbox{\normalfont\kern0.25em.\kern0.25em}\hfill
    \nobreak\makebox[1.6em][r]{\pageref*{#2}}}%
  \par\endgroup
}
\makeatother
\setlength{\baselineskip}{13pt}
\noindent\begin{minipage}[t]{0.48\linewidth}
\suppoutlineentry{0}{app:additional-distribution}
\suppoutlineentry{1}{app:publication-year}
\suppoutlineentry{1}{app:construct_validity}
\suppoutlineentry{1}{app:error_analysis}
\suppoutlineentry{1}{app:qualitative_examples}
\suppoutlineentry{0}{supp:taxonomy-development-and-validation}
\suppoutlineentry{1}{supp:taxonomy-construction}
\suppoutlineentry{1}{supp:annotation-instructions}
\suppoutlineentry{1}{supp:detailed-label-definitions}
\suppoutlineentry{1}{supp:inconsistency}
\suppoutlineentry{0}{app:ready_counterparts}
\suppoutlineentry{0}{app:additional_analysis}
\suppoutlineentry{1}{app:inference}
\suppoutlineentry{1}{app:information_bottleneck}
\suppoutlineentry{1}{app:taxonomy_free_localization}
\suppoutlineentry{1}{app:inference_time_localization}
\suppoutlineentry{1}{app:artifact_execution}
\suppoutlineentry{1}{app:executable_validation}
\suppoutlineentry{1}{app:source_uncertainty}
\suppoutlineentry{0}{app:data-annotation-details}
\suppoutlineentry{1}{app:source-verification}
\suppoutlineentry{1}{app:target_uniqueness}
\end{minipage}\hfill
\begin{minipage}[t]{0.48\linewidth}
\suppoutlineentry{1}{app:multi_defect_ablation}
\suppoutlineentry{1}{app:human-annotation-guideline}
\suppoutlineentry{1}{app:annotation-quality}
\suppoutlineentry{0}{eval}
\suppoutlineentry{1}{app:task1_metrics}
\suppoutlineentry{1}{app:task2_metrics}
\suppoutlineentry{1}{app:task3_metrics}
\suppoutlineentry{0}{supp:human-validation}
\suppoutlineentry{1}{app:judge_validation}
\suppoutlineentry{1}{app:synthetic_validation}
\suppoutlineentry{1}{app:original_proposal_validation}
\suppoutlineentry{1}{app:clarification_utility}
\suppoutlineentry{1}{app:human_baseline}
\suppoutlineentry{0}{app:scope}
\suppoutlineentry{0}{app:prompts}
\suppoutlineentry{1}{app:github_issue_prompt}
\suppoutlineentry{1}{app:reproducibility-gap-extraction}
\suppoutlineentry{1}{app:reproducibility-defect-injection}
\suppoutlineentry{1}{app:task1-rgs-judge}
\suppoutlineentry{1}{app:task2-llm-judge}
\suppoutlineentry{1}{app:task3-llm-judge}
\end{minipage}\par
\endgroup

\end{tcolorbox}

\appendix
\renewcommand{\thesection}{\Alph{section}}

\section{Additional Dataset Distribution Analysis}
\label{app:additional-distribution}
\subsection{Publication-Year Distribution}
\label{app:publication-year}

\begin{figure}[!htb]
    \centering
    \includegraphics[width=0.46\linewidth]{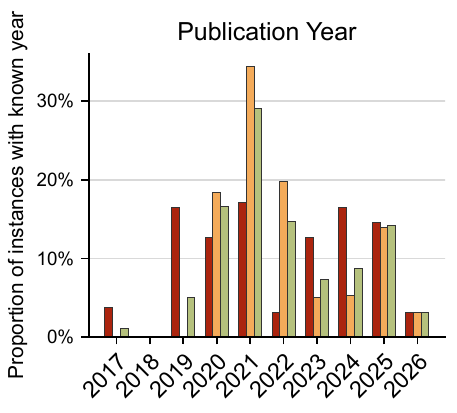}
    \caption{
    Publication-year distribution of \ourbench{} instances with known source-paper years.
    Percentages are normalized within each split after excluding instances with missing publication-year metadata.
    }
    \label{fig:publication-year-appendix}
\end{figure}

Figure~\ref{fig:publication-year-appendix} reports the publication-year distribution for instances with identifiable source-paper years. The distribution shows that \ourbench{} is concentrated in recent machine-learning literature, with most identifiable instances coming from 2020--2025 and a peak in 2021. This pattern is expected given the benchmark's focus on recent reproducibility and specification issues.

\subsection{Construct Validity Analysis: Specification Clarification vs. Idea Refinement}
\label{app:construct_validity}

A potential concern is whether IdeaAMBIG evaluates clarification of
early-stage research ideas or clarification of implementation specifications.
These two settings represent different stages of the scientific workflow.
Early-stage idea refinement concerns open-ended decisions such as research
objectives, hypotheses, motivation, and possible directions, whereas
specification clarification concerns resolving implementation-critical
choices required to faithfully instantiate an already proposed method.

To verify the intended construct of IdeaAMBIG, we conduct a human annotation
study on a stratified sample of benchmark instances. Annotators are asked to
determine whether resolving the target gap requires changing or refining the
scientific idea itself, or whether it only requires specifying missing details
needed for faithful implementation.

\paragraph{Annotation Protocol.}
We randomly sample 100 instances from IdeaAMBIG, including 50 real-world
instances and 50 controlled synthetic instances. Two annotators with machine
learning research experience independently review the original specification,
the target defect, and the supported clarification. For each instance, they
answer the following binary question:

\begin{quote}
\textit{Does resolving this gap require modifying the research idea itself,
such as changing its objective, hypothesis, motivation, or scientific
direction?}
\end{quote}

Instances are categorized as \textsc{Idea-Level Clarification} if the answer
is yes, and \textsc{Specification-Level Clarification} otherwise. The latter
category corresponds to gaps where the research objective remains unchanged
but implementation-critical choices are missing, ambiguous, or inconsistent.
\begin{table}[t]
\centering
\small
\setlength{\tabcolsep}{5pt}
\begin{tabular}{lccc}
\toprule
\textbf{Subset}
& \textbf{Idea-Level}
& \textbf{Specification-Level}
& \textbf{Total}
\\
\midrule

Real-world
& 4 (8\%)
& 46 (92\%)
& 50
\\

Synthetic
& 2 (4\%)
& 48 (96\%)
& 50
\\

\midrule

Overall
& 6 (6\%)
& 94 (94\%)
& 100
\\

\bottomrule
\end{tabular}

\caption{
Construct validity analysis distinguishing early-stage idea refinement
from implementation specification clarification.
Annotators classify whether resolving a benchmark instance requires changing
the research idea itself or only specifying implementation-critical details.
}
\label{tab:construct_validity}
\end{table}
\paragraph{Results.}
Table~\ref{tab:construct_validity} summarizes the annotation results. Across
both subsets, the vast majority of instances are categorized as
\textsc{Specification-Level Clarification}. Only a small fraction requires
changes to the underlying research idea. These results confirm that \ourbench{} primarily measures the ability of
LLMs to identify and resolve implementation-critical gaps in
research-method specifications, rather than general-purpose research idea
refinement.

This distinction is intentional: IdeaAMBIG does not aim to evaluate whether
LLMs can improve the creativity or scientific direction of early-stage ideas.
Instead, it focuses on a later but critical failure mode in automated research
pipelines, where a plausible research idea fails to provide sufficient
specification for faithful codification.

\subsection{Diagnostic Error Analysis}
\label{app:error_analysis}

We analyze GPT-5.6-Sol, the strongest model in the main experiments, to
characterize the failure patterns underlying its aggregate performance.
We reuse the original model outputs and evaluation sets without additional
sampling or generation. Task~1 uses the balanced samples of 100
\textsc{Ready} and 100 \textsc{NotReady} specifications from each subset,
while Tasks~2 and~3 use all 163 real-world instances and the fixed sample
of 200 controlled-synthetic instances.

\paragraph{Analysis Categories.}
For Task~1, we distinguish decision correctness from rationale grounding.
A prediction is \textit{correct and grounded} only when the readiness label
is correct and all three RGS components---label support, blocker match or
correct recognition of its absence, and faithfulness---receive their maximum
score. A correct prediction with at least one lower component score is
classified as \textit{correct but weakly grounded}. The remaining categories
are false-\textsc{Ready} and false-\textsc{NotReady} decisions.

For Task~2, predictions are first separated according to whether they
identify the annotated blocker using the same taxonomy-blind matcher as
Loc-Acc. Correctly localized predictions are then divided by whether both
taxonomy labels are correct. Localization failures are categorized as
neighboring blockers, different blockers, or vague or absent blockers.
These five categories are mutually exclusive.

For Task~3, we report Macro-CAS together with two component error rates:
insufficient clarification, computed as one minus Sufficiency, and
assumption-imposing clarification, computed as one minus No-Assumption.
These two error rates are evaluated independently and may overlap.
Table~\ref{tab:error_decomposition} summarizes the resulting diagnostic
outcomes for all three tasks.

\begin{table*}[t]
\centering
\small
\setlength{\tabcolsep}{0pt}
\renewcommand{\arraystretch}{1.12}
\definecolor{analysisgroupbg}{HTML}{FCF2D6}

\begin{tabularx}{\textwidth}
{@{}X
>{\centering\arraybackslash}p{0.11\textwidth}
>{\centering\arraybackslash}p{0.11\textwidth}@{}}
\toprule
\textbf{Analysis Category}
& \textbf{Real}
& \textbf{Synth.} \\
\midrule

\rowcolor{analysisgroupbg}[0pt][0pt]
\multicolumn{3}{@{}l@{}}{\textit{Task 1: Readiness assessment}} \\
Correct and grounded                 & 1.5  & 17.0 \\
Correct but weakly grounded           & 66.0 & 69.5 \\
False \textsc{Ready}                  & 15.5 & 2.5  \\
False \textsc{NotReady}               & 17.0 & 11.0 \\

\addlinespace[4pt]
\rowcolor{analysisgroupbg}[0pt][0pt]
\multicolumn{3}{@{}l@{}}{\textit{Task 2: Defect localization}} \\
Correct blocker and both taxonomy labels & 4.3  & 12.0 \\
Correct blocker, taxonomy error          & 11.7 & 8.0  \\
Neighboring blocker                      & 17.8 & 6.0  \\
Different blocker                        & 66.3 & 73.0 \\
Vague or no blocker                      & 0.0  & 1.0  \\

\addlinespace[4pt]
\rowcolor{analysisgroupbg}[0pt][0pt]
\multicolumn{3}{@{}l@{}}{\textit{Task 3: Clarification action generation}} \\
Macro-CAS                            & 80.6 & 96.2 \\
Insufficient clarification           & 19.6 & 2.0  \\
Assumption-imposing clarification    & 4.3  & 1.0  \\

\bottomrule
\end{tabularx}

\caption{
\textbf{Diagnostic analysis of GPT-5.6-Sol.}
Task~1 partitions the balanced readiness samples.
Task~2 reports mutually exclusive, micro-averaged localization outcomes.
Task~3 reports Macro-CAS across Level-2 categories and micro-averaged,
potentially overlapping error rates.
All values are percentages, subject to rounding.
}
\label{tab:error_decomposition}
\end{table*}
\paragraph{Readiness Decisions Are Frequently Weakly Grounded.}
As shown in Table~\ref{tab:error_decomposition}, GPT-5.6-Sol makes the
correct readiness decision on 67.5\% of real-world specifications, but only
1.5\% of all predictions are both correct and fully grounded. Most correct
decisions therefore fail at least one rationale criterion. Errors also occur
in both directions: false-\textsc{Ready} predictions account for 15.5\% of
all instances, equivalent to accepting 31\% of underspecified inputs, while
false-\textsc{NotReady} predictions account for 17.0\%, equivalent to
rejecting 34\% of codification-ready inputs. On controlled-synthetic
instances, decision accuracy rises to 86.5\%, although most correct
predictions remain weakly grounded.

\paragraph{Localization Failures Primarily Reflect Incorrect Target
Selection.}
Table~\ref{tab:error_decomposition} shows that the annotated blocker is
recovered in only 16.0\% of real-world instances. Of all predictions, 4.3\%
also assign both taxonomy labels correctly, while 11.7\% recover the blocker
but make a taxonomy error. Most failures occur before taxonomy assignment:
17.8\% identify a neighboring implementation decision and 66.3\% identify
a different blocker. Almost no output is vague or lacks a diagnosis,
indicating that the model generally produces a concrete methodological
concern but selects the wrong unresolved decision. Controlled-synthetic
instances show the same broad pattern, with fewer neighboring-blocker errors
but a similarly large proportion of different-blocker predictions.

\paragraph{Clarification Errors Mainly Reflect Insufficient Scope.}
As reported in Table~\ref{tab:error_decomposition}, once the annotated
blocker is supplied, GPT-5.6-Sol reaches 80.6 Macro-CAS on real-world
instances and 96.2 on controlled-synthetic instances. The real-world
insufficient-clarification rate is 19.6\%, compared with only 4.3\%
assumption-imposing actions. The corresponding synthetic rates fall to
2.0\% and 1.0\%. Thus, the main residual difficulty after the blocker is
known is requesting all information required for resolution, rather than
introducing unsupported implementation choices.

Overall, the error profiles locate the main end-to-end difficulty before
clarification generation. Models frequently make weakly grounded readiness
judgments and select plausible but incorrect blockers, whereas clarification
is usually effective once the unresolved decision is explicitly provided.

\begin{table*}[t]
\centering
\footnotesize
\setlength{\tabcolsep}{4.0pt}
\renewcommand{\arraystretch}{1.16}

\begin{tabularx}{\textwidth}{
>{\RaggedRight\arraybackslash}p{0.105\textwidth}
>{\RaggedRight\arraybackslash}p{0.215\textwidth}
>{\RaggedRight\arraybackslash}p{0.185\textwidth}
>{\RaggedRight\arraybackslash}p{0.240\textwidth}
>{\RaggedRight\arraybackslash}X
}
\toprule

\rowcolor{headerbg}
\textbf{Case}
& \textbf{Input Context}
& \textbf{Gold Annotation}
& \textbf{GPT-5.6-Sol Output}
& \textbf{Interpretation}
\\

\midrule

\casecell{caseA}{A}{Missed blocker}
&
\cellcolor{inputbg}
\textbf{Excerpt:}
``\textit{The attention mechanism takes
$[W h_i \Vert W h_j]$ and computes
$e_{ij}=a^\top[W h_i \Vert W h_j]$, followed by softmax
normalization.}''
&
\cellcolor{goldbg}
\textbf{Gold label:}
\textsc{NotReady}

\vspace{0.25em}
\textbf{Blocker:}
The specification omits the LeakyReLU transformation with negative slope
$0.2$ applied to the attention logits before softmax.
&
\cellcolor{modelbg}
\textbf{Prediction:}
\textsc{Ready}

\vspace{0.25em}
\textbf{Rationale:}
The GAT computation, softmax normalization, aggregation, and multi-head
combination are sufficiently specified.
&
\cellcolor{interpbg}
The model treats the linear attention equation as complete, overlooking
a required nonlinearity that changes the resulting attention coefficients.
\\

\midrule

\casecell{caseB}{B}{Neighboring blocker}
&
\cellcolor{inputbg}
\textbf{Excerpt:}
``\textit{The keypoint head unfolds the input image into an
$8\times8$ cell grid and processes the resulting grid with lightweight
convolutions to predict keypoint locations.}''
&
\cellcolor{goldbg}
\textbf{Gold blocker:}
The architecture omits the final $1\times1$ classifier that produces
65 logits for each cell.

\vspace{0.25em}
\textbf{Taxonomy:}
Incompleteness / Missing Model Structure
&
\cellcolor{modelbg}
\textbf{Predicted blocker:}
It is unclear whether ``$8\times8$ cell grid'' denotes
$8\times8$-pixel cells or an $8\times8$ grid over the full image.

\vspace{0.25em}
\textbf{Taxonomy:}
Ambiguity / Ambiguous Definition
&
\cellcolor{interpbg}
The model identifies a plausible issue within the same keypoint-head
module but targets the input-grid convention rather than the omitted
output classifier.
\\

\midrule

\casecell{caseC}{C}{Incomplete request}
&
\cellcolor{inputbg}
\textbf{Defect:}
How historical external variables should be incorporated into TimeMixer:
as channels in \texttt{x\_enc} or as covariates in
\texttt{x\_mark\_enc}.
&
\cellcolor{goldbg}
\textbf{Required information:}
Which input receives the variables and, if \texttt{x\_enc} is used,
whether channel independence must be disabled.
&
\cellcolor{modelbg}
\textbf{Clarification action:}
``\textit{Should historical external variables be provided as additional
channels in \textup{\texttt{x\_enc}} or as time/covariate features in
\textup{\texttt{x\_mark\_enc}}?}''
&
\cellcolor{interpbg}
The action resolves the routing choice but does not request the required
channel-independence setting.
\\

\bottomrule
\end{tabularx}

\caption{
Representative real-world errors from GPT-5.6-Sol.
Case~A shows a false-\textsc{Ready} judgment that overlooks a required
method operation.
Case~B shows a neighboring-blocker error in which the model identifies a
plausible concern within the correct architectural module but not the
annotated blocker.
Case~C shows a relevant but insufficient clarification that omits an
additional condition required for faithful implementation.
Specification excerpts are shortened without altering the information
needed to understand each error.
Taxonomy names are normalized to the final taxonomy in
Table~\ref{tab:final-taxonomy}; the displayed model diagnosis and
clarification action are unchanged.
Full details appear in
Appendix~\ref{app:qualitative_examples}.
}
\label{tab:qualitative_cases}
\end{table*}

\subsection{Qualitative Error Cases}
\label{app:qualitative_examples}

Table~\ref{tab:qualitative_cases} presents three representative
real-world errors from GPT-5.6-Sol, covering one characteristic failure
from each IdeaAMBIG task. We select cases whose gold annotation and model
error can be understood from a short, self-contained excerpt. The displayed
excerpts omit only context unrelated to the target defect and do not add
information from the hidden resolution.

\paragraph{Case A: A Required Operation Is Treated as a Routine Detail.}
The Task~1 example specifies a graph-attention score,
$e_{ij}=a^\top[W h_i \Vert W h_j]$, followed directly by softmax
normalization. The supported method instead applies a LeakyReLU
transformation with negative slope $0.2$ to the attention logits before
softmax. This omission is implementation-critical because applying the
nonlinearity changes the normalized attention coefficients and therefore
the behavior of the attention mechanism.

GPT-5.6-Sol nevertheless predicts \textsc{Ready}, reasoning that the GAT
computation, aggregation, and multi-head combination are already
sufficiently specified. The error is not caused by a completely missing
method description; most of the computation is present. Rather, the model
fails to recognize that one locally omitted operation changes the intended
method and cannot be treated like an unspecified tensor dimension or other
routine engineering choice. This case illustrates how a plausible-looking
specification can pass the readiness gate even when a method-defining
operation remains absent.

\paragraph{Case B: The Correct Module but the Wrong Blocker.}
In the Task~2 example, the specification describes a keypoint head that
unfolds the image into an $8\times8$ cell grid and applies lightweight
convolutions. The annotated blocker is the omission of the final
$1\times1$ classifier that produces 65 logits for each cell. Without this
output layer, the architecture does not determine how the convolutional
features are converted into the required keypoint predictions.

The model instead questions whether the phrase ``$8\times8$ cell grid''
refers to $8\times8$-pixel cells or to an $8\times8$ grid spanning the
full image. This is a plausible concern within the same keypoint-head
module, but it concerns the input-unfolding convention rather than the
missing output classifier. We therefore categorize it as a neighboring
blocker rather than a vague or unrelated diagnosis. The associated taxonomy
error follows from this target mismatch: the normalized predicted category is
\catname{Ambiguous Definition}, whereas the annotated category is
\catname{Missing Model Structure}. Even a correct taxonomy assignment for the model's proposed
concern would not recover the annotated implementation blocker.

\paragraph{Case C: The Clarification Targets the Defect but Does Not Fully
Resolve It.}
The Task~3 example concerns how historical external variables should be
incorporated into TimeMixer. The supported resolution must specify both
whether the variables are passed as channels through
\texttt{x\_enc} or as covariates through \texttt{x\_mark\_enc}, and,
if \texttt{x\_enc} is used, whether channel independence must be disabled.

GPT-5.6-Sol asks whether the variables should be routed through
\texttt{x\_enc} or \texttt{x\_mark\_enc}. The action is relevant and does
not presuppose either choice, but it requests only the routing decision.
A respondent could answer that \texttt{x\_enc} should be used while leaving
the required channel-independence setting unresolved. The action is
therefore insufficient under the counterfactual criterion that answering
only the explicitly requested information must be enough to recover a
codification-ready specification.

\paragraph{Summary.}
The three cases expose failures at different stages of the clarification
pipeline. Case~A fails to recognize that clarification is needed;
Case~B recognizes a plausible issue but localizes the wrong
implementation decision; and Case~C receives the correct blocker but
requests an incomplete resolution. Together with the aggregate analysis in
Table~\ref{tab:error_decomposition}, these examples show that the dominant
difficulty occurs before clarification generation, while the main residual
failure after blocker identification is incomplete clarification scope.

\section{Taxonomy Development and Validation}
\label{supp:taxonomy-development-and-validation}

\subsection{Taxonomy Construction}
\label{supp:taxonomy-construction}

\begin{figure*}[t]
\centering
\includegraphics[width=\textwidth]{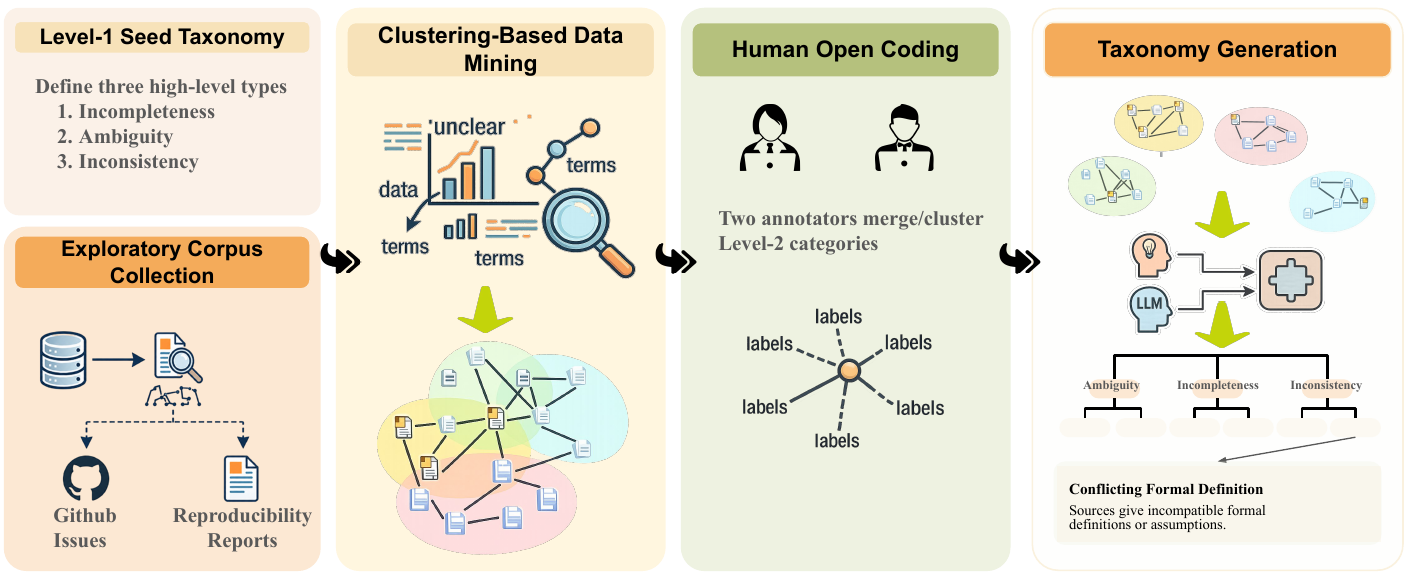}
\caption{
Taxonomy construction pipeline, from a literature-informed Level-1 defect taxonomy
to data-driven discovery and human validation of Level-2 specification defect categories.
}
\label{fig:taxonomy-construction}
\end{figure*}

To make \ourbench{} systematic and reproducible, we construct the taxonomy through a literature-informed and data-driven process rather than relying solely on intuition. Figure~\ref{fig:taxonomy-construction} provides an overview of the construction pipeline. We first define a Level-1 seed taxonomy with \numLevelOneCategories{} broad types of research idea specification defects: \textit{Incompleteness}, \textit{Ambiguity}, and \textit{Inconsistency}. This seed taxonomy is motivated by prior work on requirements quality, ambiguity, and specification problems~\citep{sommerville1997requirements,berry2004ambiguity,zave1997classification}, and is adapted to the setting of scientific research ideas.

To derive the Level-2 categories, we mine an exploratory corpus of resolved specification gaps from MLRC and TMLR reproducibility reports, together with closed or answered issues from paper-associated GitHub repositories. These sources capture cases in which independent reproducers or implementers encountered missing, ambiguous, or conflicting method specifications when reproducing or instantiating published methods. We first apply lexical and heuristic filters to identify candidate specification gaps, and then use LLM-assisted signal mining and clustering to group recurring patterns of implementation-critical defects, such as unclear definitions, missing procedures, absent implementation details, and inconsistencies between sources. The resulting clusters serve as candidate Level-2 defect categories for subsequent human open coding.

The final Level-2 taxonomy is produced through human open coding and iterative consolidation. Two annotators independently review the clustered candidates, assign provisional categories, and merge or split categories through constant comparison across examples. We refine the category names, inclusion criteria, exclusion criteria, and boundary cases into a codebook. The resulting taxonomy organizes Level-2 categories by the affected specification component, including method procedures, model structures, data specifications, evaluation specifications, configuration protocols, and formal definitions, under the corresponding Level-1 defect types. We use this codebook to verify that each category represents a recurring implementation-blocking specification defect rather than an isolated complaint.

We further assess the reliability of the human coding process through inter-annotator agreement before adjudication, as reported in Table~\ref{tab:taxonomy-agreement}. We consider the taxonomy complete for the scope of \ourbench{} when additional sampled instances no longer introduce stable new Level-2 categories and can instead be assigned to existing categories or rejected as out of scope. Therefore, the taxonomy is not intended to be universal across all scientific domains; rather, it provides a coverage-tested taxonomy for AI, NLP, and machine learning research specifications considered in this benchmark.

\begin{table}[t]
\centering
\small
\begin{tabular*}{\columnwidth}
{@{\extracolsep{\fill}}lcc@{}}
\toprule
\textbf{Annotation Target}
& \textbf{Raw Agreement}
& \textbf{Cohen's $\kappa$} \\
\midrule
Keep/reject decision & 88.7\% & 0.76 \\
Level-1 type         & 84.2\% & 0.72 \\
Level-2 category     & 73.5\% & 0.61 \\
Codification slot    & 76.4\% & 0.64 \\
\bottomrule
\end{tabular*}
\caption{Inter-annotator agreement for taxonomy construction before adjudication.}
\label{tab:taxonomy-agreement}
\end{table}

\subsection{Annotation instructions.}
\label{supp:annotation-instructions}
For each candidate instance, annotators first read the paper excerpt, the reported gap, and the supporting evidence, including the evidence text, rebuttal, issue discussion, or code-derived clarification when available. They then determine whether the instance satisfies all keep criteria. A case is marked as \texttt{is\_spec\_gap=Y} if it describes an omission, ambiguity, or conflict in the research idea or method specification. It is marked as \texttt{actionable=Y} if the gap can be addressed through a concrete clarification rather than only through a general critique or subjective preference. It is marked as \texttt{is\_method\_core\_spec\_gap=Y} if the gap affects faithful implementation of the core method, such as the algorithm, model architecture, training procedure, data or preprocessing protocol, objective, or evaluation protocol. Cases involving only environment setup, computational resources, runtime efficiency, formatting, or non-methodological presentation issues are excluded. A case is marked as \texttt{gold\_clarified\_spec\_extractable=Y} if the available evidence provides enough information to write a grounded codification-ready clarification.

After the keep criteria are verified, annotators assign exactly one Level-1 label from \textit{ambiguity}, \textit{incompleteness}, and \textit{inconsistency}, and then assign exactly one Level-2 label under the selected Level-1 type. When a case appears to fit multiple categories, annotators choose the category that best captures the primary implementation blocker. If no existing Level-2 category fits the case, annotators mark \texttt{level2\_label=other} and provide a one-sentence explanation. All disagreements are resolved through adjudication using the codebook, with priority given to evidence support, implementation relevance, and category boundary definitions.

\subsection{Detailed label definitions.}
\label{supp:detailed-label-definitions}

\begin{table*}[t]
  \centering
  \footnotesize
  \setlength{\tabcolsep}{4pt}
  \renewcommand{\arraystretch}{1.15}
  \begin{tabular}{p{0.15\textwidth}p{0.25\textwidth}p{0.52\textwidth}}
    \hline
    \textbf{Level-1 Type} & \textbf{Level-2 Category} & \textbf{Definition} \\
    \hline

    Ambiguity 

    & Ambiguous Definition 

    & A formal element, such as a symbol, notation, mathematical object, or rule, is described without a sufficiently precise meaning. Multiple plausible interpretations remain, leading implementers to compute or instantiate different objects. \\

    & Ambiguous Procedure 

    & A method operation, execution rule, inference behavior, or interaction between components is described but its operational procedure is unclear. Different implementations may follow different behaviors and produce different outcomes. \\

    \hline

    Incompleteness 

    & Missing Method Procedure 

    & A required operational step, algorithmic rule, update mechanism, decision criterion, or execution procedure is omitted. Without this information, an implementer cannot faithfully reproduce how the method operates. \\

    & Missing Model Structure 

    & A model or computational component is mentioned, but its structural configuration is insufficiently specified. Missing details may include layer composition, module organization, dimensional mapping, normalization, activation, or parameterization choices that affect the instantiated model. \\

    & Missing Data Specification

    & The construction or transformation of input data is incompletely described. Missing details may include data filtering, labeling, augmentation, normalization, tokenization, segmentation, or other preprocessing steps that affect the resulting inputs or supervision signals. \\

    & Missing Configuration Protocol 

    & A result-sensitive configuration choice is introduced, but the specification does not describe how the choice should be determined. Missing information concerns the selection, tuning, or validation procedure for important settings (e.g., hyperparameters, thresholds, initialization choices, or sampling parameters), rather than merely an omitted value. \\

    & Missing Evaluation Specification 

    & The evaluation procedure is incompletely described, including missing metric definitions, evaluation protocols, data splits, sampling procedures, prompts, thresholds, or evaluation configurations. Such omissions prevent faithful reproduction or comparison of reported results. \\

    \hline

    Inconsistency 

    & Conflicting Objective 

    & The specification and another source, such as code, appendix, or supplementary material, define different objectives, loss functions, reward signals, or optimization targets. Following different sources would optimize materially different goals. \\

    & Conflicting Model Design 

    & Different sources specify incompatible model components, architectures, preprocessing pipelines, or execution pipelines. The inconsistency makes it unclear which design should be implemented. \\

    & Conflicting Formal Definition 

    & Different sources provide incompatible formal assumptions or mathematical definitions, such as distributions, conditioning rules, aggregation operations, sampling assumptions, or inference formulations. The discrepancy changes the underlying formal model being implemented. \\

    \hline

  \end{tabular}

\caption{
Final Level-2 taxonomy used in \ourbench{}.
Each Level-2 category captures a recurring type of implementation-critical
specification defect in research ideas or method descriptions.
}
  \label{tab:final-taxonomy}
\end{table*}

\subsubsection{Ambiguity}

\paragraph{\catname{Ambiguous Definition}.}
\label{app:cat-ambiguous-definition}

\defn\ A formal element, such as a symbol, notation, mathematical object, or
rule, is introduced without a sufficiently precise definition of its meaning,
scope, or operational interpretation. The ambiguity leaves multiple plausible
interpretations that materially change what an implementer would compute or
instantiate.

\paragraph{\incl.}
This category includes undefined symbols such as $\|\cdot\|$ or $p(v)$;
ambiguous marginalization conventions; unclear variable scope, such as joint
versus marginal probability; inconsistent notation between a theorem and a
proposition; ambiguous operator semantics; unclear rules for constructing
mathematical objects, such as whether an edge feature is defined by subtraction
or concatenation; and unclear domains or codomains of functions.

\paragraph{\excl.}
This category excludes typographic errors where the intended symbol is obvious
from context, such as $L_O$ versus $L_0$. It also excludes cases where the
symbol is defined elsewhere in the same paper and the issue reflects reader
confusion rather than a specification gap.

\paragraph{\bnd.}
If a formal element is defined but the definition is still too vague to
determine what is computed, annotate the case as \catname{Ambiguous Definition}
only when the ambiguity changes the mathematical operation or object. If the
issue concerns the operational behavior or execution procedure of a method
component rather than the meaning of a formal object, use
\catname{Ambiguous Procedure} instead.

\paragraph{\pex.}
A paper defines a score as $\mathrm{Acc}(T)=\|T\|$ but does not specify whether
$\|\cdot\|$ denotes set cardinality, vector length, or another norm. The author
later clarifies that $\|\cdot\|$ denotes the cardinality of a set. This should
be included because different interpretations produce different computed
scores.

\paragraph{\cex.}
A reviewer asks whether $L_O$ and $L_0$ denote the same loss, and the author
replies that $L_0$ was a typo and should be $L_O$. This should be excluded
because the issue is a typographic correction rather than a defect of type
\catname{Ambiguous Definition}.

\paragraph{\realex.}
The notation in Theorem~3 appears inconsistent with Proposition~1, because
$p(v)$ can be read either as a joint distribution over visible and hidden
variables or as a marginal distribution over visible variables. This fits the
category because the two readings lead to different likelihood computations.
The author resolves the ambiguity by clarifying that marginalization over
hidden variables is implicit.

\paragraph{\catname{Ambiguous Procedure}.}
\label{app:cat-ambiguous-procedure}

\defn\ A method operation, execution procedure, inference rule, evaluation
behavior, or interaction between components is described, but its operational
procedure is unclear. Multiple plausible implementations are possible, and
these implementations may lead to materially different behaviors or outcomes.

\paragraph{\incl.}
This category includes ambiguous stage or branch definitions; unclear
parameter-sharing rules across model variants; unclear ordering of agent
updates; ambiguous training schedules, such as episodes versus gradient steps;
unclear whether features are frozen or fine-tuned during evaluation; unclear
true-positive matching rules; ambiguous distinctions between ablation
conditions; unclear prediction targets, such as whether the output is a binary
graph or a weighted graph; vague optimization targets, such as ``improve $X$''
without specifying the optimized quantity; and unclear success criteria without
a metric or threshold.

\paragraph{\excl.}
This category excludes undefined formal elements, which should be annotated as
\catname{Ambiguous Definition}. It excludes cases where the method behavior is
clear but a concrete structural or procedural detail is missing, which should
be annotated as \catname{Missing Method Procedure} or
\catname{Missing Model Structure}. It also excludes cases where two sources
provide conflicting implementations, which should be annotated as
\catname{Conflicting Model Design} or another inconsistency category depending
on the source of conflict.

\paragraph{\bnd.}
If the ambiguity affects both training and evaluation, assign the category
according to the primary implementation decision. If two method variants
cannot be distinguished because the formal objects are unclear, use
\catname{Ambiguous Definition}. Otherwise, use
\catname{Ambiguous Procedure} when the uncertainty concerns how the method
operates. If the target behavior is clear but a required execution step is
missing, use \catname{Missing Method Procedure}.

\paragraph{\pex.}
A reviewer asks whether a transfer experiment trains linear classifiers on
fixed representations or fine-tunes the backbone. The author clarifies that
features before the final classification layer are fixed and a new linear SVM
is trained for each target dataset. This should be included because
fixed-feature transfer and fine-tuning define different evaluation procedures.

\paragraph{\cex.}
A reviewer asks how many attention heads are used, and the author points to
Appendix~C, where the model uses 8 heads. This should be excluded because the
issue concerns a missing architectural parameter rather than ambiguous
procedure. If the value is absent and prevents reconstruction of the model
structure, use \catname{Missing Model Structure}.

\paragraph{\realex.}
The paper does not explain how the order of agent Q-function updates is chosen,
nor how agents agree on this order in a decentralized setting. This fits the
category because resampling the order in each round and fixing one order
throughout training imply different communication and convergence behavior.
The author clarifies that a single random permutation is sampled once and kept
fixed throughout training.

\subsubsection{Incompleteness}

\paragraph{\catname{Missing Method Procedure}.}
\label{app:cat-missing-method-procedure}

\defn\ A required operational step, algorithmic rule, update mechanism,
decision criterion, module interaction rule, or execution procedure is omitted.
Without this information, an implementer cannot faithfully reproduce how the
method operates or may implement materially different procedures.

\paragraph{\incl.}
This category includes missing decision rules or bookkeeping sets in search
algorithms; missing termination conditions; missing routing rules for which
representation feeds which loss; missing special-token construction and
training procedures; missing detokenization rules; missing weight-sharing
rules between training stages; missing kernel, stride, or padding rules for
length-preserving convolution; missing quantization rules or gradient
estimators; and missing execution rules that determine how intermediate
representations or decisions are produced.

\paragraph{\excl.}
This category excludes missing configuration selection procedures, which should
be annotated as \catname{Missing Configuration Protocol}. It excludes missing
structural choices, such as pooling functions, normalization layers, module
organization, or dimensional mappings, which should be annotated as
\catname{Missing Model Structure}. If both a procedure and a configuration
choice are absent, use this category when the missing execution rule is the
primary blocker.

\paragraph{\bnd.}
If the paper states that a procedure is used but does not describe how the
procedure is executed, annotate the case as
\catname{Missing Method Procedure}. If the procedure is sufficiently specified
but the missing information concerns how a tunable choice is selected,
validated, or adapted, use \catname{Missing Configuration Protocol}.

\paragraph{\pex.}
A paper introduces two representations, $\mathbf{u}_L(x_t)$ and
$\mathbf{u}_L(x_t,y_\tau)$, but does not specify which representation is used
for MLM, NSP, or SOP. This should be included because the loss computation and
gradient flow cannot be implemented faithfully without this routing rule.

\paragraph{\cex.}
A reviewer asks for the value of $k$ in a Top-$k$ compression module, and the
author replies that $k=0.1$. This should be excluded because the issue concerns
a configuration value rather than a missing execution procedure. If the paper
omits how $k$ is selected or tuned, use
\catname{Missing Configuration Protocol}.

\paragraph{\realex.}
The paper introduces unary and binary output representations but does not state
which representation feeds MLM, NSP, or SOP, or whether binary predicates are
supervised. This fits the category because the training objective cannot be
implemented faithfully without these routing decisions. The rebuttal resolves
the issue by specifying the use of per-token $\mathbf{u}_L(x_t)$ for MLM,
\texttt{[CLS]} $\mathbf{u}_L(x)$ for NSP/SOP, and no direct loss on
$\mathbf{u}_L(x,y)$.

\paragraph{\catname{Missing Model Structure}.}
\label{app:cat-missing-model-structure}

\defn\ The paper states that a model, computational component, or architectural
module is used, but omits structural details that materially affect the
implemented model. These details may concern component organization, layer
composition, dimensional mapping, normalization, activation functions,
initialization, or other structural choices.

\paragraph{\incl.}
This category includes missing normalization layers; missing graph-level
readout functions; missing pooling composition rules; missing activation
functions when non-standard; missing L2 normalization between layers; missing
weight initialization schemes when they differ from defaults; missing fixation
point generation structures; missing frame selection strategies; and missing
dimensional mappings needed to connect model components.

\paragraph{\excl.}
This category excludes missing configuration selection procedures for otherwise
defined architectures, which should be annotated as
\catname{Missing Configuration Protocol}. It also excludes missing execution
rules for algorithm behavior, which should be annotated as
\catname{Missing Method Procedure}. If the paper and another source provide
contradictory architectures or pipelines, use
\catname{Conflicting Model Design}.

\paragraph{\bnd.}
If the missing detail is a structural choice that changes model capacity,
information flow, or representation shape, annotate the case as
\catname{Missing Model Structure}. If the missing detail is only a numerical
setting within an otherwise complete architecture, use
\catname{Missing Configuration Protocol} only when the selection procedure is
missing and result-sensitive.

\paragraph{\pex.}
A paper states that a graph representation is converted into a graph-level
vector but does not specify whether the readout is mean pooling, max pooling,
sum pooling, or a concatenation of multiple pooling functions. This should be
included because the readout function changes the representation passed to
downstream layers.

\paragraph{\cex.}
A reviewer asks why U-Net was chosen over other architectures, and the author
replies that it performed best empirically. This should be excluded because
the question asks for a design justification rather than a missing structural
specification.

\paragraph{\realex.}
The paper mentions ``simulating the pupil function of an ideal imaging system''
but does not specify how fixation locations are generated, how many fixations
are used, or how the mask geometry is defined. This fits the category because
the method depends on the structure of the fixation sampling module. The
rebuttal answer specifies a $2{\times}2$ grid, four cell centroids plus the
global centroid, and a rectangular fixation mask with the same aspect ratio as
the input and spatial size equal to 25\% of the image.

\paragraph{\catname{Missing Data Specification}.}
\label{app:cat-missing-data-specification}

\defn\ The paper does not fully specify how data are constructed, filtered,
labeled, normalized, tokenized, segmented, augmented, or transformed before
being used by the method. As a result, different implementers may construct
different inputs, labels, or supervision signals.

\paragraph{\incl.}
This category includes missing dataset filtering rules; missing inclusion or
exclusion criteria; missing label construction rules; missing normalization
or standardization procedures; missing tokenization or serialization
templates; missing segmentation rules for long documents, videos, or time
series; missing train/validation/test split construction when it is part of
data preparation; missing augmentation procedures; missing negative sampling
rules; missing prompt construction rules for data generation; and missing
procedures for removing duplicates or leakage.

\paragraph{\excl.}
This category excludes missing evaluation-only details, which should be
annotated as \catname{Missing Evaluation Specification}. It excludes missing
model structural details that transform representations inside the model,
which should be annotated as \catname{Missing Model Structure}. It also
excludes cases where two sources provide concrete but conflicting data
pipelines, which should be annotated as \catname{Conflicting Model Design}.

\paragraph{\bnd.}
If the gap affects the construction of examples, inputs, labels, or
supervision signals before training or evaluation, use
\catname{Missing Data Specification}. If the gap affects how predictions are
scored after the model produces outputs, use
\catname{Missing Evaluation Specification}. If the paper omits a preprocessing
detail but the code provides one valid implementation without contradiction,
annotate the case as incompleteness rather than inconsistency.

\paragraph{\pex.}
A paper states that long documents are split into chunks for retrieval but
does not specify the chunk length, overlap, section filtering rule, or whether
tables and captions are retained. This should be included because different
preprocessing choices create different retrieval inputs.

\paragraph{\cex.}
A paper uses ImageNet normalization and cites a standard preprocessing
pipeline. A reviewer asks whether pixel values are scaled to $[0,1]$. This
should be excluded when the cited pipeline already determines the
preprocessing rule and the issue reflects clarification rather than an
implementation-blocking gap.

\paragraph{\realex.}
A paper-associated GitHub issue reports that a released method requires
converting paper PDFs into text chunks, but the paper does not specify whether
the conversion uses raw PDF text, LaTeX source, OCR output, or Markdown
extraction, nor how figures, equations, and references are handled. This
fits the category because different preprocessing pipelines produce different
textual inputs and retrieval contexts.

\paragraph{\catname{Missing Configuration Protocol}.}
\label{app:cat-missing-configuration-protocol}

\defn\ A result-sensitive configuration choice is introduced, but the
specification does not describe how the choice should be determined. The
missing information concerns the selection, tuning, validation, or adaptation
procedure for important settings rather than merely an omitted numerical value.

\paragraph{\incl.}
This category includes missing search ranges; missing validation metrics;
missing validation data fractions; missing stopping criteria for configuration
selection; missing procedures for selecting parameters that depend on unknown
quantities; missing prior variance or posterior summary selection procedures
in Bayesian methods; and missing selection rules for settings such as
$\gamma$, $\lambda$, temperature, thresholds, regularization strengths, or
sampling parameters when these choices materially affect results.

\paragraph{\excl.}
This category excludes ordinary unreported values when the selection procedure
is standard and unambiguous. It also excludes structural model choices whose
absence prevents construction of a component, which should be annotated as
\catname{Missing Model Structure}. Missing execution procedures that happen to
contain tunable parameters should not be annotated here if the procedure itself
is the primary missing element.

\paragraph{\bnd.}
If the paper states that a configuration choice is tuned but omits the
selection criterion, search range, candidate set, or validation procedure,
annotate the case as \catname{Missing Configuration Protocol}. If the tuning
procedure is clear but the final selected value is not reported, do not
annotate the case as a specification gap unless the missing value prevents
reproduction in a non-standard way.

\paragraph{\pex.}
A paper introduces a temperature parameter for decoding and reports that it
strongly affects results, but does not specify whether the value is fixed,
tuned on a validation set, or selected per task. This should be included
because different selection protocols can produce materially different
outputs.

\paragraph{\cex.}
A paper omits the batch size used in training, and the author later reports
that the batch size is 256. This should be excluded when batch size is a
standard training parameter and the configuration selection procedure is not
part of the method specification.

\paragraph{\realex.}
The algorithm requires choosing $\lambda$ according to a covering number
$\mathcal{N}(\mathcal{H},\varepsilon)$ that depends on the misspecification
level $\xi$, the switch bound $S$, and the path-length bound $P$, all of which
are unknown in practice. This fits the category because the paper does not
provide a usable protocol for setting $\lambda$. The author answer resolves
the issue by recommending grid search in practical use.

\paragraph{\catname{Missing Evaluation Specification}.}
\label{app:cat-missing-evaluation-specification}

\defn\ The evaluation procedure is incompletely specified, leaving out details
required to reproduce reported results or compare systems fairly. Missing
details may affect how performance is measured, aggregated, filtered, or
interpreted.

\paragraph{\incl.}
This category includes missing metric computation rules; missing thresholds
for converting scores into labels; missing evaluation data splits; missing
prompt sets or templates used for evaluation; missing decoding settings;
missing sampling procedures; missing judge model configurations; missing
aggregation rules across seeds, tasks, or datasets; missing tie-breaking rules;
and missing matching rules for comparing predictions with references.

\paragraph{\excl.}
This category excludes missing training procedures, which should be annotated
as \catname{Missing Method Procedure}. It excludes missing data construction
or preprocessing rules, which should be annotated as
\catname{Missing Data Specification}. It also excludes cases where different
sources provide conflicting evaluation pipelines, which should be annotated as
\catname{Conflicting Model Design}.

\paragraph{\bnd.}
If the missing detail determines how reported performance is computed,
aggregated, or interpreted, use \catname{Missing Evaluation Specification}.
If the missing detail determines how evaluation inputs or examples are
constructed, use \catname{Missing Data Specification}. If the evaluation
procedure is explicitly specified but another source provides a conflicting
procedure, use \catname{Conflicting Model Design}.

\paragraph{\pex.}
A paper reports F1 for extracted claims but does not specify whether matching
is exact, token-level, span-level, or semantic, and does not state how partial
matches are scored. This should be included because different matching rules
can produce different F1 values.

\paragraph{\cex.}
A reviewer asks why AUROC rather than accuracy is reported, and the author
explains that the dataset is imbalanced. This should be excluded because the
metric is already specified and the question concerns justification rather
than a missing evaluation specification.

\paragraph{\realex.}
A reproducibility report for an LLM evaluation study finds that the paper
reports win rate under model-based judging but does not specify the judge
prompt, judge model version, decoding temperature, or whether pair order is
randomized. This fits the category because these choices directly affect the
evaluation outcome and must be specified for faithful reproduction.

\subsection{Inconsistency}
\label{supp:inconsistency}

\paragraph{\catname{Conflicting Objective}.}
\label{app:cat-conflicting-objective}

\defn\ The specification and another source, such as released code, an
appendix, supplementary material, or configuration file, define incompatible
objectives, loss functions, reward signals, or optimization targets. Following
the different sources would optimize materially different behaviors or
solutions.

\paragraph{\incl.}
This category includes conflicts in the mathematical form of a loss function;
incompatible weighting or aggregation of loss terms; discrepancies between
the stated and implemented optimization targets; conflicts in reward
definitions for reinforcement learning; incompatible regularization terms;
differences in the signs, coefficients, or normalization of objective
components; and cases where the implementation optimizes a surrogate that
changes the intended methodological behavior.

\paragraph{\excl.}
This category excludes cases where an objective or loss component is omitted
rather than contradicted, which should be annotated as
\catname{Missing Method Procedure}. It excludes unclear descriptions that
permit multiple interpretations without another source explicitly defining an
incompatible objective, which should be annotated as
\catname{Ambiguous Definition}. It also excludes conflicts concerning
model components, architecture, preprocessing, or execution order, which
should be annotated as \catname{Conflicting Model Design}. Conflicts in formal
assumptions or mathematical definitions that do not directly change the
optimized quantity should be annotated as
\catname{Conflicting Formal Definition}.

\paragraph{\bnd.}
If two sources define different quantities to be minimized or maximized, or
apply incompatible weighting, aggregation, signs, or coefficients that
materially change the optimization behavior, use
\catname{Conflicting Objective}. If the objective is consistent but the
sources define different model architectures or processing pipelines, use
\catname{Conflicting Model Design}. If the discrepancy concerns an underlying
mathematical assumption, distribution, sampling rule, or inference
formulation rather than the optimized quantity itself, use
\catname{Conflicting Formal Definition}. If the objective is not specified at
all, use \catname{Missing Method Procedure}.

\paragraph{\pex.}
A paper defines the training loss as a weighted element-wise sum of per-sample
log-probabilities, whereas the released implementation multiplies the weight
vector and log-probability vector using an outer product before aggregation.
This should be included because the two formulations optimize different
quantities and induce materially different training behavior.

\paragraph{\cex.}
A paper defines a classification loss but does not report the coefficient of
an auxiliary regularization term. This should be excluded because no
incompatible objective is specified; the coefficient is missing and should
instead be annotated as \catname{Missing Method Procedure}.

\paragraph{\realex.}
A reproducibility report for a fair offline reinforcement-learning method
finds that the stated objective requires element-wise weighting of
log-probabilities by importance weights, whereas the released implementation
computes an outer product between the two vectors. This fits
\catname{Conflicting Objective} because the two formulations define
incompatible loss computations: the implemented objective effectively reduces
the method to behavior cloning rather than optimizing the intended
fairness-weighted objective.

\paragraph{\catname{Conflicting Model Design}.}
\label{app:cat-conflicting-model-design}

\defn\ Different sources specify incompatible model components,
architectures, preprocessing pipelines, or execution pipelines, leaving it
unclear which design represents the intended method. Following the different
sources would produce materially different computational structures or
processing behaviors.

\paragraph{\incl.}
This category includes conflicts in the number, type, order, or connectivity
of model components; incompatible descriptions of feature aggregation,
residual connections, attention mechanisms, normalization layers, or
prediction heads; conflicts between single-stage and multi-stage pipelines;
incompatible preprocessing or postprocessing pipelines; and discrepancies
between the architecture described in a paper and the architecture implemented
in released code.

\paragraph{\excl.}
This category excludes cases where an architectural detail is absent rather
than contradicted, which should be annotated as
\catname{Missing Model Structure}. It excludes cases where a single
description permits multiple architectural interpretations without explicitly
specifying incompatible alternatives, which should be annotated as
\catname{Ambiguous Procedure}. It also excludes conflicts in loss
functions, reward signals, or optimization targets, which should be annotated
as \catname{Conflicting Objective}, and conflicts in mathematical assumptions
or formal operations, which should be annotated as
\catname{Conflicting Formal Definition}.

\paragraph{\bnd.}
If two sources explicitly define different model components, architectural
connections, preprocessing stages, or execution pipelines, use
\catname{Conflicting Model Design}. If the relevant design choice is simply
omitted, use \catname{Missing Model Structure}. If the specification uses
unclear language that supports multiple interpretations but no source commits
to an incompatible alternative, use \catname{Ambiguous Procedure}. If
the architecture is consistent but the sources optimize different objectives,
use \catname{Conflicting Objective}.

\paragraph{\pex.}
A paper states that representations from the final encoder layer are passed to
the prediction head, whereas the accompanying implementation concatenates
representations from all encoder layers before prediction. This should be
included because the two sources define incompatible feature aggregation and
prediction architectures.

\paragraph{\cex.}
A paper states that a graph neural network is used but does not report the
number of message-passing layers. This should be excluded because no
incompatible architecture is specified; the architectural detail is missing
and should instead be annotated as \catname{Missing Model Structure}.

\paragraph{\realex.}
A reproducibility report for a graph neural network explainer finds that the
paper describes three GCN layers feeding directly into a classifier, whereas
the released code concatenates the three intermediate GCN outputs before
classification. This fits \catname{Conflicting Model Design} because the two
sources define incompatible feature aggregation and classification
architectures, leading to materially different implementations.

\paragraph{\catname{Conflicting Formal Definition}.}
\label{app:cat-conflicting-formal-definition}

\defn\ Different sources provide incompatible formal assumptions,
mathematical definitions, or derivations for the same part of a method.
These conflicts may concern distributions, conditioning rules, aggregation
operations, sampling assumptions, normalization procedures, inference
formulations, or other mathematical operations. Following the different
definitions would implement materially different formal models or produce
different outputs.

\paragraph{\incl.}
This category includes incompatible equations for the same quantity;
conflicts in element-wise, vector, or matrix operations; discrepancies in
summation, averaging, normalization, or aggregation rules; incompatible
probability distributions or conditioning assumptions; conflicts in sampling
and inference formulations; inconsistencies between a stated equation and its
derivation; and cases where a simplified formula is not mathematically
equivalent to the formal definition used elsewhere.

\paragraph{\excl.}
This category excludes cases where a formal definition is missing rather than
contradicted, which should be annotated as
\catname{Missing Method Procedure}. It excludes cases where a formula is
present but its notation or intended interpretation is unclear without another
source specifying an incompatible definition, which should be annotated as
\catname{Ambiguous Definition}. It also excludes conflicts that directly
define different optimization objectives or loss functions, which should be
annotated as \catname{Conflicting Objective}, and conflicts in model
components, architectures, preprocessing pipelines, or execution pipelines,
which should be annotated as \catname{Conflicting Model Design}.

\paragraph{\bnd.}
If two sources explicitly provide mathematically incompatible definitions,
assumptions, derivations, or operations for the same method component, use
\catname{Conflicting Formal Definition}. If the conflicting formulas define
different quantities to be optimized, use \catname{Conflicting Objective}.
If the formulas are consistent but their implementation occurs within
different model architectures or processing pipelines, use
\catname{Conflicting Model Design}. If the required mathematical definition
is absent, use \catname{Missing Method Procedure}; if a single definition
admits multiple interpretations because of unclear notation, use
\catname{Ambiguous Definition}.

\paragraph{\pex.}
A paper defines a covariance contribution using an element-wise sum of
variance ratios, whereas an appendix replaces the same term with a ratio of
products. This should be included when the two expressions are not
mathematically equivalent and therefore produce different values for the same
model quantity.

\paragraph{\cex.}
A paper defines a probability score using a summation but does not specify
whether the index ranges over documents, tokens, or latent variables. This
should be excluded because no incompatible alternative is explicitly given;
the notation is ambiguous and should instead be annotated as
\catname{Ambiguous Definition}.

\paragraph{\realex.}
A reproducibility report for a probabilistic retrieval model finds that the
paper simplifies a trace term in the KL divergence as a ratio of products,
whereas the corrected derivation computes an element-wise sum of variance
ratios. This fits \catname{Conflicting Formal Definition} because the two
expressions are not mathematically equivalent and therefore produce different
relevance scores under the same probabilistic model.

\section{Construction of Codification-Ready Counterparts}
\label{app:ready_counterparts}

Each retained benchmark record contains an underspecified
\textsc{NotReady} specification and a corresponding
codification-ready \textsc{Ready} specification. The counterpart is
constructed by resolving the annotated target defect using only the
available source evidence. Reviewers verify that the resulting
specification resolves the implementation-critical decision, introduces
no unsupported methodological assumptions, and preserves all
non-target content.

For real-world instances derived from reproducibility reports, the
resolution is obtained from the implementation choice or clarification
documented by the reproducing authors and, when available, corroborated
by the associated paper or implementation. For GitHub-derived
instances, the resolution is obtained from the author or maintainer
response that closes or resolves the issue. For controlled synthetic
instances, the codification-ready source reference before defect
injection serves directly as the \textsc{Ready} counterpart.

The two versions are written in the same specification format and retain
the same task, inputs, outputs, method context, training context, and
evaluation context except where the target clarification necessarily
modifies one of these components. This paired construction limits
surface-form differences unrelated to codification readiness.

\section{Additional Experimental Analysis}
\label{app:additional_analysis}

\subsection{Inference and Deployment Details}
\label{app:inference}

\paragraph{Model Access and Deployment.}
We evaluate the 13 models reported in
Table~\ref{tab:main_results}, accessing all models through the OpenRouter API.
The frontier models are
GPT-5.6-Sol, Claude Sonnet 5, Gemini 3.1 Pro Preview, and
DeepSeek-V3.2.
The open-weight reasoning models are Qwen3.5-397B-A17B,
DeepSeek-R1-0528, GLM-5.2, and Kimi-K3. The open-weight general
models are GPT-OSS-120B, Gemma-4-31B-IT, Qwen3.5-9B, Qwen3-8B,
and Qwen3-32B. We use the same benchmark inputs,
task definitions, and output requirements across model families.

\paragraph{Shared Prompting Protocol.}
All models are evaluated using the fixed task-specific prompts provided
in Appendix~\ref{app:prompts}. Each benchmark record is processed
independently, without access to predictions for its paired specification
or to examples from other benchmark records. Models receive only the
information defined by the corresponding task. In particular, they never
receive the downstream papers, codebases, issue threads, reproducibility
evidence, or source artifacts used to construct or validate the instance.

For Task~1, the model receives a single specification and predicts a
\textsc{Ready} or \textsc{NotReady} label together with a supporting
rationale. For Task~2, it receives an underspecified specification and
returns a Level-1 label, a Level-2 label, and a natural-language
description of the unresolved implementation decision. For Task~3, it
receives the underspecified specification together with the annotated
target-defect description and returns an action type, a concrete
clarification action, and the information expected from carrying out
that action.

\paragraph{Decoding and Structured Outputs.}
We use deterministic decoding whenever it is supported by the model
provider, setting the temperature to zero. For tasks requiring structured
generation, we request JSON Schema-constrained outputs whenever this
functionality is available. The requested schema follows the output
fields defined for each task and is held fixed across models.

We set the maximum output length to 2,048 tokens for Task~1 readiness
assessment and Task~2 defect localization, and to 1,024 tokens for
Task~3 clarification-action generation. These limits provide sufficient
space for the required structured fields and their accompanying
natural-language descriptions while keeping the inference protocol
consistent across models.

\paragraph{Evaluator Separation.}
Model generation and output evaluation are performed as separate stages.
The evaluated models do not receive evaluator judgments, gold resolutions,
or feedback during generation. Semantic and rubric-based components are
scored using the shared evaluator described in
Appendix~\ref{app:judge_validation}. The same evaluation criteria are
applied to outputs from all model families.

\subsection{Information Bottleneck Analysis for Clarification}
\label{app:information_bottleneck}

Our main results show a substantial gap between defect localization
(Task~2) and clarification action generation (Task~3). However, these tasks
differ in both inputs and outputs, making it unclear whether the observed
gap reflects difficulty in discovering the unresolved specification defect
or difficulty in formulating clarification actions. To isolate whether defect discovery or clarification formulation is the
primary bottleneck, we perform an information bottleneck analysis by
controlling the availability of defect information during clarification
generation.

\paragraph{Experimental Setup.}
We fix GPT-5.6-Sol as the backbone model and evaluate clarification action
generation under two input conditions. The \textsc{End-to-End} setting
receives only the underspecified research idea and must independently
identify the missing specification information before generating a
clarification action. The \textsc{Defect-Guided} setting additionally
receives the annotated target defect while keeping the original
specification unchanged. Both settings are evaluated on the same Task~3 instances and use
identical prompts and decoding configurations.

We evaluate clarification quality using the Task~3 metrics: Macro-CAS,
Sufficiency, and No-Assumption. Macro-CAS measures whether the generated action targets the
annotated defect, requests sufficient information for resolution,
and avoids unsupported assumptions. Sufficiency
measures whether the requested information would be sufficient to resolve
the gap. No-Assumption measures whether the model introduces unsupported
implementation choices.

\paragraph{Results.}
As shown in Table~\ref{tab:information_bottleneck}, providing the target
defect substantially improves clarification quality. Macro-CAS increases
from 13.6 to 80.6, a gain of 67.0 points, while Sufficiency increases from
8.6 to 80.4, a gain of 71.8 points. These improvements show that models can
generate effective clarification actions once the unresolved implementation
decision is explicitly identified, but struggle when they must first discover
the relevant specification gap from the input alone.

No-Assumption remains nearly unchanged across the two settings
(96.3 vs.\ 95.7), indicating that the performance gap is not primarily caused
by models introducing unsupported implementation choices. Instead, without
access to the target defect, models tend to produce cautious but insufficient
or misdirected clarification requests.

Overall, these results support our central finding that defect discovery,
rather than clarification formulation, is the primary bottleneck. Current
LLMs can effectively request the missing information once directed to the
correct implementation-critical gap, but remain substantially less reliable
at identifying which unresolved decision requires clarification.

\subsection{Taxonomy-Free Defect Localization}
\label{app:taxonomy_free_localization}

Task~2 requires a model to identify an implementation-critical defect
and predict its Level-1 and Level-2 taxonomy labels. To separate defect
localization from taxonomy prediction, we evaluate GPT-5.6-Sol without
requiring defect-category outputs on all 163 real-world instances.

\paragraph{Experimental Setting.}
The taxonomy-free setting uses the same underspecified specifications
as the original Task~2 evaluation. The model predicts only a
natural-language defect description, without Level-1 or Level-2 labels:
\[
\hat e_i=f(x_i).
\]
The prediction should identify the unresolved implementation decision
and explain why it affects faithful implementation. We compare these
outputs with the original Task~2 predictions on the same 163 instances.

\paragraph{Evaluation Protocol.}
Both conditions use the same Claude Opus~4.8 semantic matcher and
same-blocker criterion used for Loc-Acc
(Appendices~\ref{app:task2_metrics} and~\ref{app:judge_validation}).
A prediction is correct when its defect description identifies the
same atomic, implementation-critical decision as the annotated defect,
rather than a neighboring decision or a general methodological concern.
Blocker accuracy is computed over the full real-world set:
\[
\mathrm{Blocker\ Acc.}
=
\frac{100}{N}\sum_{i=1}^{N}m_i,
\qquad N=163,
\]
where $m_i=1$ if the predicted description matches the annotated defect
and $m_i=0$ otherwise.

Neither condition requires correct taxonomy labels for blocker scoring.
For the original Task~2 outputs, this measure is Loc-Acc: it scores
only defect identification even though the model also predicts taxonomy
labels. The taxonomy-free condition removes category prediction from
the requested output while retaining the same localization criterion.
Macro DRR is a separate metric that additionally requires correct
Level-1 and Level-2 labels and macro-averages recovery over Level-2
categories.

\paragraph{Results.}
We compare the two output requirements.
Removing category prediction raises blocker accuracy from 16.0\%
to 40.5\%. Taxonomy prediction therefore adds substantial difficulty,
while defect discovery remains challenging without it: 59.5\% of the
annotated defects are not recovered in the taxonomy-free setting.

\subsection{Inference-Time Search for Defect Localization}
\label{app:inference_time_localization}

Since defect localization is the main bottleneck in \ourbench{}, we test whether
simple inference-time search can improve Task~2 performance without changing
the model or providing additional evidence.

\paragraph{Setup.}
We evaluate GPT-5.6-Sol on all 163 real-world Task~2 instances using the same
benchmark inputs, taxonomy, localization criteria, evaluator, and deterministic
decoding configuration as in the main experiments. The original Task~2 setting
requires a single blocker prediction. We additionally make
$K\in\{3,5\}$ independent calls per instance and pool the candidate blockers,
while keeping the same definition of an implementation-critical defect.

Each candidate is matched to the annotated blocker using the same taxonomy-blind
target-identity evaluator used for Loc-Acc. We define blocker Recall@$K$ as
\begin{equation}
\mathrm{Recall@}K =
\frac{1}{N}\sum_{i=1}^{N}
\mathbb{I}\!\left[
\exists j\leq K:\hat d_{ij}\equiv d_i^\star
\right],
\end{equation}
where $d_i^\star$ is the annotated blocker and $\hat d_{ij}$ is the $j$-th
generated candidate. Recall@$K$ is an oracle coverage metric and does not
measure deployable top-1 localization.

We also evaluate a gold-blind self-selection condition for $K=5$. The model
receives only the original specification and its five generated candidates and
selects one candidate as the primary blocker. No gold defect, taxonomy label,
source artifact, or resolution is provided. The selected candidate is evaluated
with the original Loc-Acc and Macro DRR metrics.

\paragraph{Results.}
As shown in Table~\ref{tab:inference_scaling}, expanding the candidate set
provides only modest gains in blocker coverage. Recall increases from 16.0\%
under single-shot inference to 21.5\% at $K=3$ and 23.3\% at $K=5$. Thus, even
with five candidates, the annotated blocker is absent from the candidate set in
76.7\% of real-world instances.

The additional coverage also does not translate into better top-1 localization.
Gold-blind self-selection over the $K=5$ candidates yields 8.0\% Loc-Acc and
3.0\% Macro DRR, compared with 16.0\% and 9.6\% under single-shot inference.
Among instances where the annotated blocker appears in the top-5 candidate set,
gold-blind self-selection recovers it in only $8.0/23.3 \approx 34.3\%$ of
cases.

\paragraph{Interpretation.}
These results suggest that the localization bottleneck is not primarily caused
by premature commitment to a single prediction. Broader candidate generation
only modestly improves discovery of the annotated blocker, and the model further
struggles to distinguish the correct blocker from plausible alternatives once
it is generated. Simple inference-time search therefore does not substantially
alleviate the Task~2 localization bottleneck.

\subsection{Executing Evidence-Seeking Actions over Research Artifacts}
\label{app:artifact_execution}

The main Task~3 evaluation assesses clarification actions counterfactually:
would the information requested by an action resolve the annotated blocker if
it were obtained? To test the additional execution step, we conduct a small
diagnostic study in which a model generates an evidence-seeking action and an
executor carries it out over a fixed research artifact.

\paragraph{Setup.}
We select $N=20$ controlled instances from 20 distinct source papers. Each
has a bounded implementation or evaluation detail that is recoverable from
the locally available paper; selection is based on artifact availability and
resolvability, not model predictions. For each instance, GPT-5.6-Sol receives
only the underspecified specification and annotated target defect, as in the
defect-guided Task~3 setting. A prompt adapted from Task~3 requires it to
produce one concrete \textsc{Evidence-Seeking} action specifying what to
inspect and what unknown detail to determine, without supplying an answer.

A GPT-5.6-Sol executor then receives the specification, target defect,
generated action, and a fixed bundle containing only the associated paper.
It can list the bundle, search its text, and open bounded line ranges, with
a limit of 10 tool calls. It returns a resolved detail or abstains, together
with cited artifact locations. Cited passages are checked against lines the
executor actually inspected before evidence is judged. Neither stage receives
the gold resolution, resolved specification, or evaluator feedback; the gold
resolution enters only the post-execution recovery evaluation.

\paragraph{Evaluation.}
For instance $i$, let $R_i=1$ if a semantic judge determines that the
executor's answer fully recovers the implementation-critical gold detail, and
let $E_i=1$ if a separate, gold-blind judge determines that the verified
artifact passage supports the executor's answer. An abstention receives zero
on both criteria. The three reported metrics are
\begin{equation}
\begin{aligned}
\mathrm{Recovery} &= \frac{1}{N}\sum_{i=1}^{N} R_i, \\
\mathrm{Support}  &= \frac{1}{N}\sum_{i=1}^{N} E_i, \\
\mathrm{E2E}      &= \frac{1}{N}\sum_{i=1}^{N} R_iE_i.
\end{aligned}
\end{equation}
Both judgments use GPT-5.4. Recovery is semantic rather than lexical and
requires the complete target detail; Support concerns the executor's own
answer and does not use the gold resolution.

\begin{table}[H]
\centering
\footnotesize
\setlength{\tabcolsep}{2pt}
\renewcommand{\arraystretch}{1.05}

\begin{tabularx}{\columnwidth}
{@{}>{\raggedright\arraybackslash}m{0.32\columnwidth}
*{3}{>{\centering\arraybackslash}X}@{}}

\toprule
\textbf{Setting}
& \textbf{\shortstack{Resolution\\Recovery $\uparrow$}}
& \textbf{\shortstack{Evidence\\Support $\uparrow$}}
& \textbf{\shortstack{End-to-End\\Success $\uparrow$}} \\
\midrule
Generated Action $\rightarrow$ Artifact Execution
& 45.0 & 40.0 & 25.0 \\
\bottomrule
\end{tabularx}

\caption{Execution of evidence-seeking actions over fixed paper artifacts
for $N=20$ artifact-resolvable Task~3 instances. Resolution Recovery measures
whether the target detail is recovered; Evidence Support requires the cited
artifact passage to substantiate the executor's answer; End-to-End Success
requires both. Values are percentages, corresponding to 9/20, 8/20, and
5/20 instances, respectively.}
\label{tab:artifact_execution}
\end{table}

\paragraph{Results and Scope.}
Table~\ref{tab:artifact_execution} shows that the executor recovers the gold
detail in 9/20 cases and provides sufficient artifact support for its answer
in 8/20; both conditions hold in 5/20. Among the remaining cases, four
recover the detail without sufficient cited support, three provide evidence
for an answer that does not fully recover the gold detail, and eight satisfy
neither criterion. Thus, even when the target blocker is supplied and the
paper contains the answer, generating an inspection action does not ensure
an evidence-supported resolution. This study probes the artifact-acquisition
step preceding the oracle-resolved downstream codification studies; it does
not measure end-to-end performance on the full benchmark. Because the sample
contains only 20 selected, paper-resolvable controlled instances, the rates
should not be generalized to all \ourbench{} instances or other artifact types.

\subsection{Small-Scale Executable Validation}
\label{app:executable_validation}

The oracle clarification utility study in Section~\ref{sec:oracle-clarification-utility}
evaluates whether resolving a specification gap improves the quality of the
resulting implementation specification. However, codification-readiness is still
an intermediate outcome: a natural question is whether these improvements carry
over to executable code. To address this question without turning IdeaAMBIG into
a full paper-reproduction benchmark, we conduct a small-scale executable
validation study.

\paragraph{Setup.}
We sample 20 controlled instances derived from the ideation--execution
trajectories described in Section~\ref{sec:data_collection}. Each source
project includes a final paper and an available reference codebase, allowing
the target implementation detail to be verified against an executed research
artifact. We stratify the sample by Level-1 defect category and retain
instances whose target detail can be evaluated through a bounded
implementation component, such as a model module, preprocessing rule, loss
definition, inference procedure, or evaluation computation.

We exclude cases requiring large-scale model training, private datasets,
unavailable external services, or reproduction of a complete experimental
pipeline. For each retained instance, we construct fixed instance-specific
tests or verification criteria from the final paper and reference code before
inspecting either generated implementation.

For each instance, we use two specifications: a
\textsc{Direct Generation} specification produced from the underspecified
idea alone, and a \textsc{Clarification-Assisted Generation} specification
produced with the annotated defect, clarification action, and oracle
resolution. We then ask GPT-5.6-Sol to implement each specification as minimal
runnable code. The implementation prompt and decoding settings are identical
across conditions. The model is not given the original paper, reference
codebase, hidden gold resolution, or generation-condition label.

\paragraph{Executable Evaluation.}
Because all selected instances are derived from ideation--execution
trajectories, each instance is associated with an available final paper and
reference codebase. We use this paper--code pair to identify the
evidence-supported target implementation detail and to construct
instance-specific executable checks before inspecting any generated code.

We define \textsc{Target Fidelity} as whether the generated implementation
faithfully instantiates the target method detail encoded in the reference
implementation. Depending on the target component, fidelity is assessed
through output equivalence on fixed synthetic inputs, targeted unit tests, or
blinded inspection against a predefined implementation criterion. The same
inputs, tests, and criteria are applied to both generation conditions.

\paragraph{Results.}
Table~\ref{tab:executable_validation} shows that the benefits of oracle
clarification extend from textual specifications to executable
implementations. Code generated from \textsc{Clarification-Assisted}
specifications is more often runnable than code generated through
\textsc{Direct Generation} (95.0\% vs.\ 70.0\%) and is nearly twice as likely
to pass all predefined instance-specific tests (85.0\% vs.\ 45.0\%).
Most importantly, clarification increases target fidelity from 30.0\% to
90.0\%, where fidelity is evaluated against the method detail encoded in the
available final paper and reference implementation. The proportion of cases
requiring further clarification also decreases from 35.0\% to 5.0\%.

The gap between runnability and target fidelity is particularly informative.
Although 70.0\% of implementations generated directly from underspecified
inputs execute successfully, only 30.0\% faithfully instantiate the target
method detail. The remaining implementations often adopt a plausible but
unsupported default for the unresolved choice. Oracle-resolved specifications
therefore improve not only code executability, but also agreement with the
executed reference method. Because the study evaluates bounded components
rather than complete training and evaluation pipelines, these results provide
component-level evidence of improved codification fidelity rather than full
reproduction of paper-level experimental results.
\subsection{Source-Level Robustness and Uncertainty}
\label{app:source_uncertainty}

Multiple evaluation instances may originate from the same underlying
research source. For example, multiple GitHub issues may come from the
same repository, and multiple controlled defects may be constructed
from the same reproduced paper or executed research project. Such
instances can share terminology, methodological structure,
specification style, and source artifacts. Treating them as fully
independent may therefore underestimate uncertainty and give
disproportionate influence to sources contributing multiple instances.

\paragraph{Source Clusters and Evaluation Sets.}
We assign each benchmark instance a source-cluster identifier.
For GitHub-derived instances, the source cluster is the repository.
For real-world and controlled-synthetic instances derived from
reproducibility reports, it is the original paper being reproduced.
For instances derived from ideation--execution trajectories, it is the
executed research project. All evaluation inputs derived from the same
repository, source paper, or project share one source-cluster
identifier.

Task~1 contains 100 READY and 100 NOTREADY specifications sampled
independently from the corresponding benchmark records. The two
readiness classes are not restricted to matched counterpart pairs, and
each specification is evaluated in isolation. A source cluster may
therefore contribute inputs to both readiness classes. Tasks~2 and~3
use the same NOTREADY benchmark instances within each subset.
Table~\ref{tab:source_cluster_statistics} reports the source-cluster
composition of the exact fixed evaluation sets.

\paragraph{Source-Clustered Bootstrap.}
We estimate uncertainty using a source-clustered bootstrap with
10,000 replicates. For each task--subset evaluation set, we sample the
observed source clusters with replacement. Whenever a source cluster is
sampled, all of its associated evaluation inputs are included with the
same bootstrap multiplicity. We then recompute the complete primary
metric on the resampled data: Macro-F1 for Task~1, Macro Defect Recovery
Rate for Task~2, and Macro Clarification Action Success Rate for
Task~3. Resampling therefore occurs before metric aggregation rather
than over already aggregated instance-level scores.

For Task~1, each retained replicate contains both readiness classes.
For Tasks~2 and~3, the Level-2 category set is fixed to the categories
represented in the original evaluation set, and a replicate is retained
only when every such category is represented. We report percentile
95\% confidence intervals using the 2.5th and 97.5th percentiles of
the bootstrap distribution.

\paragraph{Source-Balanced Estimates.}
We additionally compute source-balanced estimates to test whether the
results are disproportionately influenced by sources contributing
multiple evaluation inputs.

For Task~1, let $\mathcal{I}_{g,y}$ denote the inputs associated with
source cluster $g$ and gold readiness class
$y\in\{\textsc{Ready},\textsc{NotReady}\}$, and let
$\mathcal{G}_{y}$ denote the source clusters represented in class $y$.
Each input $i\in\mathcal{I}_{g,y}$ receives weight
\begin{equation}
w_i
=
\frac{1}
{|\mathcal{G}_{y}|\,
 |\mathcal{I}_{g,y}|}.
\end{equation}
This assigns equal total weight to each source cluster within each gold
readiness class. We construct the corresponding weighted confusion
matrix, compute the class-specific F1 scores, and take their unweighted
mean.

For Tasks~2 and~3, let $h_i^{(t)}\in\{0,1\}$ denote whether instance
$i$ satisfies the complete primary success criterion for task $t$.
For Task~2, this requires correct blocker localization and both
taxonomy labels. For Task~3, it requires an action that addresses the
target blocker, obtains sufficient information, and introduces no
unsupported assumption.

For source cluster $g$ and gold Level-2 category $c$, we first compute
\begin{equation}
\bar{h}_{g,c}^{(t)}
=
\frac{1}{|\mathcal{I}_{g,c}|}
\sum_{i\in\mathcal{I}_{g,c}} h_i^{(t)},
\end{equation}
where $\mathcal{I}_{g,c}$ contains the evaluated instances from source
cluster $g$ with gold Level-2 category $c$. The source-balanced primary
metric is then
\begin{equation}
M_{\mathrm{src}}^{(t)}
=
\frac{1}{|\mathcal{C}^{(2)}|}
\sum_{c\in\mathcal{C}^{(2)}}
\frac{1}{|\mathcal{G}_{c}|}
\sum_{g\in\mathcal{G}_{c}}
\bar{h}_{g,c}^{(t)},
\end{equation}
where $\mathcal{G}_{c}$ is the set of source clusters represented in
category $c$. This retains the original category-level macro averaging
while assigning equal weight to each represented source within a
category.

\begin{table*}[t]
\centering
\small
\setlength{\tabcolsep}{5.0pt}
\renewcommand{\arraystretch}{1.10}

\begin{tabular}{llcccc}
\toprule
\textbf{Task}
& \textbf{Subset}
& \textbf{$N$}
& \textbf{Point}
& \shortstack{\textbf{Source-Clustered}\\\textbf{95\% CI}}
& \shortstack{\textbf{Source-}\\\textbf{Balanced}}
\\
\midrule

Task 1
& Real
& 200
& 67.5
& [59.8, 74.6]
& 66.9
\\

Task 1
& Synthetic
& 200
& 86.4
& [80.9, 91.2]
& 85.8
\\

Task 2
& Real
& 163
& 9.6
& [3.1, 18.7]
& 9.4
\\

Task 2
& Synthetic
& 200
& 12.2
& [6.8, 18.5]
& 11.7
\\

Task 3
& Real
& 163
& 80.6
& [70.1, 89.0]
& 81.2
\\

Task 3
& Synthetic
& 200
& 96.2
& [92.0, 99.0]
& 96.0
\\

\bottomrule
\end{tabular}

\caption{
Source-level robustness of the primary GPT-5.6-Sol results reported in
Table~\ref{tab:main_results}. Point reproduces the original
instance-level metric on the exact fixed evaluation set.
Source-clustered 95\% confidence intervals are obtained by resampling
complete repositories, source papers, or executed projects.
Source-Balanced reports the corresponding estimate under the
source-balanced weighting procedure. Source-balanced estimates differ
from the instance-level results by at most 0.6 points.
}
\label{tab:source_level_primary_results}
\end{table*}

\paragraph{Paired Source-Clustered Comparisons.}
We use paired source-clustered bootstrap comparisons to assess whether
differences between GPT-5.6-Sol and Claude Sonnet~5 are stable after
accounting for source dependence. In each replicate, the same sampled
source-cluster multiplicities are applied to both models, and we
compute
\begin{equation}
\Delta^{(b)}
=
M_A^{(b)}-M_B^{(b)}.
\end{equation}
We report the 2.5th and 97.5th percentiles of the paired-difference
distribution. Two-sided bootstrap $p$-values are computed from the
proportion of replicates on either side of zero. The six model
comparisons are exploratory, and their $p$-values are Holm-adjusted
across the six task--subset combinations.

We separately apply the paired source-clustered bootstrap to the
\textsc{End-to-End} and \textsc{Defect-Guided} clarification settings.
These conditions use the same GPT-5.6-Sol model, decoding
configuration, evaluator, and real-world Task~3 instances. Because this
is a pre-specified test of the blocker-availability effect, it is
reported separately from the exploratory model-comparison family.

\begin{table*}[t]
\centering
\small
\setlength{\tabcolsep}{5.0pt}
\renewcommand{\arraystretch}{1.10}

\begin{tabular}{lllrrrr}
\toprule
\textbf{Task}
& \textbf{Subset}
& \textbf{Paired Contrast}
& \textbf{$\Delta$}
& \textbf{95\% CI}
& \textbf{$p$}
& \shortstack{\textbf{CI Excludes}\\\textbf{Zero}}
\\
\midrule

Task 1
& Real
& GPT-5.6-Sol $-$ Claude Sonnet 5
& +8.2
& [-0.3, 16.5]
& 0.210
& No
\\

Task 1
& Synthetic
& GPT-5.6-Sol $-$ Claude Sonnet 5
& +10.2
& [3.4, 17.0]
& 0.024
& Yes
\\

Task 2
& Real
& GPT-5.6-Sol $-$ Claude Sonnet 5
& +2.8
& [-0.2, 7.3]
& 0.320
& No
\\

Task 2
& Synthetic
& GPT-5.6-Sol $-$ Claude Sonnet 5
& +0.8
& [-3.2, 5.5]
& 1.000
& No
\\

Task 3
& Real
& GPT-5.6-Sol $-$ Claude Sonnet 5
& +3.8
& [-8.0, 15.5]
& 1.000
& No
\\

Task 3
& Synthetic
& GPT-5.6-Sol $-$ Claude Sonnet 5
& +1.7
& [-1.9, 6.5]
& 0.720
& No
\\

\midrule

Task 3
& Real
& \textsc{Defect-Guided} $-$ \textsc{End-to-End}
& +67.0
& [59.0, 75.2]
& $<0.001$
& Yes
\\

\bottomrule
\end{tabular}

\caption{
Paired source-clustered bootstrap comparisons. The same sampled
source-cluster multiplicities are applied to both systems in every
replicate. For the first six rows, $p$-values are Holm-adjusted across
the exploratory GPT-5.6-Sol versus Claude Sonnet~5 comparisons. The
final row reports the separate pre-specified comparison between
\textsc{Defect-Guided} and \textsc{End-to-End} clarification on the
same real-world Task~3 instances. The final column indicates whether
the corresponding source-clustered 95\% confidence interval excludes
zero.
}
\label{tab:source_clustered_comparisons}
\end{table*}

\paragraph{Results.}
The evaluation sets exhibit nontrivial source-level clustering
(Table~\ref{tab:source_cluster_statistics}). For example, the 200
Task~1 real-world and controlled-synthetic inputs originate from only
34 and 31 source clusters, respectively. Nevertheless, the
source-balanced GPT-5.6-Sol estimates differ from the corresponding
instance-level estimates by at most 0.6 points across all six
task--subset combinations
(Table~\ref{tab:source_level_primary_results}). The primary cross-task
pattern is therefore not driven by a small number of sources
contributing multiple evaluation inputs.

On real-world instances, Macro DRR remains low at 9.6
($95\%~\mathrm{CI}=[3.1,18.7]$), whereas defect-guided Macro-CAS
reaches 80.6 ($95\%~\mathrm{CI}=[70.1,89.0]$). Five of the six
exploratory GPT-5.6-Sol versus Claude Sonnet~5 comparisons have
confidence intervals containing zero. Only the Task~1
controlled-synthetic difference remains reliable after Holm
adjustment. We therefore interpret most small differences between the
two leading models descriptively rather than as stable rankings.

In contrast, providing the annotated blocker improves real-world
Macro-CAS by 67.0 points under paired source-clustered resampling
($95\%~\mathrm{CI}=[59.0,75.2]$, $p<0.001$). The confidence interval
remains far from zero after resampling complete research sources.
Thus, the central conclusion that blocker identification, rather than
clarification formulation once the blocker is known, is the primary
bottleneck remains robust to dependence among instances derived from
the same research source.

\section{Data Construction and Annotation Details}
\label{app:data-annotation-details}

\subsection{Source-Specific Candidate Verification}
\label{app:source-verification}

\paragraph{Review Structure.}
Candidates from all three construction paths---GitHub issues,
reproducibility reports, and ideation--execution trajectories---undergo the
same two-stage human review. A primary annotator with experience in
machine-learning research screens every extracted or generated candidate and
assigns a preliminary \textsc{Keep}, \textsc{Uncertain}, or
\textsc{Reject} decision. A second annotator with machine-learning research
experience independently reviews all candidates provisionally retained by
the primary annotator, together with all candidates marked as uncertain.
This design provides complete initial coverage while concentrating
independent verification on cases that may enter the benchmark or require
additional judgment.

This stage evaluates candidate validity and evidence support only. Level-1
and Level-2 taxonomy labels are assigned and validated separately after
candidate inclusion, as described in
Appendix~\ref{app:annotation-quality}.

\paragraph{Source Materials.}
Reviewers receive the source materials relevant to each construction path.
For reproducibility-report candidates, these include the extracted gap and
proposed resolution, the corresponding report passages, relevant portions of
the original paper, and available supporting artifacts such as appendices,
source code, configuration files, or author responses. For GitHub candidates,
reviewers inspect the issue thread, the associated repository and paper
context, and the discussion or implementation evidence supporting the
resolution. For ideation--execution candidates, reviewers receive the final
paper--code pair, the reconstructed reference specification, the modified
candidate, and the implementation-critical detail removed or abstracted
during controlled construction.

Across all sources, reviewers are instructed to rely only on the available
evidence. They must not infer the intended method from conventional practice,
personal preference, or an implementation choice that merely appears
plausible.

\paragraph{Verification of Real-World Candidates.}
A real-world candidate from a reproducibility report or GitHub issue is
retained only if it satisfies all of the following criteria:

\begin{enumerate}
    \item \textbf{Genuine specification gap.}
    The candidate identifies an omission, ambiguity, or inconsistency in the
    original description of the research method, rather than reader
    confusion, an implementation error introduced by the reproducer, or a
    request for an optional improvement.

    \item \textbf{Method-core relevance.}
    The unresolved choice affects faithful implementation of the method,
    including its objective, formal definition, model structure, algorithmic
    procedure, training protocol, data construction, preprocessing,
    inference behavior, or evaluation protocol.

    \item \textbf{Atomicity.}
    The candidate contains exactly one primary implementation blocker.
    Candidates containing multiple independent gaps are decomposed when the
    source evidence permits; otherwise, they are excluded.

    \item \textbf{Self-contained underspecified input.}
    The benchmark input contains enough context to understand the proposed
    method and the unresolved implementation decision without revealing the
    resolution.

    \item \textbf{Evidence-supported clarification.}
    The source discussion, reproducibility report, paper, or associated
    artifact provides enough information to formulate a concrete
    clarification that resolves the gap without unsupported inference.
\end{enumerate}

We exclude cases involving only software installation, dependency versions,
hardware resources, runtime, credentials, inaccessible external resources,
formatting, or other non-methodological engineering concerns. We also exclude
performance discrepancies when the available evidence does not identify a
specific defect in the original method specification.

\paragraph{Verification of Codification-Ready References.}
Controlled synthetic construction begins from a reference specification
derived either from a successfully reproduced paper--report pair or from a
completed ideation--execution paper--code pair. A reconstructed reference is
accepted as codification-ready only when the available artifacts jointly
expose the method-defining choices required to recover the reproduced or
executed method. In particular, reviewers verify that the component selected
for controlled modification is explicitly specified or directly instantiated
in the source artifacts and that no unresolved paper--code or paper--report
conflict remains for that component.

The purpose of this review is not to repeat the reproduction or execution
study. Instead, it verifies that the reconstructed specification is a valid
starting point for controlled defect injection and that the target detail has
a source-supported value that can serve as the reference resolution.

\paragraph{Verification of Controlled Synthetic Candidates.}
A controlled synthetic candidate is retained only if it satisfies all of the
following criteria:

\begin{enumerate}
    \item \textbf{Single controlled modification.}
    Exactly one implementation-critical detail is removed, abstracted, or
    altered relative to the verified reference.

    \item \textbf{Preservation of non-target content.}
    All implementation-critical information unrelated to the target defect
    remains unchanged.

    \item \textbf{Method-core consequence.}
    The modification leaves a method-defining choice underdetermined and
    permits materially different implementations, rather than merely omitting
    a routine engineering preference.

    \item \textbf{Recoverable target detail.}
    The removed or altered information is directly recoverable from the
    source paper, report, codebase, or other supporting artifact.

    \item \textbf{No resolution leakage.}
    The modified specification does not directly reveal the hidden reference
    detail or otherwise make the clarification unnecessary.

    \item \textbf{Resolvable clarification.}
    A targeted clarification question or evidence-seeking action could obtain
    the information needed to restore codification readiness without
    introducing an unsupported implementation choice.
\end{enumerate}

Candidates are rejected when the transformation introduces multiple
interacting defects, changes the scientific objective, distorts non-target
method content, or relies on a reference detail that cannot be grounded in
the source artifacts.

\paragraph{Decisions and Adjudication.}
For each candidate, the primary annotator records the preliminary decision,
the supporting evidence spans, the target implementation blocker, and the
source-supported resolution. The second annotator independently reassesses
all \textsc{Keep} and \textsc{Uncertain} cases using the same source
materials and verification criteria.

When the reviewers disagree on inclusion, method-core relevance, atomicity,
reference readiness, transformation validity, or evidence sufficiency, they
jointly re-examine the source materials and resolve the case through
discussion. Inclusion requires agreement that the target gap is
implementation-critical, atomic, and supported by the available evidence.
Cases that remain ambiguous, contain unresolved evidence conflicts, or
require unsupported interpretation after adjudication are discarded.

\subsection{Target-Uniqueness and Alternative-Blocker Audit}
\label{app:target_uniqueness}

\ourbench{} evaluates defect localization under a single-target assumption:
each \textsc{NotReady} specification should contain one annotated
implementation-critical blocker, and a prediction is counted as localized only
when it recovers that target. A potential validity concern is that a
specification may inadvertently contain another independent blocker that is
also sufficient to prevent faithful codification. In such a case, a model
could identify a genuine defect but be penalized for not matching the
benchmark-selected target. We therefore conduct a target-uniqueness audit to
verify that the annotated defect is the unique primary blocker in each retained
instance.

\paragraph{Audit Scope.}
We audit all \numInstances{} \textsc{NotReady} specifications in \ourbench{},
including \numRealInstances{} real-world and \numSyntheticInstances{}
controlled-synthetic instances. The audit is conducted on the
\textsc{NotReady} inputs rather than on their corresponding \textsc{Ready}
counterparts because Task~2 requires models to localize the blocker from the
underspecified specification alone.

\paragraph{Two-Stage Annotation Protocol.}
Two annotators with machine-learning research experience independently review
each instance using a two-stage protocol designed to reduce anchoring to the
existing gold annotation. In the first stage, annotators receive only the
\textsc{NotReady} specification and the minimum source context needed to
interpret it. They do not receive the annotated defect, taxonomy labels, gold
clarification, or resolved specification. Each annotator determines whether
the specification is codification-ready and enumerates up to three candidate
implementation blockers. For each candidate, the annotator records the
affected methodological component, unresolved implementation decision, and
whether competent implementers could make materially different
method-defining choices.

In the second stage, annotators receive the annotated target defect, its
supporting evidence, and the evidence-grounded resolution. They then assess
whether the target is valid, whether it is the primary implementation blocker,
and whether any independently identified alternative constitutes a separate
implementation-critical defect. An alternative is considered
\emph{co-primary} if it would independently leave the specification
\textsc{NotReady} even after the annotated target were resolved. Routine
engineering choices, non-core hyperparameters, presentation omissions, and
details that do not change the instantiated scientific method are not treated
as alternative blockers.

For each instance, annotators assign \texttt{gold\_target\_valid},
\texttt{gold\_target\_primary}, and \texttt{target\_unique} labels from
\{\textsc{Yes}, \textsc{No}, \textsc{Unclear}\}. When an alternative issue is
identified, they additionally record its description, affected component,
unresolved decision, severity (\textsc{Co-Primary},
\textsc{Secondary-Critical}, or \textsc{Non-Blocking}), and relationship to
the gold target (\textsc{Independent}, \textsc{Overlapping}, or
\textsc{Dependent}).

\paragraph{Adjudication and Dataset Revision.}
All cases receiving a \textsc{No} or \textsc{Unclear} judgment on target
validity, target primacy, or target uniqueness, together with all annotator
disagreements, are reviewed by a third machine-learning researcher. The
adjudicator assigns one of five final decisions: \textsc{Keep},
\textsc{Revise-Gold}, \textsc{Revise-Specification}, \textsc{Split}, or
\textsc{Drop}.

We retain an instance unchanged only when the annotated target is valid,
primary, and unique. When the target is correct but another blocker is
accidentally present, we revise the specification using source-supported
information so that only the intended target remains unresolved. When the
existing gold does not capture the primary blocker, we revise the gold
annotation and taxonomy labels. Cases containing multiple independent,
evidence-resolved blockers are split only when each can be converted into a
self-contained single-target instance by resolving all non-target blockers.
Cases that cannot be repaired without unsupported inference are removed. All
revised or newly split instances undergo a second verification round.

\paragraph{Evaluation Measures.}
We report the percentage of audited instances for which the original target is
valid, primary, and the unique primary blocker. We also report the frequency
of independently identified co-primary alternatives and the number of original
instances retained, revised, split, or removed after adjudication.

\paragraph{Results.}
Table~\ref{tab:target_uniqueness_audit} summarizes the audit results. Across
the real-world and controlled-synthetic subsets, 91.4\% and 95.6\% of the
original instances, respectively, contain a target independently judged to be
the unique primary implementation blocker. Co-primary alternative blockers are
identified in 4.9\% of real-world instances and 2.4\% of
controlled-synthetic instances. Following adjudication, 624 original instances
are retained unchanged, 21 are revised, 7 are split into 15 final
single-target instances, and 8 are removed. The final benchmark therefore
retains \numInstances{} single-target instances. Only instances whose
annotated target remains the unique primary blocker after adjudication are
retained.

\begin{table*}[t]
\centering
\scriptsize
\setlength{\tabcolsep}{3.6pt}
\renewcommand{\arraystretch}{1.02}

\begin{tabular*}{\textwidth}{@{\extracolsep{\fill}}lrrrrrrr@{}}
\toprule
\textbf{Subset}
& \textbf{$N$}
& \shortstack{\textbf{Orig. Unique}\\\textbf{Primary (\%)}}
& \shortstack{\textbf{Co-primary}\\\textbf{Alt. (\%)}}
& \textbf{Revised}
& \textbf{Split}
& \textbf{Dropped}
& \shortstack{\textbf{Final}\\\textbf{$N$}}
\\
\midrule

Real-world
& 163
& 91.4
& 4.9
& 8
& 3
& 3
& 163
\\

Controlled synthetic
& 497
& 95.6
& 2.4
& 13
& 4
& 5
& 497
\\

\midrule

\textbf{Overall}
& \textbf{660}
& \textbf{94.5}
& \textbf{3.0}
& \textbf{21}
& \textbf{7}
& \textbf{8}
& \textbf{660}
\\

\bottomrule
\end{tabular*}

\caption{
Target-uniqueness audit of all \textsc{NotReady} specifications.
\textbf{Orig. Unique Primary} reports original targets judged valid, primary,
and unique; \textbf{Co-primary Alt.} reports independent alternative blockers.
\textbf{Revised}, \textbf{Split}, and \textbf{Dropped} count original audited
instances. \textbf{Final $N$} reports retained single-target instances after
adjudication; 7 split instances yield 15 final instances, offsetting 8 drops.
}
\label{tab:target_uniqueness_audit}
\end{table*}

\subsection{Multi-Defect Readiness and Localization Ablation}
\label{app:multi_defect_ablation}

\ourbench{} evaluates the controlled single-target setting defined in
Section~\ref{sec:task_formulation}. To probe whether this setting understates the
difficulty of specifications that contain more than one
implementation-critical defect, we conduct an exploratory ablation in
which two known, independently validated defects are combined into one
specification.

\paragraph{Construction.}
The controlled-synthetic construction pipeline (Appendix~\ref{app:ready_counterparts})
injects each defect into a project's shared codification-ready reference
independently, so a project with multiple synthetic instances has several
single-defect specifications that all derive from the same reference. We
identify all projects with at least two such sibling instances and, for
\numMultiDefectInstances{} of them, use \texttt{deepseek/\allowbreak
deepseek-v4-pro} to rewrite the shared reference into one specification
that simultaneously obscures two of the project's already-validated target
defects, reusing each defect's existing gold detail and surface-form
style rather than inventing new content. We stratify the
\numMultiDefectInstances{} pairs evenly by whether the two defects share
the same \texttt{codification\_slot} (25 same-slot, 25 different-slot),
since this is the axis most likely to affect whether the two defects are
confusable. A preservation check (both details obscured, no third defect
introduced, non-target content preserved) is required for every
constructed instance, and all \numMultiDefectInstances{} pass.

\paragraph{Evaluation.}
We evaluate GPT-5.6-Sol on Task~1 (readiness) and a multi-defect variant
of the Task~2 prompt that keeps the same taxonomy, boundary rules, and
selectivity criteria as the single-target prompt (Figure~\ref{fig:task2_prompt})
but no longer caps the response at one defect, asking the model to
report only the defect or defects it is confident meet the criteria
rather than a fixed count. Task~2 predictions are scored against each of
the two gold targets independently with the same Claude Opus~4.8
semantic judge used elsewhere in the paper (Appendix~\ref{app:judge_validation}),
and we report \textsc{Any-of-2} (at least one gold target recovered) and
\textsc{All-of-2} (both recovered). As a paired baseline, we evaluate the
same two single-defect siblings independently under the original
single-target prompt and combine their outcomes (AND for readiness and
All-of-2, OR for Any-of-2), so both conditions cover the identical pair
of defects. 
\paragraph{Results.}
Table~\ref{tab:multi_defect_ablation} reports the comparison. Readiness
assessment is directionally higher when both defects are combined into
one specification (88.0\% vs.\ 80.0\%, $\Delta=+8.0$, McNemar
$p=0.289$), consistent with the intuition that a specification with more
independent problems is at least as likely to be flagged \textsc{NotReady}.
We find no evidence that combining two defects into one specification
makes strict localization harder: All-of-2 is directionally higher, not
lower, when the defects are combined (8.0\% vs.\ 2.0\%, $\Delta=+6.0$,
$p=0.250$), and Any-of-2 is significantly higher (64.0\% vs.\ 36.0\%,
$\Delta=+28.0$, $p=0.0013$). 

\paragraph{Scope.}
This ablation uses only the controlled-synthetic subset, because
real-world sibling instances derived from the same paper or repository do
not share a common reference specification and cannot be combined without
additional content synthesis. Appendix~\ref{app:target_uniqueness} shows
that 94.5\% of \ourbench{} instances overall have a unique primary
blocker under the existing single-target construction, so this ablation
is a secondary, exploratory check rather than a claim about the frequency
of naturally co-occurring defects.

\subsection{Benchmark Annotation Guidelines}
\label{app:human-annotation-guideline}

\begingroup
\setlength{\parindent}{0pt}
\setlength{\parskip}{0.35em}

This appendix presents the complete human annotation guideline used to assess
whether a research idea or method specification is sufficiently clear for
faithful initial implementation. Annotators judge each specification using only
the supplied text and are not permitted to consult the original paper, source
code, or additional implementation materials.

\subsubsection{Annotation Objective}
\label{app:annotation-objective}

\begin{guidelinebox}{Core Annotation Question}
Can a competent machine-learning researcher construct a faithful initial
implementation or experimental prototype from the supplied specification
without inventing an unsupported assumption about the core method?
\end{guidelinebox}

Annotators should assume that the implementer may use:

\begin{itemize}[leftmargin=1.5em, itemsep=1pt, topsep=2pt]
    \item standard machine-learning libraries;
    \item canonical implementations of named standard components; and
    \item ordinary engineering defaults.
\end{itemize}

The target is a faithful initial implementation or experimental prototype,
rather than exact reproduction of every reported numerical result.

\subsubsection{Readiness Labels}
\label{app:readiness-labels}

\begin{readybox}
\textbf{\READY.}
Label the specification \READY{} when the implementer can construct a faithful
initial implementation or experimental prototype without inventing an
unsupported assumption about the core method.

The specification does not need to fix every ordinary hyperparameter,
engineering detail, random seed, software version, or explicitly open modular
choice.
\end{readybox}

\begin{notreadybox}
\textbf{\NOTREADY.}
Label the specification \NOTREADY{} when an omission, ambiguity, or internal
inconsistency leaves an implementation-critical choice underdetermined.
\end{notreadybox}

\begin{unsurebox}
\textbf{\UNSURE.}
Use \UNSURE{} only when the case cannot be reliably classified and requires
expert adjudication. It should not be used merely because the annotator finds
the method technically unfamiliar.
\end{unsurebox}

\subsubsection{Implementation-Critical Choices}
\label{app:implementation-critical}

Implementation-critical choices are decisions that affect the identity,
behavior, training logic, inference logic, or faithful evaluation of the
proposed method. They may concern:

\begin{itemize}[leftmargin=1.5em, itemsep=1pt, topsep=2pt]
    \item the research task, inputs, or outputs;
    \item the core algorithmic procedure or method operation;
    \item the model structure or computational components;
    \item the objective, loss, reward, or supervision specification;
    \item the training or optimization procedure;
    \item the data specification, including construction or preprocessing procedures;
    \item the inference or decision rule;
    \item the evaluation specification; or
    \item consistency across different parts of the specification.
\end{itemize}

\begin{guidelinebox}{Ordinary Engineering Choices}
Ordinary engineering choices generally do not make a specification
\NOTREADY{}, unless the choice is itself central to the proposed contribution.

Examples include:

\begin{itemize}[leftmargin=1.4em, itemsep=1pt, topsep=2pt]
    \item random seeds, hardware, file paths, and logging frequency;
    \item software patch versions;
    \item conventional batch sizes;
    \item conventional optimizer settings; and
    \item minor implementation details that do not materially change the method.
\end{itemize}
\end{guidelinebox}

\subsubsection{Specification Slots}
\label{app:specification-slots}

For a \NOTREADY{} case, the annotator selects the \emph{single specification
slot most directly affected by the primary implementation blocker}. The slot
should correspond to the information that must be clarified, rather than every
component that may be indirectly affected.

\subsubsection{Counterfactual Test}
\label{app:counterfactual-test}

\begin{guidelinebox}{Material-Divergence Test}
Could two competent implementers follow the specification and produce
materially different versions of the core method, with no evidence indicating
which version is intended?
\end{guidelinebox}

If the answer is yes, the specification is generally \NOTREADY{}.

Differences caused only by ordinary engineering choices do not make the
specification \NOTREADY{}. Annotators should focus on whether two
implementations would differ materially in the method's identity, behavior,
training logic, inference logic, or faithful evaluation.

\subsubsection{Annotation Procedure}
\label{app:annotation-procedure}

Annotators follow five steps. Table~\ref{tab:annotation-output-fields}
lists the output fields and allowed values.

\begin{enumerate}[leftmargin=1.7em, itemsep=3pt, topsep=3pt]

\item \textbf{Assess readiness.}
Assign \READY{}, \NOTREADY{}, or \UNSURE{} using the decision
rules below.

\item \textbf{Select the affected slot.}
For a \NOTREADY{} case, select the single specification slot
most directly affected by the primary blocker.

\item \textbf{Locate the blocker.}
Highlight the smallest text span that reveals the defect.
For an inconsistency, highlight both conflicting spans.

\item \textbf{Describe the required clarification.}
In one or two sentences, identify the unresolved decision and
the information needed to resolve it, without proposing an
unsupported solution.

\item \textbf{Assess atomicity.}
Record whether the specification contains no blocker, one
primary blocker, multiple independent blockers, or unclear
blocker boundaries. Count independent defects rather than
affected sentences or components. Multiple blockers are
independent when resolving one would leave another unresolved.

\end{enumerate}

For \READY{} cases, set the affected slot to \code{NONE},
leave the blocking span and required clarification empty,
and record \code{NO_BLOCKER}.

\subsubsection{Decision Rules}
\label{app:decision-rules}

\begin{readybox}
\textbf{Label the specification \READY{} when all of the following hold:}

\begin{enumerate}[leftmargin=1.7em, itemsep=2pt, topsep=3pt]
    \item The task and intended output are sufficiently clear.
    \item The core method can be implemented without inventing an unsupported
    assumption.
    \item No implementation-critical contradiction is present.
    \item Any remaining unspecified choices are ordinary engineering choices or
    explicitly open modular choices.
    \item Two competent implementers would not produce materially different
    versions of the core method solely because of missing information in the
    specification.
\end{enumerate}
\end{readybox}

\begin{notreadybox}
\textbf{Label the specification \NOTREADY{} when at least one of the following
holds:}

\begin{enumerate}[leftmargin=1.7em, itemsep=2pt, topsep=3pt]
    \item A core algorithmic operation is missing or ambiguous.
    \item The structure or interaction of essential model components is unclear.
    \item The objective, supervision, or training logic is underdetermined.
    \item Essential data construction, preprocessing, inference, or decision
    logic is missing.
    \item The evaluation protocol is too unclear to faithfully assess the
    central research claim.
    \item Two or more parts of the specification are internally inconsistent.
    \item The implementer would need to invent an unsupported assumption that
    could materially change the method.
\end{enumerate}
\end{notreadybox}

\begin{guidelinebox}{Exact Reproduction Is Not Required}
Do not label a specification \NOTREADY{} solely because it lacks details needed
for exact numerical reproduction. The target is a faithful initial
implementation or experimental prototype, not exact replication of every
reported result.
\end{guidelinebox}

\subsubsection{Important Annotation Reminders}
\label{app:annotation-reminders}

\begin{guidelinebox}{Important Reminders}
\begin{itemize}[leftmargin=1.5em, itemsep=2pt, topsep=2pt]
    \item Judge only the supplied specification.
    \item Do not use prior knowledge of the original paper or method to fill in
    missing details.
    \item Do not assume that a plausible implementation is necessarily the
    intended implementation.
    \item Do not penalize explicitly open modular choices when the
    specification clearly states that multiple alternatives are allowed.
    \item Select the slot corresponding to the information that must be
    clarified, rather than every component that may be indirectly affected.
    \item When multiple issues are present, identify the primary blocker and
    select \texttt{MULTIPLE\_INDEPENDENT\_BLOCKERS} if additional independent
    blockers remain.
\end{itemize}
\end{guidelinebox}

\subsubsection{Annotation Output Fields}
\label{app:output-fields}

Table~\ref{tab:annotation-output-fields} lists the fields collected for each case.

\subsubsection{Annotation Summary}
\label{app:annotation-summary}

\begin{summarybox}
For each specification, annotators produce a structured annotation consisting
of a readiness judgment and, when applicable, a defect characterization.

For a \READY{} decision:

\begin{enumerate}[leftmargin=1.7em, itemsep=3pt, topsep=3pt]
    \item Assign the \READY{} label.
    \item Set the affected specification slot to \texttt{NONE}.
    \item Leave defect-related fields empty.
\end{enumerate}

For a \NOTREADY{} decision:

\begin{enumerate}[leftmargin=1.7em, itemsep=3pt, topsep=3pt]
    \item Assign the \NOTREADY{} label.
    \item Identify the primary implementation blocker.
    \item Assign the corresponding Level-1 defect type:
    \textsc{Ambiguity}, \textsc{Incompleteness}, or
    \textsc{Inconsistency}.
    \item Assign the corresponding Level-2 defect category that best
    characterizes the primary blocker.
    \item Highlight the smallest relevant text span revealing the defect.
    \item Identify the affected specification slot.
    \item Describe the specific implementation-critical information that must
    be clarified before implementation.
    \item Determine whether the case contains exactly one primary blocker or
    multiple independent blockers.
\end{enumerate}
\end{summarybox}

\endgroup

\subsection{Annotation Reliability}
\label{app:annotation-quality}
To ensure the reliability of benchmark annotations, we conduct an
inter-annotator agreement study on a manually reviewed subset of benchmark
instances. We randomly sample \numRealInstances{} real-world instances and \numSyntheticEvalInstances{} controlled
synthetic instances, covering diverse research domains and defect categories.
Two graduate-level annotators with backgrounds in computer science and machine
learning independently annotate each instance following the annotation
guidelines described in Appendix~\ref{app:human-annotation-guideline}, without
accessing each other's decisions.
After independent annotation, all disagreements are adjudicated by a senior
annotator with machine-learning research experience. The adjudicator reviews
the instance, both annotations, and the relevant source evidence, then assigns
the final readiness label, Level-1 label, Level-2 label, and primary target
defect when applicable. These adjudicated decisions replace the preliminary
annotations and constitute the final gold labels used in \ourbench{}.

We measure agreement on three core annotation components: codification
readiness labels, Level-1 defect types, and Level-2 defect categories. For
\NOTREADY{} instances, we additionally evaluate whether annotators identify
the same primary target defect. We report raw agreement and Cohen's
$\kappa$, separately for the real-world and controlled synthetic subsets.
Cohen's $\kappa$ is omitted for target defect matching because the annotation
setting does not yield a stable categorical distribution for reliable
chance-corrected agreement.

Table~\ref{tab:annotation_agreement} summarizes the annotation agreement
results over the manually reviewed subset. Annotators achieve consistently
high agreement across readiness classification and defect taxonomy labels.
For readiness assessment, the agreement reaches 92.0\% and 96.0\% on the
real-world and synthetic subsets, with Cohen's $\kappa$ values of 0.84 and
0.92, respectively, indicating strong consistency beyond chance agreement.
Defect taxonomy annotations show similar reliability, with Level-1 label
agreement above 93\% on both subsets and substantial agreement for Level-2
categories ($\kappa=0.92$ on real-world and $\kappa=0.85$ on synthetic
instances). These pre-adjudication agreement results demonstrate that the
proposed annotation guidelines enable consistent identification of
implementation-critical specification defects, while the final benchmark labels
reflect the adjudicated gold decisions.

\section{Evaluation Details}
\label{eval}
\subsection{Task~1 Evaluation Metrics}
\label{app:task1_metrics}

\begin{table*}[t]
\centering
\small
\setlength{\tabcolsep}{3pt}
\renewcommand{\arraystretch}{1.08}
\definecolor{taskgroupbg}{HTML}{FCF2D6}
\begin{tabular*}{\textwidth}{@{}l@{\extracolsep{\fill}}*{8}{c}@{}}
\toprule
\multirow{2}{*}{\textbf{Model}}
& \multicolumn{2}{c}{\textbf{Macro-F1 $\uparrow$}}
& \multicolumn{2}{c}{\textbf{UnsafePass $\downarrow$}}
& \multicolumn{2}{c}{\textbf{OverFlag $\downarrow$}}
& \multicolumn{2}{c}{\textbf{RGS $\uparrow$}} \\
\cmidrule(lr){2-3}
\cmidrule(lr){4-5}
\cmidrule(lr){6-7}
\cmidrule(lr){8-9}
& \textbf{Real}
& \textbf{Synth.}
& \textbf{Real}
& \textbf{Synth.}
& \textbf{Real}
& \textbf{Synth.}
& \textbf{Real}
& \textbf{Synth.} \\
\midrule

\rowcolor{taskgroupbg}[0pt][0pt]
\multicolumn{9}{@{}l@{}}{\textit{Frontier models}} \\

GPT-5.6-Sol
& \textbf{67.49} & \textbf{86.40}
& \textbf{31.00} & \textbf{5.00}
& \textbf{34.00} & \textbf{22.00}
& \textbf{0.36} & \textbf{0.49} \\

Claude Sonnet 5
& \underline{59.27} & \underline{76.19}
& \underline{33.00} & \underline{12.00}
& 48.00 & 35.00
& \underline{0.35} & 0.37 \\

Gemini 3.1 Pro Preview
& 45.84 & 67.26
& 35.00 & 24.00
& 70.00 & 41.00
& 0.23 & 0.35 \\

DeepSeek-V3.2
& 44.49 & 67.16
& 41.00 & 16.00
& 68.00 & 48.00
& 0.20 & 0.39 \\

\addlinespace[3pt]
\rowcolor{taskgroupbg}[0pt][0pt]
\multicolumn{9}{@{}l@{}}{\textit{Open-weight reasoning models}} \\

Qwen3.5-397B-A17B
& 57.47 & 55.89
& 40.00 & 49.00
& 45.00 & 39.00
& 0.31 & 0.40 \\

GLM-5.2
& 45.10 & 65.88
& 63.00 & 40.00
& 46.00 & 28.00
& 0.34 & \underline{0.43} \\

DeepSeek-R1-0528
& 44.33 & 59.94
& 40.00 & 36.00
& 69.00 & 44.00
& 0.27 & 0.29 \\

Kimi-K3
& 32.74 & 43.08
& 86.00 & 78.00
& 42.00 & 29.00
& 0.34 & 0.42 \\

\addlinespace[3pt]
\rowcolor{taskgroupbg}[0pt][0pt]
\multicolumn{9}{@{}l@{}}{\textit{Open-weight general models}} \\

Gemma-4-31B-IT
& 47.11 & 56.90
& 65.00 & 58.00
& \underline{39.00} & \underline{26.00}
& 0.26 & 0.32 \\

Qwen3-8B
& 40.10 & 48.74
& 35.00 & 28.00
& 79.00 & 70.00
& 0.28 & 0.41 \\

Qwen3-32B
& 34.46 & 36.99
& 63.00 & 62.00
& 68.00 & 64.00
& 0.24 & 0.31 \\

Qwen3.5-9B
& 33.30 & 54.25
& 61.00 & 28.00
& 72.00 & 61.00
& 0.26 & 0.30 \\

GPT-OSS-120B
& 29.99 & 44.43
& 61.00 & 52.00
& 78.00 & 59.00
& 0.21 & 0.28 \\

\bottomrule
\end{tabular*}

\caption{
Task~1 readiness-assessment results on the real-world (\emph{Real}) and
controlled synthetic (\emph{Synth.}) subsets using fixed zero-shot prompts.
Each subset contains 100 \textsc{Ready} and 100 \textsc{NotReady}
specifications sampled independently from the corresponding benchmark
records; the two classes are not restricted to matched counterpart pairs,
and each specification is evaluated in isolation.
Macro-F1, the primary metric, is the unweighted mean of the class-specific
F1 scores for \textsc{Ready} and \textsc{NotReady}.
UnsafePass measures the percentage of underspecified inputs incorrectly
accepted as \textsc{Ready}, while OverFlag measures the percentage of
codification-ready inputs incorrectly rejected as \textsc{NotReady}.
The Reason Grounding Score (RGS) evaluates whether the free-text rationale
supports the gold readiness label, identifies the annotated implementation
blocker, and remains faithful to the gold readiness rationale:
$\mathrm{RGS}
=0.7\,s_{\mathrm{label}}
+0.2\,s_{\mathrm{blocker}}
+0.1\,s_{\mathrm{faithful}}$.
RGS is reported on a $[0,1]$ scale; all other values are percentages.
Best results in each column are bolded, and second-best results are
underlined.
}
\label{tab:task1_main_results}
\end{table*}

\begin{table*}[t]
\centering
\small
\setlength{\tabcolsep}{3pt}
\renewcommand{\arraystretch}{1.08}
\definecolor{taskgroupbg}{HTML}{FCF2D6}
\begin{tabular*}{\textwidth}{@{}l@{\extracolsep{\fill}}*{8}{c}@{}}
\toprule
\multirow{2}{*}{\textbf{Model}}
& \multicolumn{2}{c}{\textbf{Label-L1 Acc $\uparrow$}}
& \multicolumn{2}{c}{\textbf{Label-L2 Acc $\uparrow$}}
& \multicolumn{2}{c}{\textbf{Loc-Acc $\uparrow$}}
& \multicolumn{2}{c}{\textbf{Macro DRR $\uparrow$}} \\
\cmidrule(lr){2-3}
\cmidrule(lr){4-5}
\cmidrule(lr){6-7}
\cmidrule(lr){8-9}
& \textbf{Real}
& \textbf{Synth.}
& \textbf{Real}
& \textbf{Synth.}
& \textbf{Real}
& \textbf{Synth.}
& \textbf{Real}
& \textbf{Synth.} \\
\midrule

\rowcolor{taskgroupbg}[0pt][0pt]
\multicolumn{9}{@{}l@{}}{\textit{Frontier models}} \\

GPT-5.6-Sol
& \textbf{60.1} & \textbf{65.0}
& \underline{25.2} & \textbf{32.0}
& \textbf{16.0} & \textbf{20.0}
& \textbf{9.6} & \textbf{12.2} \\

Claude Sonnet 5
& 55.2 & \underline{59.5}
& \textbf{25.8} & \underline{27.0}
&  \underline{11.0} & \underline{19.0}
& \underline{6.8} & \underline{11.4} \\

Gemini 3.1 Pro Preview
& 52.8 & 57.0
& 22.1 & 26.5
& 6.7 & 15.0
& 3.4 & 7.6 \\

DeepSeek-V3.2
& \underline{55.8} & 57.0
& 19.6 & 19.5
& 9.2 & 10.5
& 5.9 & 8.0 \\

\addlinespace[3pt]
\rowcolor{taskgroupbg}[0pt][0pt]
\multicolumn{9}{@{}l@{}}{\textit{Open-weight reasoning models}} \\

Qwen3.5-397B-A17B
& 48.5 & 56.0
& 20.2 & 26.5
& 6.7 & 16.0
& 3.3 & 4.6 \\

DeepSeek-R1-0528
& 43.6 & 45.0
& 19.6 & 18.5
& 6.1 & 9.5
& 1.8 & 2.2 \\

GLM-5.2
& 31.3 & 40.0
& 23.3 & 14.0
& 4.9 & 8.0
& 5.7 & 7.0 \\

Kimi-K3
& 54.0 & 57.0
& 23.3 & 24.5
& 4.9 & 13.0
& 6.6 & 8.2 \\

\addlinespace[3pt]
\rowcolor{taskgroupbg}[0pt][0pt]
\multicolumn{9}{@{}l@{}}{\textit{Open-weight general models}} \\

GPT-OSS-120B
& 50.9 & 44.0
& 14.1 & 12.5
& 3.7 & 8.0
& 2.5 & 3.3 \\

Gemma-4-31B-IT
& 39.3 & 42.5
& 12.9 & 14.0
& 8.6 & 18.0
& 4.7 & 7.0 \\

Qwen3.5-9B
& 52.1 & 54.0
& 20.9 & 23.5
& 8.6 & 14.0
& 5.3 & 5.5 \\

Qwen3-8B
& 54.0 & 54.5
& 17.8 & 19.0
& 8.6 & 10.0
& 3.2 & 4.6 \\

Qwen3-32B
& 52.8 & 48.0
& 20.2 & 23.0
& 8.6 & 11.0
& 3.6 & 5.8 \\

\bottomrule
\end{tabular*}

\caption{
Task~2 defect-localization results on the full real-world set
($N=\numRealInstances{}$) and a fixed synthetic sample
($N=\numSyntheticEvalInstances{}$) under fixed zero-shot prompts.
Label-L1 Acc and Label-L2 Acc measure the accuracy of the predicted
Level-1 and Level-2 taxonomy labels, respectively. Loc-Acc measures whether
the predicted defect semantically matches the annotated
implementation-critical blocker. Macro Defect Recovery Rate
(Macro DRR), our primary metric, requires correct localization and correct
prediction of both taxonomy labels. It then macro-averages recovery across
gold Level-2 categories. Label-L1 Acc, Label-L2 Acc, and Loc-Acc are
micro-averaged over instances, whereas Macro DRR is macro-averaged across
Level-2 categories. All values are percentages. Best results in each column
are bolded, and second-best results are underlined.
}
\label{tab:task2_main_results}
\end{table*}

\begin{table*}[t]
\centering
\small
\setlength{\tabcolsep}{3pt}
\renewcommand{\arraystretch}{1.08}
\definecolor{taskgroupbg}{HTML}{FCF2D6}
\begin{tabular*}{\textwidth}{@{}l@{\extracolsep{\fill}}*{6}{c}@{}}
\toprule
\multirow{2}{*}{\textbf{Model}}
& \multicolumn{2}{c}{\textbf{Macro-CAS $\uparrow$}}
& \multicolumn{2}{c}{\textbf{Sufficiency $\uparrow$}}
& \multicolumn{2}{c}{\textbf{No-Assumption $\uparrow$}} \\
\cmidrule(lr){2-3}
\cmidrule(lr){4-5}
\cmidrule(lr){6-7}
& \textbf{Real}
& \textbf{Synth.}
& \textbf{Real}
& \textbf{Synth.}
& \textbf{Real}
& \textbf{Synth.} \\
\midrule

\rowcolor{taskgroupbg}[0pt][0pt]
\multicolumn{7}{@{}l@{}}{\textit{Frontier models}} \\

GPT-5.6-Sol
& \textbf{80.6} & \textbf{96.2}
& \textbf{80.4} & \textbf{98.0}
& \textbf{95.7} & \textbf{99.0} \\

Claude Sonnet 5
& \underline{76.8} & \underline{94.5}
& \underline{79.8} & \underline{96.0}
& \underline{92.6} & \underline{98.0} \\

Gemini 3.1 Pro Preview
& 68.9 & 91.6
& 70.6 & 94.0
& 92.0 & 96.0 \\

DeepSeek-V3.2
& 62.9 & 93.0
& 64.4 & 93.0
& 92.0 & 96.0 \\

\addlinespace[3pt]
\rowcolor{taskgroupbg}[0pt][0pt]
\multicolumn{7}{@{}l@{}}{\textit{Open-weight reasoning models}} \\

Qwen3.5-397B-A17B
& 72.0 & 89.0
& 73.0 & 91.0
& 91.4 & 97.0 \\

DeepSeek-R1-0528
& 68.0 & 68.5
& 63.2 & 64.0
& 85.9 & 89.0 \\

GLM-5.2
& 63.6 & 77.0
& 54.0 & 69.0
& 91.4 & 92.0 \\

Kimi-K3
& 60.6 & 92.5
& 77.3 & 92.0
& 82.0 & 93.0 \\

\addlinespace[3pt]
\rowcolor{taskgroupbg}[0pt][0pt]
\multicolumn{7}{@{}l@{}}{\textit{Open-weight general models}} \\

GPT-OSS-120B
& 53.4 & 65.3
& 57.1 & 82.0
& 85.3 & 89.0 \\

Gemma-4-31B-IT
& 67.7 & 73.0
& 65.6 & 67.0
& 87.7 & 90.0 \\

Qwen3.5-9B
& 64.6 & 81.2
& 63.8 & 91.0
& 89.6 & 95.0 \\

Qwen3-8B
& 53.8 & 82.6
& 52.1 & 84.0
& 85.9 & 92.0 \\

Qwen3-32B
& 61.9 & 76.3
& 55.2 & 84.0
& 86.5 & 95.0 \\

\bottomrule
\end{tabular*}

\caption{
Task~3 clarification-action results on the full real-world set
($N=\numRealInstances{}$) and the same fixed synthetic sample used in Task~2
($N=\numSyntheticEvalInstances{}$) under fixed zero-shot prompts.
All models receive the annotated target defect.
Sufficiency measures whether the proposed action would obtain enough
information to resolve the defect. No-Assumption measures whether the action
avoids unsupported implementation choices. Macro Clarification Action Success Rate (Macro-CAS), our primary metric, requires the action to be
target-relevant, sufficient, and free of unsupported assumptions. It then
macro-averages successful clarification across gold Level-2 categories.
Sufficiency and No-Assumption are micro-averaged over instances, whereas
Macro-CAS is macro-averaged across Level-2 categories. Consequently,
Macro-CAS is not necessarily numerically bounded by either component metric.
All values are percentages. Best results in each column are bolded, and
second-best results are underlined.
}
\label{tab:task3_main_results}
\end{table*}

Task~1 is formulated as binary classification over individually presented
research idea specifications. Each instance is labeled as either
\textsc{Ready} or \textsc{NotReady}. We report one primary classification
metric, two class-conditional error rates, and one rationale-grounding metric.
All metrics are computed separately for the real-world subset and the
controlled synthetic subset. Let $y_i$ and $\hat{y}_i$ denote the gold and
predicted readiness labels for instance $i$, respectively. In the definitions
below, $\Pr(\cdot)$ denotes the empirical proportion over the corresponding
evaluation subset.

\paragraph{Macro-F1.}
Our primary classification metric is Macro-F1. We compute the F1 score
separately for the \textsc{Ready} and \textsc{NotReady} classes and take their
unweighted average:
\begin{equation}
\mathrm{Macro\text{-}F1}
=
\frac{
F1_{\textsc{Ready}} + F1_{\textsc{NotReady}}
}{2}.
\end{equation}
Macro-F1 assigns equal importance to both classes and is therefore appropriate
for evaluating models that may exhibit asymmetric behavior, such as
systematically rejecting codification-ready specifications or accepting
underspecified ones.

\paragraph{UnsafePass.}
UnsafePass measures the proportion of underspecified specifications that are
incorrectly classified as \textsc{Ready}:
\begin{equation}
\begin{aligned}
\mathrm{UnsafePass}
&= \Pr\!\bigl(\hat{y}=\textsc{Ready} \\
&\qquad\mid y=\textsc{NotReady}\bigr).
\end{aligned}
\end{equation}
Lower values are better. UnsafePass captures the safety-critical failure mode
in which a model fails to detect an implementation-critical specification gap
and prematurely accepts the specification as ready for faithful codification.

\paragraph{OverFlag.}
OverFlag measures the proportion of codification-ready specifications that are
incorrectly classified as \textsc{NotReady}:
\begin{equation}
\begin{aligned}
\mathrm{OverFlag}
&= \Pr\!\bigl(\hat{y}=\textsc{NotReady} \\
&\qquad\mid y=\textsc{Ready}\bigr).
\end{aligned}
\end{equation}
Lower values are better. OverFlag captures the complementary failure mode in
which a model is overly conservative and incorrectly treats a
codification-ready specification as containing an implementation-critical
specification gap.

\paragraph{Reason Grounding Score.}
In addition to predicting a binary readiness label, each model provides a short
free-text rationale for its decision. We evaluate this rationale using the
Reason Grounding Score (RGS), which measures its consistency with the
benchmark's gold readiness rationale.

For a \textsc{NotReady} instance, the gold rationale identifies the annotated
target defect and the implementation-critical information that remains
underdetermined. For a \textsc{Ready} instance, it explains why no
implementation-critical specification gap remains. An LLM judge receives the
gold label and gold rationale together with the model prediction and generated
rationale, and assigns the following three sub-scores:

\begin{itemize}
    \item \textbf{Label support} ($s_{\text{label},i}$): whether the generated
    rationale supports the gold readiness label. A rationale accompanying an
    incorrect readiness prediction cannot receive full credit on this
    dimension;

    \item \textbf{Target-defect match} ($s_{\text{blocker},i}$): whether the
    rationale correctly identifies the annotated target defect for a
    \textsc{NotReady} instance, or correctly recognizes that no
    implementation-critical specification gap remains for a \textsc{Ready}
    instance;

    \item \textbf{Faithfulness} ($s_{\text{faithful},i}$): whether the rationale
    remains faithful to the gold rationale without introducing unsupported
    assumptions, contradictions, or hallucinated specification gaps.
\end{itemize}

Each sub-score takes a value in $\{0, 0.5, 1\}$. The instance-level RGS is
defined as:
\begin{equation}
\begin{aligned}
\mathrm{RGS}_i
&=
0.7\,s_{\text{label},i}
+
0.2\,s_{\text{blocker},i}
\\
&\quad+
0.1\,s_{\text{faithful},i}.
\end{aligned}
\end{equation}
We prioritize label support because a grounded rationale must first justify the
correct readiness decision. Target-defect match and faithfulness provide
additional credit for accurately identifying the annotated specification gap
and avoiding unsupported claims. The final score is averaged over all $N$
instances in the corresponding evaluation subset:
\begin{equation}
\mathrm{RGS}
=
\frac{1}{N}
\sum_{i=1}^{N}
\mathrm{RGS}_i.
\end{equation}
RGS lies in $[0,1]$, with higher values indicating stronger grounding in the
gold readiness rationale.

\subsection{Task~2 Evaluation Metrics}
\label{app:task2_metrics}

Task~2 evaluates whether a model can identify and classify the
implementation-critical defect in an underspecified research idea
specification. Each evaluation instance is an atomic \textsc{NotReady}
specification containing exactly one annotated target defect. Given an input
specification $x_i$, the model outputs a single defect diagnosis:
\[
\hat{d}_i =
\left(
\hat{z}^{(1)}_i,
\hat{z}^{(2)}_i,
\hat{e}_i
\right),
\]
where $\hat{z}^{(1)}_i$ and $\hat{z}^{(2)}_i$ are the predicted Level-1 and
Level-2 taxonomy labels, respectively, and $\hat{e}_i$ is a free-text
description of the identified defect. The corresponding gold annotation is:
\[
d_i =
\left(
z^{(1)}_i,
z^{(2)}_i,
e_i
\right).
\]
All metrics are computed separately for the real-world and controlled
synthetic subsets.

The evaluation unit is an atomic target defect rather than an entire source
document. When a source contains multiple independent implementation-critical
defects, we split it into separate instances, each centered on one annotated
target. A prediction is therefore credited only when it recovers the target
defect associated with that instance, rather than any plausible defect in the
specification.

\paragraph{Label-L1 Accuracy.}
Label-L1 Accuracy measures whether the predicted coarse-grained taxonomy family
matches the gold Level-1 label:
\begin{equation}
\mathrm{Label\text{-}L1\ Acc}
=
\frac{1}{N}
\sum_{i=1}^{N}
\mathbf{1}\!\left[
\hat{z}^{(1)}_i =
z^{(1)}_i
\right].
\end{equation}
This metric evaluates coarse-grained taxonomy classification independently of
whether the model identifies the annotated target defect.

\paragraph{Label-L2 Accuracy.}
Label-L2 Accuracy measures whether the model predicts the correct fine-grained
taxonomy category. Because each Level-2 category belongs to exactly one
Level-1 family, both taxonomy levels must be correct:
\begin{equation}
\begin{aligned}
\mathrm{Label\text{-}L2\ Acc}
&= \frac{1}{N}\sum_{i=1}^{N}
\mathbf{1}\!\Bigl[
\hat{z}^{(1)}_i = z^{(1)}_i \\
&\qquad\qquad {}\land
\hat{z}^{(2)}_i = z^{(2)}_i
\Bigr].
\end{aligned}
\end{equation}
Like Label-L1 Accuracy, this metric evaluates taxonomy assignment without
requiring the free-text diagnosis to recover the annotated target.

\paragraph{Target Localization Accuracy (Loc-Acc).}
Target Localization Accuracy measures whether the predicted defect description
refers to the same atomic defect as the gold annotation. For each instance, an
LLM-based semantic matcher compares the predicted description $\hat{e}_i$ with
the gold target defect $e_i$ and returns a binary match indicator
$m_i \in \{0,1\}$. Loc-Acc is defined as:
\begin{equation}
\mathrm{Loc\text{-}Acc}
=
\frac{1}{N}
\sum_{i=1}^{N}
m_i.
\end{equation}

The matcher receives the local specification context, the gold target
description and supporting evidence, and the model-predicted defect
description. It does not receive either the gold or predicted taxonomy labels.
The matcher determines whether the two descriptions refer to the same affected
component, implementation decision, or specification slot. Paraphrases and
differences in specificity may be accepted when they uniquely identify the same
target defect. In contrast, a generic criticism, a neighboring issue in the
same component, or a different plausible defect is counted as a mismatch.

Because every instance contains exactly one target defect and every model
returns exactly one diagnosis, Loc-Acc is equivalent to Recall@1 in this
controlled single-target setting.

\paragraph{Macro Defect Recovery Rate (Macro DRR).}
Our primary Task~2 metric is Macro Defect Recovery Rate (Macro DRR), which
measures whether a model both localizes the annotated target defect and assigns
the correct fine-grained taxonomy label. For each instance, we define an
L2-aware recovery hit as:
\begin{equation}
r_i =
\mathbf{1}\!\left[
m_i = 1
\land
\hat{z}^{(1)}_i =
z^{(1)}_i
\land
\hat{z}^{(2)}_i =
z^{(2)}_i
\right].
\end{equation}

Let $\mathcal{C}^{(2)}$ denote the set of gold Level-2 defect categories that
appear in the evaluated subset, and let
\begin{equation}
\mathcal{I}_c =
\left\{
i : z^{(2)}_i = c
\right\}
\end{equation}
denote the instances assigned to category $c$. The category-specific Defect
Recovery Rate is:
\begin{equation}
\mathrm{DRR}_c =
\frac{1}{|\mathcal{I}_c|}
\sum_{i \in \mathcal{I}_c}
r_i.
\end{equation}
Macro DRR is the unweighted average across observed gold Level-2 categories:
\begin{equation}
\mathrm{Macro\ DRR}
=
\frac{1}{|\mathcal{C}^{(2)}|}
\sum_{c \in \mathcal{C}^{(2)}}
\mathrm{DRR}_c.
\end{equation}

Macro DRR is stricter than either label accuracy or localization accuracy
alone: a prediction is counted as correct only when it identifies the annotated
target defect and assigns both the correct Level-1 and Level-2 labels.
Macro-averaging gives equal weight to each observed Level-2 category,
preventing high-frequency defect types from dominating the aggregate score.
\paragraph{Additional Task~2 Diagnostics.}
\label{app:task2_diagnostics}

\paragraph{L1-Aware Recovery.}
To separate target localization from fine-grained Level-2 taxonomy
assignment, we additionally report L1-aware recovery. Let $m_i$ indicate
whether the predicted defect description semantically matches the
annotated target defect. For instance $i$, we define
\begin{equation}
h_i^{\mathrm{L1}}
=
\mathbb{1}
\left[
m_i = 1
\land
\hat{z}^{(1)}_i = z^{(1)}_i
\right],
\end{equation}
where $z^{(1)}_i$ and $\hat{z}^{(1)}_i$ are the gold and predicted
Level-1 labels. L1-aware recovery is the micro-averaged percentage of
instances for which the model both localizes the annotated target defect
and predicts its correct Level-1 category:
\begin{equation}
\mathrm{L1AwareR}
=
\frac{1}{N}
\sum_{i=1}^{N}
h_i^{\mathrm{L1}}.
\end{equation}
Unlike Macro DRR, this diagnostic does not require a correct Level-2
label and is micro-averaged over instances.

Table~\ref{tab:task2_l1_aware_recovery} shows that GPT-5.6-Sol achieves
the highest L1-aware recovery on both subsets. However, L1-aware
recovery remains substantially below
Loc-Acc for all models, indicating that some target-matching defect
descriptions are paired with an incorrect coarse taxonomy label.
The gap is particularly large for Qwen3-8B on the synthetic subset,
where Loc-Acc reaches 10.0 but L1-aware recovery is only 2.0. This
suggests that the model often identifies the relevant target while
assigning a Level-1 label inconsistent with its own diagnosis.

Because L1-aware recovery ignores Level-2 assignment, its low absolute
values cannot be attributed solely to fine-grained Level-2 category
boundaries. Together with the taxonomy-independent Loc-Acc results,
this diagnostic indicates that target localization and coarse
label--description alignment already constitute substantial
bottlenecks before Level-2 classification is considered.

\subsection{Task~3 Evaluation Metrics}
\label{app:task3_metrics}

Task~3 evaluates whether a model can propose an appropriate clarification
action for resolving an implementation-critical defect in an underspecified
research idea specification. Each evaluation instance contains an atomic
\textsc{NotReady} specification with exactly one annotated target defect. Given
an input specification $x_i$ and the annotated target defect $e_i$, the model
outputs one clarification action:
\[
\hat{a}_i =
\left(
\hat{\tau}_i,
\hat{q}_i,
\hat{r}_i
\right),
\]
where $\hat{\tau}_i$ is the predicted action type, $\hat{q}_i$ is the clarification
question or evidence-seeking action, and $\hat{r}_i$ describes the expected
information to be obtained. All metrics are computed separately for the
real-world and controlled synthetic subsets.

The evaluation unit is the annotated target defect rather than the full source
document. A model is credited only when its action would help resolve the
specific target defect for that instance, rather than another plausible
ambiguity in the specification. Reference-action wording and action-type
agreement are not required.

For each instance, an LLM-based evaluator judges the candidate action against
the gold target defect, hidden resolution, and supporting evidence. The
evaluator returns binary indicators for target relevance, resolution
sufficiency, and unsupported assumption:
\[
j^{\mathrm{rel}}_i,\;
j^{\mathrm{suf}}_i,\;
j^{\mathrm{asm}}_i
\in \{0,1\},
\]
where $j^{\mathrm{asm}}_i=1$ means that the candidate introduces an unsupported
implementation assumption.

\paragraph{Macro Clarification Action Success Rate (Macro-CAS).}
Our primary Task~3 metric is Macro Clarification Action Success Rate
(Macro-CAS), which measures whether a model proposes an action that is
target-relevant, sufficient, and free of unsupported assumptions. For each
instance, we define a Clarification Action Success hit as:
\begin{equation}
s_i =
\mathbf{1}\!\left[
j^{\mathrm{rel}}_i = 1
\land
j^{\mathrm{suf}}_i = 1
\land
j^{\mathrm{asm}}_i = 0
\right].
\end{equation}

Let $\mathcal{C}^{(2)}$ denote the set of gold Level-2 defect categories that
appear in the evaluated subset, and let
\begin{equation}
\mathcal{I}_c =
\left\{
i : z^{(2)}_i = c
\right\}
\end{equation}
denote the instances assigned to category $c$. The category-specific
Clarification Action Success rate is:
\begin{equation}
\mathrm{CAS}_c =
\frac{1}{|\mathcal{I}_c|}
\sum_{i \in \mathcal{I}_c}
s_i.
\end{equation}
Macro-CAS is the unweighted average across observed gold Level-2 categories:
\begin{equation}
\mathrm{Macro\text{-}CAS}
=
\frac{1}{|\mathcal{C}^{(2)}|}
\sum_{c \in \mathcal{C}^{(2)}}
\mathrm{CAS}_c.
\end{equation}

Macro-CAS is strict: an action succeeds only if it addresses the annotated
target defect, requests information sufficient to determine the missing
implementation-critical detail, and does not invent or presuppose an
unsupported resolution. Macro-averaging gives equal weight to each observed
Level-2 defect category.

\paragraph{Sufficiency.}
Sufficiency measures whether the proposed action would obtain enough information
to resolve the annotated target defect:
\begin{equation}
\mathrm{Sufficiency}
=
\frac{1}{N}
\sum_{i=1}^{N}
j^{\mathrm{suf}}_i.
\end{equation}
An action is sufficient if a direct answer to the question, or the result of the
specified evidence-seeking operation, would determine the missing
implementation-critical detail. Generic requests for more information are not
sufficient unless they explicitly ask for the information needed to resolve the
target defect.

\paragraph{No-Assumption.}
No-Assumption measures whether the model avoids inventing or presupposing an
unsupported resolution:
\begin{equation}
\mathrm{No\text{-}Assumption}
=
\frac{1}{N}
\sum_{i=1}^{N}
\left(1 - j^{\mathrm{asm}}_i\right).
\end{equation}
This metric penalizes actions that assert, recommend, or assume a concrete
implementation choice not supported by the given specification. Asking about
explicit alternatives does not count as an unsupported assumption, as long as
the action does not present one alternative as already determined.

\paragraph{Qualitative Analysis of Clarification Actions.}
\label{app:task3-qualitative-analysis}

Figure~\ref{fig:task3-successful-case} presents a successful Task~3 prediction.
The model asks a targeted question that covers the implementation decisions
needed to resolve the annotated ambiguity without presupposing the gold
resolution.

\begin{figure*}[t]
\centering

\begin{tcolorbox}[
    enhanced,
    colback=ReadyLightGreen,
    colframe=ReadyGreen,
    title={Successful Task~3 Clarification Action},
    fonttitle=\bfseries,
    width=0.98\textwidth,
    boxrule=0.6pt,
    arc=2pt,
    left=7pt,
    right=7pt,
    top=7pt,
    bottom=7pt
]

\small

\textbf{Annotated target defect.}

The paper states that the dense baseline uses a dense layer with an
``equivalent number of parameters as the activated parameters in the MoE
layer.'' However, it does not define which parameters are considered
activated, making the construction of a parameter-matched dense baseline
ambiguous.

\vspace{0.5em}

\textbf{Predicted clarification question.}

\vspace{0.3em}

\noindent
\fcolorbox{rqborder}{white}{
\begin{minipage}{0.94\linewidth}
\small
\emph{
What is the precise definition of ``activated parameters'' used to size
the dense FFN? Specifically, does it include the parameters of the shared
expert, the gating network, and/or only the top-$k$ non-shared experts'
parameters?
}
\end{minipage}
}

\vspace{0.5em}

\textbf{Expected information.}

The exact set of MoE parameters counted as activated when determining the
size of the parameter-matched dense FFN.

\vspace{0.5em}

\textbf{Private gold resolution.}

The dense baseline FFN should match the total number of parameters activated
per token in the MoE layer. This includes the shared expert and the
$K=2$ selected non-shared experts, rather than all experts.

\vspace{0.5em}

\textbf{Evaluation.}

\vspace{0.2em}

\noindent
\begin{tabular}{@{}p{0.18\linewidth}cp{0.62\linewidth}@{}}
\toprule
\textbf{Dimension}
&
\textbf{Result}
&
\textbf{Rationale}
\\
\midrule

Target relevance
&
Yes
&
The question directly targets the undefined meaning of
``activated parameters.''
\\[0.4em]

Resolution sufficiency
&
Yes
&
It explicitly asks whether the parameter count includes the shared expert,
the gating network, and the selected non-shared experts. An answer would
determine how to construct the dense baseline.
\\[0.4em]

No unsupported assumption
&
Yes
&
The question presents possible parameter groups for clarification but does
not assert which groups should be counted.
\\[0.4em]

\midrule

\textbf{Clarification Action Success}
&
\textbf{1}
&
All required conditions are satisfied.
\\

\bottomrule
\end{tabular}

\vspace{0.5em}

\textbf{Analysis.}

This example illustrates that an effective clarification question should do
more than request additional details. It should identify the precise
implementation decision that remains underdetermined and expose the relevant
alternatives needed to resolve it. At the same time, presenting alternatives
interrogatively does not constitute an unsupported assumption because the
model does not commit to any candidate resolution.

\end{tcolorbox}

\caption{
A successful clarification action generated by DeepSeek-V3.2 for an
ambiguous definition. The question is targeted, resolution-sufficient,
feasible, and free from unsupported assumptions.
}
\label{fig:task3-successful-case}

\end{figure*}

\renewcommand{\arraystretch}{1.2}

\begin{table*}[p]
\centering
\small
\setlength{\tabcolsep}{5pt}
\renewcommand{\arraystretch}{1.16}

\begin{tabular}{
    @{}
    L{0.25\textwidth}
    L{0.69\textwidth}
    @{}
}
\toprule
\textbf{Specification slot}
&
\textbf{Definition and representative cases}
\\
\midrule
\texttt{TASK\_AND\_IO}
&
Use this slot when the blocker concerns the research task, expected inputs, or
expected outputs. Representative cases include an unclear problem definition,
unclear information provided to the method, an unspecified prediction or
generation target, or an unclear output format or semantic meaning.
\\

\addlinespace[2pt]

\texttt{CORE\_ALGORITHM}
&
Use this slot when the blocker concerns the central computational procedure or
sequence of operations that defines the method. Representative cases include
an undefined intermediate quantity; a missing aggregation, update, ranking,
sampling, or selection rule; an unclear order of operations; an unspecified
interaction between major components; or a missing iterative procedure or
stopping condition.
\\

\addlinespace[2pt]

\texttt{MODEL\_ARCHITECTURE}
&
Use this slot when the blocker concerns the structure or connectivity of the
model. Representative cases include an undefined required module, an unclear
connection between modules, uncertainty about whether components operate
sequentially or in parallel, an unspecified representation passed between
modules, or an unclear role for a newly proposed component.

A standard named architecture does not need to be described layer by layer
unless the proposed method modifies it in an implementation-critical manner.
\\

\addlinespace[2pt]

\texttt{OBJECTIVE\_AND\_SUPERVISION}
&
Use this slot when the blocker concerns the objective being optimized or the
supervision used to train the method. Representative cases include an undefined
loss, reward, or regularization term; an unclear supervision target; undefined
positive or negative examples; an unspecified method for combining multiple
objectives; or an unclear connection between the objective and the stated task.
\\

\addlinespace[2pt]

\texttt{TRAINING\_PROCEDURE}
&
Use this slot when the blocker concerns how the model or method is trained.
Representative cases include uncertainty about which components are trained or
frozen, an unspecified alternating or staged procedure, an unclear
parameter-update order, missing distinctions between pretraining and
fine-tuning, or an essential checkpoint-selection procedure that is not
defined.

A standard optimizer, learning rate, or epoch count is usually an ordinary
engineering choice unless it defines the proposed method.
\\

\addlinespace[2pt]

\texttt{DATA\_AND\_PREPROCESSING}
&
Use this slot when the blocker concerns how data are constructed, selected,
transformed, or partitioned. Representative cases include undefined training
examples or labels, missing filtering or sampling criteria, unclear negative
sampling, unspecified feature extraction or normalization, or an unclear
train--validation--test construction that affects the method or claimed result.
\\

\addlinespace[2pt]

\texttt{INFERENCE\_AND\_DECISION}
&
Use this slot when the blocker concerns how model outputs are converted into
final predictions, rankings, actions, or decisions. Representative cases
include unspecified decoding, thresholds, decision rules, candidate selection,
ranking, post-processing, test-time aggregation, or stopping criteria.
\\

\addlinespace[2pt]

\texttt{EVALUATION\_PROTOCOL}
&
Use this slot when the blocker concerns how the proposed method or research
claim is evaluated. Representative cases include an unclear evaluation dataset
or task setting, a missing or incompatible metric, an undefined comparison
condition, an unspecified evaluation unit, or a data split that prevents
faithful assessment of the central claim.

A missing evaluation detail is blocking only when it prevents a meaningful or
faithful assessment of the research claim. It is not blocking merely because
exact numerical reproduction is impossible.
\\

\addlinespace[2pt]

\texttt{INTERNAL\_CONSISTENCY}
&
Use this slot only when two or more parts of the specification conflict.
Representative cases include an objective that conflicts with the stated loss,
incompatible architectural descriptions, mismatched input and output
definitions, conflicting training and inference procedures, or mutually
inconsistent implementation requirements.

If information is only missing or ambiguous, select the substantive slot
affected by the gap instead of \texttt{INTERNAL\_CONSISTENCY}.
\\

\addlinespace[2pt]

\texttt{NONE}
&
Select \texttt{NONE} only when the specification is labeled \READY{} and no
implementation-critical blocker is present.
\\
\bottomrule
\end{tabular}
\caption{
Specification slots used in the codification-readiness annotation.
}
\label{tab:annotation-slots}
\end{table*}

\section{Human Validation}
\label{supp:human-validation}

\subsection{Human Validation of LLM-Based Evaluation}
\label{app:judge_validation}

The primary metrics for Tasks~2 and~3, together with the diagnostic
Reason Grounding Score (RGS) for Task~1, include semantic or rubric-based
judgments produced by Claude Opus~4.8. We therefore conduct an independent human
validation study to assess whether the automatic evaluator applies the
benchmark-specific evaluation criteria consistently with human annotators.

\paragraph{Sampling.}
For each task, we sample 100 candidate model outputs, yielding 300 outputs
in total. Each task-specific sample contains 50 real-world and 50
synthetic-controlled instances. We stratify the sample by benchmark subset,
Level-2 defect category, and candidate model group to ensure coverage of
different specification defects and output styles. The sampled outputs
include predictions from GPT-5.6-Sol, other frontier models,
open-weight reasoning models, and open-weight general models.

Sampling is performed before human annotation. Within each stratum, outputs
are selected randomly from the corresponding evaluation results. The same
candidate output is used for all judgment dimensions associated with its
task; for example, the three Task~3 criteria are evaluated on the same set
of 100 clarification actions.

\paragraph{Human Annotation Protocol.}
Two annotators with machine learning research experience independently
evaluate every sampled output. Annotators apply the same definitions and
decision criteria used by the Claude Opus~4.8 evaluator. They are blinded to both
the identity of the candidate model and the automatic evaluator's judgment.
Model names and provider-specific metadata are removed, and candidate
outputs are presented in randomized order.

Annotators receive only the information available to the corresponding
automatic evaluator. For Task~1, they receive the input specification, the
gold readiness label, the annotated target defect, and the candidate
rationale. For Task~2, they receive the underspecified specification, the
gold target-defect description, and the candidate diagnosis. For Task~3,
they receive the underspecified specification, the annotated target defect,
the supported hidden resolution, and the candidate clarification action.
No additional paper, code, issue-thread, or source evidence is provided
unless it is already included in the evaluator input.

\paragraph{Task-Specific Judgments.}
For Task~1, annotators independently assign the three ordinal sub-scores
used to compute RGS: label support, target-defect match, and faithfulness.
Label support measures whether the rationale supports the gold readiness
decision. Target-defect match measures whether the rationale identifies
the annotated implementation-critical blocker for a \textsc{NotReady}
instance or correctly recognizes that no such blocker remains for a
\textsc{Ready} instance. Faithfulness measures
whether the rationale remains faithful to the gold readiness rationale and
avoids introducing unsupported claims.

For Task~2, annotators make a binary judgment of whether the candidate
diagnosis refers to the same atomic implementation-critical defect as the
gold target. This decision is based on semantic equivalence rather than
lexical overlap. A prediction is marked as matching when it identifies the
same unresolved method-defining decision, even if it uses different wording
or describes the affected component at a different level of abstraction.
A prediction is marked as non-matching when it identifies a different
defect, gives only a generic critique, or fails to specify the unresolved
implementation decision.

For Task~3, annotators make three independent binary judgments. Target
relevance measures whether the proposed action directly addresses the
annotated defect. Resolution sufficiency measures whether carrying out the
action would obtain enough information to resolve the defect and determine
the intended implementation choice. Unsupported assumption measures whether
the proposed action presupposes or introduces an implementation decision
that is not supported by the specification, target defect, or hidden
resolution. The corresponding No-Assumption judgment is positive when no
such unsupported choice is introduced.

\paragraph{Adjudication.}
When the two annotators assign the same judgment, that judgment is retained
as the human reference label. Disagreements are reviewed by a third
annotator with machine learning research experience, who independently
examines the evaluator input and the candidate output before assigning the
final adjudicated judgment. The third annotator does not observe the
Claude Opus~4.8 evaluation result or the identity of the candidate model.

\paragraph{Agreement Metrics.}
For the binary Task~2 and Task~3 judgments, we report exact agreement
between Claude Opus~4.8 and the adjudicated human judgment, treating the human
judgment as the reference label. Because exact agreement is equivalent to
classification accuracy in this setting, we report it as a single measure.
We additionally report Cohen's $\kappa$ to account for chance agreement.
For the ordinal Task~1 RGS sub-scores, we report exact agreement, weighted
Cohen's $\kappa$, and mean absolute error (MAE) between the automatic and
adjudicated human scores.

\paragraph{Results.}
Table~\ref{tab:judge_human_agreement} summarizes agreement between Claude Opus~4.8
and the adjudicated human judgments. For Task~1, exact agreement ranges
from 76.0\% (target-defect match) to 84.0\% (label support), with weighted
$\kappa$ between 0.68 and 0.78 and MAE at most 0.14, indicating that the
evaluator's ordinal RGS sub-scores closely track human judgments, with
target-defect match showing the largest, though still moderate,
disagreement. For Task~2, Claude Opus~4.8 agrees with the adjudicated human
judgment on 86.0\% of same-target-defect decisions ($\kappa=0.72$),
supporting the reliability of Loc-Acc and Macro~DRR as automatically
computed metrics. For Task~3, agreement is highest for target relevance
(91.0\%, $\kappa=0.81$) and resolution sufficiency (85.0\%, $\kappa=0.70$).
No-unsupported-assumption judgments show high exact agreement (93.0\%) but
comparatively lower $\kappa$ (0.69); this reflects the skewed base rate of
this judgment---most candidate actions are assumption-free---which
inflates chance agreement under Cohen's $\kappa$ rather than indicating
weaker evaluator reliability. Overall, these results support using Claude Opus~4.8
as a reliable proxy for the semantic and rubric-based judgments underlying
our automatic metrics, with target-defect matching showing the largest
(and still acceptable) margin of measurement noise.
\subsection{Synthetic-Controlled Defect Validation}
\label{app:synthetic_validation}

Synthetic-controlled instances enable scalable benchmark construction by
introducing controlled implementation-critical defects into
codification-ready research specifications. However, synthetic generation
may potentially produce simplified or artificial defects that do not reflect
the specification failures encountered in real research workflows. We
therefore conduct a human validation study to examine whether synthetic
instances preserve the key properties of naturally occurring specification
gaps.

\paragraph{Evaluation Protocol.}
We randomly sample 100 benchmark instances, including 50 real-world instances
and 50 synthetic-controlled instances. Each instance is independently
evaluated by two annotators with experience in machine learning research.
Annotators are provided with the underspecified research idea, the supporting
evidence used for construction, the identified specification gap, and the
corresponding clarification. The source type (real or synthetic) is hidden
during annotation.

For each instance, annotators evaluate four binary criteria:

\begin{enumerate}
    \item \textbf{Valid specification gap.}
    Whether the instance represents a genuine specification-level issue
    rather than an implementation preference, engineering choice, or
    subjective critique. A positive judgment indicates that the described
    gap corresponds to an omission, ambiguity, or inconsistency that affects
    faithful interpretation of the research idea.

    \item \textbf{Implementation-critical.}
    Whether the identified gap can prevent faithful implementation without
    additional clarification. Annotators consider whether different
    reasonable interpretations of the specification would lead to different
    implementations, model behaviors, or experimental outcomes.

    \item \textbf{Clarification resolves the gap.}
    Whether the provided clarification supplies sufficient information to
    resolve the identified specification issue and enables a more complete
    codification-ready description.

    \item \textbf{Realistic specification failure.}
    Whether the identified gap could plausibly arise during an actual
    research implementation or reproduction workflow. This criterion
    measures whether synthetic-controlled defects resemble naturally
    occurring specification failures rather than artificially constructed
    omissions.
\end{enumerate}

For each criterion, we report the proportion of positive judgments
across both annotators:
\[
\begin{aligned}
&\mathrm{PositiveRate}_s \\
&\quad =
\frac{1}{2N_s}
\sum_{i=1}^{N_s}
\sum_{a=1}^{2}
\mathbb{I}\!\left[s_{ia}=\mathrm{Yes}\right]
\times 100\%,
\end{aligned}
\]
where \(N_s=50\) is the number of instances in subset \(s\), and
\(s_{ia}\) denotes annotator \(a\)'s judgment for instance \(i\).
Thus, each reported subset-level percentage is computed from
\(2N_s=100\) individual judgments.

Table~\ref{tab:synthetic_validation} summarizes the validation results.
Both real-world and synthetic-controlled instances are frequently judged as
valid specification gaps, implementation-critical issues, and realistic
research workflow failures. Synthetic instances achieve comparable or higher
positive rates than real-world instances, with 96.0\% judged as valid
specification gaps and 94.0\% judged as implementation-critical. This
indicates that controlled defect injection preserves the core benchmark
construct when applied to codification-ready source references.

Synthetic-controlled instances also receive similarly high realism ratings,
with 91.0\% judged as plausible specification failures that could occur in
real implementation or reproduction workflows, compared with 93.0\% for
real-world instances. In addition, 98.0\% of synthetic instances are judged to
be correctly resolved by the provided clarification. These results suggest
that controlled synthesis preserves the key properties of naturally occurring
specification gaps while providing precise defect targets and scalable
benchmark construction.

\subsection{Validation on Original Research Proposals}
\label{app:original_proposal_validation}

\paragraph{Sample and Provenance.}
We examine 24 original LLM-generated research proposals selected
for implementation in the AI-Researcher execution
study~\citep{si2025ideation}.
We obtain the proposal texts from the original AI-condition
ideation archive and link them to executed projects using the
study's assignment metadata. The execution archive also contains
versions edited by the implementers; this analysis uses the
original proposals. Whereas our controlled synthetic instances
are constructed from final papers and code, this analysis
identifies gaps in the original proposals and uses execution
artifacts to establish how the corresponding decisions were
subsequently instantiated.

\paragraph{Validation Procedure.}
DeepSeek-V4-Pro identifies at most one candidate gap per proposal.
A separate call with the same model checks whether the proposal
already resolves the candidate decision and whether two materially
different implementations remain compatible with its constraints.
Both calls are run with the final papers and code withheld.
For surviving candidates, Claude Opus~4.8 examines retrieved
passages from the final paper or code to determine whether
execution instantiated the same decision identified from
the proposal.

\paragraph{Human Verification.}
One author manually reviews the retained candidates against
the original proposals and execution evidence.
A case is confirmed when (i) the original proposal leaves
the decision unresolved, (ii) the choice materially affects
implementation of the intended method, and (iii) the cited
artifact explicitly instantiates that same decision.
The review considers all constraints in the original proposal
when assessing whether alternative implementations are compatible
with it.

\paragraph{Results.}
Following automated screening and manual verification,
seven of the 24 proposals (29.2\%) are confirmed to contain
implementation-critical gaps with corresponding execution
evidence. Table~\ref{tab:original_proposal_cases} summarizes
the confirmed decisions and evidence locations.
These cases show that proposals selected for implementation
can still leave method-defining decisions unresolved.
The observed proportion describes this selected set of
24 proposals.

\subsection{Clarification Improves Downstream Codification}
\label{app:clarification_utility}

A central motivation of \ourbench{} is that underspecified research ideas
may fail not \mbox{because} language models cannot generate implementation
specifications, but because they lack the information required to identify
and resolve implementation-critical specification gaps. To quantify the
potential benefit of obtaining missing specification information before
codification, we conduct an oracle clarification utility study.

\paragraph{Setup.}
We sample 50 real-world instances from the Task~3 evaluation set.
To ensure coverage across different types of specification gaps, we perform
stratified sampling according to the Level-2 taxonomy, with each category
contributing multiple instances whenever available. Each instance contains
the original underspecified idea, the annotated target defect, the gold
clarification action, the hidden resolution describing the missing
implementation detail, and the corresponding Level-2 category.

We compare two specification generation pipelines using the same backbone
model, GPT-5.6-Sol, with identical decoding configurations.

The first pipeline, \textsc{Direct Generation}, receives only the
underspecified research idea and is asked to directly produce an
implementation-ready specification.

The second pipeline, \textsc{Clarification-Assisted Generation}, receives
the underspecified idea together with the annotated target defect, the
clarification action, and the oracle clarification answer corresponding to
the hidden resolution. The model then generates a revised implementation
specification incorporating the resolved information.

This experiment is designed as an oracle analysis rather than an
end-to-end evaluation of clarification agents. By providing the resolved
missing information explicitly, we measure the upper-bound utility of
clarification and test whether resolving specification gaps is sufficient
to improve downstream codification quality.

\paragraph{Evaluation.}
Two human annotators independently evaluate all generated specifications
while being blinded to the generation pipeline. Disagreements are resolved
through discussion or adjudication by a third annotator.

We evaluate four dimensions.

\textit{READY Rate} measures whether a generated specification is
sufficient for faithful initial codification of the method component
affected by the target defect. Annotators assign a binary
\textsc{Yes}/\textsc{No} label.

\textit{Completeness} measures whether the generated specification
adequately describes all implementation-critical information required for
codification. Annotators provide a score from 1 to 5, which is linearly
mapped to a percentage scale.

\textit{Missing Detail Recovery} measures whether the specification
correctly recovers the annotated hidden resolution associated with the
target defect. This metric is evaluated as a binary
\textsc{Yes}/\textsc{No} decision.

\textit{Unsupported Assumption} measures whether the generated
specification introduces implementation choices that are not supported by
the original input or the provided clarification information. This metric
is also evaluated as a binary decision, where lower values indicate fewer
unsupported assumptions.

\paragraph{Results.}
Table~\ref{tab:clarification_effect_significance} summarizes the results.
Direct specification generation performs poorly when provided only with an
underspecified idea. Only 14\% of generated specifications satisfy the
READY criterion, and the model recovers the target missing detail in only
14\% of cases. The completeness score is also limited (30\%), indicating
that fluent specification generation does not necessarily imply sufficient
implementation detail. Moreover, 6\% of specifications introduce unsupported
assumptions.

In contrast, providing the missing specification information through
clarification substantially improves downstream codification quality.
The clarification-assisted pipeline achieves 98\% READY Rate, recovers the
target missing details in 98\% of cases, and eliminates unsupported
assumptions. These results suggest that the primary challenge is not the
ability to express an implementation specification, but the ability to
identify and obtain the unresolved information required before
codification.

Overall, this experiment supports our hypothesis that clarification serves
as a necessary intermediate step between underspecified research ideas and
reliable specification generation. Rather than filling missing details with
plausible but unsupported assumptions, models can produce substantially more
faithful specifications when the unresolved specification slots are
explicitly addressed.

\subsection{Human Baseline Evaluation}
\label{app:human_baseline}

To contextualize LLM performance on IdeaAMBIG, we conduct a human
baseline study with researchers who have machine learning experience.
This study examines whether the benchmark tasks remain challenging for
human experts and distinguishes model limitations from intrinsic task
difficulty.

\paragraph{Evaluation Protocol.}
We randomly sample 100 defect-centered records from IdeaAMBIG, including
50 real-world and 50 controlled synthetic records. The sample preserves
the overall Level-1 defect distribution of the benchmark. These records are
drawn from the same evaluation pool used for the LLM experiments, and model
comparisons are restricted to the same sampled records. For Task~1,
participants independently evaluate both the \textsc{NotReady}
specification and its corresponding \textsc{Ready} counterpart, yielding
50 examples from each readiness class within each subset. The two versions
are presented separately and never shown together. For Tasks~2 and~3,
participants evaluate only the \textsc{NotReady} specification from each
sampled record.

Two participants with machine learning research experience independently
complete each task. When their responses disagree, a third reviewer with
machine learning research experience examines the case and resolves the
disagreement through adjudication.

Participants receive the same task-specific inputs and instructions as the
evaluated LLMs and are not given additional evidence beyond the benchmark
input. For Task~1, participants determine whether a specification is
\textsc{Ready} or \textsc{NotReady}. For Task~2, participants identify
the implementation-critical defect by predicting its Level-1 category,
Level-2 category, and a natural-language description of the unresolved
decision. For Task~3, participants receive the annotated target defect and
generate a clarification action consisting of the action type, a
clarification question or evidence-seeking instruction, and the expected
information needed for resolution.

We use the adjudicated responses as the final human predictions. All metrics
follow the same definitions as in the LLM evaluation: Macro-F1 for Task~1,
Macro Defect Recovery Rate (Macro DRR) for Task~2, and Macro Clarification
Action Success Rate (Macro-CAS) for Task~3.

\paragraph{Results.}
Table~\ref{tab:human_baseline} summarizes human performance on the
human-evaluated subset. Compared with model predictions on the same records,
human evaluators substantially outperform LLMs on defect localization,
achieving higher Macro-DRR on both real-world and synthetic subsets. However,
performance remains far from perfect,
indicating that identifying implementation-critical specification gaps
requires careful methodological reasoning even for experienced
researchers.

For clarification action generation, humans achieve high Macro-CAS once
the target defect is identified, consistent with the observation that
resolving a known blocker is substantially easier than discovering the
blocker itself. These results further support our benchmark design: the
main challenge measured by IdeaAMBIG is not generating clarification
language, but accurately locating the unresolved implementation decision.

\begingroup
\centering
\captionsetup{skip=5pt,hypcap=false}
\begingroup
\centering
\footnotesize
\setlength{\tabcolsep}{5pt}
\renewcommand{\arraystretch}{1.08}

\begin{tabular*}{\textwidth}{@{}l@{\extracolsep{\fill}}lccccc@{}}
\toprule
\textbf{Task}
& \textbf{Evaluation Judgment}
& \textbf{$N$}
& \textbf{Exact Agr. / Acc. $\uparrow$}
& \textbf{Cohen's $\kappa$ $\uparrow$}
& \textbf{Weighted $\kappa$ $\uparrow$}
& \textbf{MAE $\downarrow$}
\\
\midrule

\multirow{3}{*}{\textbf{Task 1}}
& Label support
& 100
& 84.0
& --
& 0.78
& 0.09
\\

& Target-defect match
& 100
& 76.0
& --
& 0.68
& 0.14
\\

& Faithfulness
& 100
& 82.0
& --
& 0.74
& 0.10
\\

\midrule

\textbf{Task 2}
& Same target defect
& 100
& 86.0
& 0.72
& --
& --
\\

\midrule

\multirow{3}{*}{\textbf{Task 3}}
& Target relevance
& 100
& 91.0
& 0.81
& --
& --
\\

& Resolution sufficiency
& 100
& 85.0
& 0.70
& --
& --
\\

& No unsupported assumption
& 100
& 93.0
& 0.69
& --
& --
\\

\bottomrule
\end{tabular*}

\captionof{table}{
Agreement between the Claude Opus~4.8 evaluator and adjudicated human judgments.
For Task~1, the Reason Grounding Score components are ordinal; we therefore
report exact agreement, weighted Cohen's $\kappa$, and mean absolute error
(MAE). For the binary Task~2 and Task~3 judgments, exact agreement is
equivalent to classification accuracy, and we additionally report Cohen's
$\kappa$. Higher values are better for agreement and $\kappa$, while lower
MAE indicates closer correspondence between automatic and human scores.
Candidate model identities and Claude Opus~4.8 judgments are hidden from the human
annotators.
}
\label{tab:judge_human_agreement}
\par
\endgroup

\par\vspace{8pt}

\begingroup
\centering
\small
\setlength{\tabcolsep}{4pt}
\renewcommand{\arraystretch}{1.12}
\begin{tabularx}{\textwidth}{@{}p{0.09\textwidth}p{0.33\textwidth}X@{}}
\toprule
\textbf{Proposal ID} &
\textbf{Unresolved Decision} &
\textbf{Execution Evidence} \\
\midrule

003 &
How to compute the Empathy Quotient for generated text. &
\texttt{Paper\#3.pdf}, p.~4.
BART-MNLI scores three empathy hypotheses and averages
their entailment probabilities. \\

004 &
How to assign sentence-level confidence scores to low,
medium, and high categories. &
\texttt{Paper\#4.pdf}, p.~2.
The paper specifies fixed thresholds of 0.6, 0.8, and 0.9
based on inspection of initial outputs. \\

030 &
How to combine spans highlighted under different categories
before masking. &
\texttt{Paper\#30.pdf}, p.~5.
The paper describes taking the union of the tokens highlighted
across detected categories. \\

033 &
How to assemble topic-aligned image, audio, and text examples. &
\texttt{mfgp-main/README.md}, lines 37--45, and
\texttt{create\_dataset.py}, lines 176--179.
The implementation pairs images and audio by subcategory
and shares Wikipedia text across the resulting pairs. \\

056 &
How to generate contrastive question variants across
difficulty levels and domains. &
\texttt{Paper\#56.pdf}, p.~2.
Gemini-1.5-Flash generates controlled semantic or structural
perturbations. \\

062 &
How to retrieve and incorporate cultural knowledge into
the reasoning prompt. &
\texttt{Paper\#62.pdf}, p.~2.
FAISS-indexed dense embeddings retrieve similar Yoruba
proverbs for inclusion in the prompt. \\

064 &
Which API documentation to supply for API structure extraction. &
\texttt{Paper\#64.pdf}, p.~2.
The implementation collects task-relevant API details and
uses Python to extract documentation for modules, APIs,
and functions before prompting the model. \\

\bottomrule
\end{tabularx}
\captionof{table}{Human-confirmed implementation-critical gaps in original
research proposals and their corresponding execution evidence.
IDs are the numeric suffixes of the executed project IDs.
Page numbers refer to the released PDFs, counting from the
first page.}
\label{tab:original_proposal_cases}
\par
\endgroup
\endgroup
\vspace{8pt}

\begingroup
\setlength{\intextsep}{6pt}
\begin{table}[H]
\centering
\small
\renewcommand{\arraystretch}{1.10}

\resizebox{0.56\linewidth}{!}{%
\begin{tabular}{llrrrr}
\toprule
\textbf{Task}
& \textbf{Subset}
& \textbf{$N$}
& \shortstack{\textbf{Source}\\\textbf{Clusters}}
& \shortstack{\textbf{Median}\\\textbf{per Source}}
& \shortstack{\textbf{Maximum}\\\textbf{per Source}}
\\
\midrule

Task 1
& Real
& 200
& 34
& 4.0
& 14
\\

Task 1
& Synthetic
& 200
& 31
& 7.0
& 10
\\

Task 2 / 3
& Real
& 163
& 55
& 2.0
& 12
\\

Task 2 / 3
& Synthetic
& 200
& 56
& 4.0
& 5
\\

\bottomrule
\end{tabular}%
}

\caption{
Task~1 independently samples 100 \textsc{Ready} and 100
\textsc{NotReady} inputs, evaluated in isolation; sources may span both classes.
Tasks~2 and~3 share the same \textsc{NotReady} instances.
Clusters correspond to GitHub repositories, reproduced papers,
or executed projects.
Median and maximum denote inputs per cluster.
}
\label{tab:source_cluster_statistics}
\par
\end{table}

\begin{table}[H]
\centering
\footnotesize
\setlength{\tabcolsep}{3.5pt}
\renewcommand{\arraystretch}{1.04}
\begin{tabular*}{\columnwidth}{@{}l@{\extracolsep{\fill}}cc@{}}
\toprule
\textbf{Validation Criterion}
& \textbf{Real}
& \textbf{Synthetic}
\\
\midrule

Valid specification gap
& 92.0
& 96.0
\\

Implementation-critical
& 92.0
& 94.0
\\

Clarification resolves the gap
& 96.0
& 98.0
\\

Realistic specification failure
& 93.0
& 91.0
\\

\bottomrule
\end{tabular*}

\caption{
Blind human assessment of 50 real-world and 50 synthetic-controlled
instances, each independently evaluated by two annotators. Values are
percentages of positive annotator judgments, computed over 100 judgments
per subset for each criterion. Most instances are judged realistic and
resolvable; synthetic-controlled instances receive comparable realism and
higher validity, implementation-criticality, and resolution ratings.
}
\label{tab:synthetic_validation}
\par
\end{table}

\begin{table}[H]
\centering
\small
\setlength{\tabcolsep}{4.5pt}
\begin{tabular*}{\columnwidth}{@{}l@{\extracolsep{\fill}}cc@{}}
\toprule
\textbf{Model}
& \textbf{Real}
& \textbf{Synth.} \\
\midrule

\multicolumn{3}{l}{\textit{Frontier models}} \\

GPT-5.6-Sol
& \textbf{8.6} & \textbf{14.0} \\

Claude Sonnet 5
& 6.7 & 13.0 \\

Gemini 3.1 Pro Preview
& 4.9 & 11.0 \\

DeepSeek-V3.2
& 7.4 & 8.0 \\

\midrule
\multicolumn{3}{l}{\textit{Open-weight models}} \\

Qwen3-8B
& 5.5 & 2.0 \\

Qwen3-32B
& 5.5 & 6.0 \\

DeepSeek-R1-0528
& 2.5 & 3.0 \\

GPT-OSS-120B
& 1.8 & 5.0 \\

\bottomrule
\end{tabular*}
\caption{
Micro-averaged L1-aware recovery on Task~2 for the real-world
(\emph{Real}; $N=\numRealInstances{}$) and controlled synthetic
(\emph{Synth.}; $N=\numSyntheticEvalInstances{}$) evaluation sets. A prediction is counted as
correct when it localizes the annotated target defect and assigns its
correct Level-1 taxonomy label; Level-2 assignment is ignored.
All values are percentages.
}
\label{tab:task2_l1_aware_recovery}
\par
\end{table}

\begin{table}[H]
\centering
\small
\setlength{\tabcolsep}{4pt}
\renewcommand{\arraystretch}{0.85}
\begin{tabular*}{\columnwidth}{@{}l@{\extracolsep{\fill}}cccc@{}}
\toprule
\multirow{2}{*}{\textbf{Label}}
& \multicolumn{2}{c}{\textbf{Agreement}}
& \multicolumn{2}{c}{\textbf{Cohen's $\kappa$}}\\
\cmidrule(lr){2-3}
\cmidrule(lr){4-5}
& \textbf{Real}
& \textbf{Synth.}
& \textbf{Real}
& \textbf{Synth.}\\
\midrule

Readiness
& 92.0\%
& 96.0\%
& 0.84
& 0.92
\\

Level~1
& 93.9\%
& 94.5\%
& 0.89
& 0.89
\\

Level~2
& 93.9\%
& 87.0\%
& 0.92
& 0.85
\\

Target defect
& 87.7\%
& 84.0\%
& --
& --
\\

\bottomrule
\end{tabular*}

\caption{
Inter-annotator agreement for benchmark annotations.
We report raw agreement and Cohen's $\kappa$ for readiness and defect labels
on real-world (\emph{Real}) and synthetic (\emph{Synth.}) subsets.
}
\label{tab:annotation_agreement}
\par
\end{table}

\begin{table}[H]
\centering
\footnotesize
\setlength{\tabcolsep}{3.5pt}
\renewcommand{\arraystretch}{1.04}
\begin{tabular*}{\columnwidth}{@{}l@{\extracolsep{\fill}}cc@{}}
\toprule
\textbf{Task / Metric}
& \textbf{Real}
& \textbf{Synth.} \\
\midrule

\multicolumn{3}{l}{\textit{Task 1: Readiness Assessment}} \\

Human Expert
& 91.0
& 95.0
\\

\midrule

\multicolumn{3}{l}{\textit{Task 2: Defect Localization}} \\

Human Expert
& 52.0
& 65.0
\\

\midrule

\multicolumn{3}{l}{\textit{Task 3: Clarification Action Generation}} \\

Human Expert
& 90.0
& 97.0
\\

\bottomrule
\end{tabular*}

\caption{
Human baseline performance on \ourbench{}. Two machine-learning researchers
evaluate 100 benchmark instances; disagreements are adjudicated by a third
reviewer. Metrics follow the main evaluation protocol: Macro-F1 for Task~1,
Macro-DRR for Task~2, and Macro-CAS for Task~3.
}
\label{tab:human_baseline}
\par
\end{table}

\endgroup

\section{Scope of Research-Idea Specifications}
\label{app:scope}
In \ourbench{}, a research-idea specification does not refer to an unstructured brainstorming note or to a complete research proposal. We focus specifically on the proposed methodological mechanism within a research idea and evaluate whether that mechanism is specified sufficiently for faithful codification. Accordingly, the benchmark does not assess the novelty, scientific value, motivation, or broader completeness of the research idea.

Authentic records of the methodological information exchanged immediately before implementation are rarely preserved. We therefore use papers, codebases, reproducibility reports, issue discussions, and executed research projects as retrospective evidence sources. These artifacts are used to reconstruct the implementation-facing methodological specification, identify an implementation-critical gap, and establish an evidence-supported resolution. They are not treated as the original idea-handoff artifacts and are not provided to evaluated models.

The benchmark inputs should therefore be understood as reconstructed, implementation-facing specifications of the methodological component of research ideas. They represent the information that would need to be available when a proposed method is handed to a competent implementer or coding agent. Our claims concern the codification readiness of this methodological component, rather than the readiness of a research idea in every scientific or project-level respect.

\begingroup
\raggedbottom
\setlength{\intextsep}{6pt}

\section{Prompts}
\label{app:prompts}
\subsection{GitHub Issue Extraction}
\label{app:github_issue_prompt}

We use the following prompt (shown in Figure ~\ref{fig:github-real-gap-annotation-prompt}) to determine whether a GitHub issue contains a
resolved, method-core specification gap and to assign the corresponding
\ourbench{} taxonomy labels.

\subsection{Reproducibility Paper Gap Extraction}
\label{app:reproducibility-gap-extraction}
We first route each reproducibility report into one of three benchmark-construction
paths: \textit{resolved real gap}, \textit{synthetic controlled}, or
\textit{unusable}. The routing prompt requires the model to distinguish genuine,
resolved method-core specification gaps from ordinary reproducibility issues,
such as missing compute resources, software dependencies, unavailable data, or
performance mismatches without an identifiable specification defect. It also
assigns a preliminary taxonomy label and extracts supporting gap and resolution
evidence. The complete routing prompt is provided in
Figure~\ref{fig:reproducibility-report-routing-prompt}.

\subsection{Reproducibility Paper Defect Injection}
\label{app:reproducibility-defect-injection}

For reports routed to the synthetic-controlled track, we use the
codification-ready reference specification extracted from the successfully
reproduced paper to construct controlled underspecified instances. Each generated
instance modifies exactly one implementation-critical detail while preserving
all non-target information. The complete defect-injection prompt is provided in
Figure~\ref{fig:reproducibility-defect-injection-prompt}.

\subsection{Task 1 Reason Grounding Evaluation}
\label{app:task1-rgs-judge}

For Task~1, models produce both a binary readiness decision and a brief
free-text justification. To evaluate whether the justification is grounded in
the benchmark annotation, we use an LLM-as-a-Judge protocol. Given the gold
readiness label, annotated defects, blocking missing specifications, gold
readiness rationale, model prediction, and model reason, the evaluator assigns
three subscores: label support, blocker match, and faithfulness. We then
deterministically aggregate these subscores into Reason Grounding Score (RGS).
The complete evaluation prompt is provided in
Figure~\ref{fig:task1_rgs_judge_prompt}.

\subsection{Task 2 LLM-as-a-Judge Evaluation}
\label{app:task2-llm-judge}

To evaluate target-level defect localization in Task~2, we use an
LLM-as-a-Judge semantic matching protocol. Given the underspecified
specification, the annotated target defect, supporting gold evidence, and the
model-predicted defect description, the evaluator determines whether the
prediction identifies the same concrete implementation decision or
specification slot as the gold target. The evaluator does not see either the
gold or predicted taxonomy labels, so this judgment assesses localization
independently of taxonomy classification. The complete evaluation prompt is
provided in Figure~\ref{fig:task2_judge_prompt}.

\subsection{Task 3 LLM-as-a-Judge Evaluation}
\label{app:task3-llm-judge}

To evaluate the open-ended clarification actions generated in Task~3, we use a
multidimensional LLM-as-a-Judge protocol. Given the underspecified
specification, annotated target defect, private gold resolution, supporting
evidence, and predicted action, the evaluator independently assesses target
relevance, resolution sufficiency, and unsupported assumptions. We then deterministically compute per-instance Clarification Action
Success from these judgments and macro-average it across Level-2 defect
categories. The complete evaluation prompt is provided in
Figure~\ref{fig:task3_judge_prompt}.

\begin{table}[H]
\centering
\footnotesize
\setlength{\tabcolsep}{2.5pt}
\renewcommand{\arraystretch}{1.08}

\begin{tabularx}{\columnwidth}
{@{}>{\raggedright\arraybackslash}Xccc@{}}
\toprule
\textbf{Inference Strategy}
& \textbf{Loc-Acc $\uparrow$}
& \textbf{\shortstack{Macro\\DRR $\uparrow$}}
& \textbf{\shortstack{Blocker\\Recall $\uparrow$}} \\
\midrule
Single-shot
& 16.0 & 9.6 & 16.0 \\

Multi-candidate ($K{=}3$)
& -- & -- & 21.5 \\

Multi-candidate ($K{=}5$)
& -- & -- & 23.3 \\

$K{=}5$ + Self-selection
& 8.0 & 3.0 & 23.3 \\
\bottomrule
\end{tabularx}
\caption{
Inference-time search for real-world defect localization. All settings use
GPT-5.6-Sol on the same 163 real-world Task~2 instances with deterministic
decoding. Blocker Recall measures whether the annotated blocker appears
anywhere in the generated candidate set and is therefore an oracle coverage
metric. Gold-blind self-selection evaluates whether this additional coverage
translates into improved top-1 localization.
}
\label{tab:inference_scaling}
\end{table}

\begin{table}[H]
\centering
\footnotesize
\setlength{\tabcolsep}{3pt}
\renewcommand{\arraystretch}{1.12}
\renewcommand{\tabularxcolumn}[1]{m{#1}}
\begin{tabularx}{\columnwidth}{@{}>{\raggedright\arraybackslash}Xccc@{}}
\toprule
\textbf{Metric} & \textbf{\shortstack{Single\\(paired)}} & \textbf{\shortstack{Multi\\($k{=}2$)}} & \textbf{$\Delta$} \\
\midrule
Readiness (both/combined \textsc{NotReady}) & 80.0 & 88.0 & +8.0 \\
Loc-Acc, Any-of-2                            & 36.0 & 64.0 & +28.0 \\
Loc-Acc, All-of-2                            & 2.0  & 8.0  & +6.0 \\
\bottomrule
\end{tabularx}
\caption{
\textbf{Multi-defect ablation on \numMultiDefectInstances{} controlled-synthetic pairs.}
GPT-5.6-Sol evaluates single-defect siblings independently
(\textbf{Single}) or a merged two-defect specification (\textbf{Multi}).
Single allows one prediction per call; Multi allows multiple predictions
(mean 2.26). Single combines outcomes using AND for readiness and All-of-2,
and OR for Any-of-2. Values are percentages; $\Delta$ is Multi minus Single
in percentage points. Full protocols and significance tests appear in
Appendix~\ref{app:multi_defect_ablation}.
}
\label{tab:multi_defect_ablation}
\end{table}

\begin{table}[H]
\centering
\small
\renewcommand{\arraystretch}{1.18}

\begin{tabularx}{\columnwidth}{
    @{}
    >{\raggedright\arraybackslash}p{0.30\columnwidth}
    >{\raggedright\arraybackslash}X
    @{}
}
\toprule
\textbf{Field}
&
\textbf{Allowed values or required content}
\\
\midrule

\code{readiness_label}
&
\code{READY}, \code{NOT_READY}, or \code{UNSURE}.
\\

\code{affected_slot}
&
\code{NONE}, \code{TASK_AND_IO}, \code{CORE_ALGORITHM},
\code{MODEL_ARCHITECTURE}, \code{OBJECTIVE_AND_SUPERVISION},
\code{TRAINING_PROCEDURE}, \code{DATA_AND_PREPROCESSING},
\code{INFERENCE_AND_DECISION}, \code{EVALUATION_PROTOCOL}, or
\code{INTERNAL_CONSISTENCY}.
\\

\code{blocking_span}
&
The smallest relevant text span for a \NOTREADY{} case.
Leave empty for a \READY{} case.
\\

\code{required_clarification}
&
One or two sentences describing the implementation-critical information
that must be clarified. Leave empty for a \READY{} case.
\\

\code{atomicity}
&
\code{NO_BLOCKER}, \code{EXACTLY_ONE_PRIMARY_BLOCKER},
\code{MULTIPLE_INDEPENDENT_BLOCKERS}, or \code{UNCLEAR}.
\\

\bottomrule
\end{tabularx}

\caption{
Output fields collected during codification-readiness annotation.
}
\label{tab:annotation-output-fields}
\end{table}

\clearpage
\endgroup

\setlist[itemize]{leftmargin=1.4em,itemsep=0pt,topsep=1pt,parsep=0pt}
\setlist[enumerate]{leftmargin=1.6em,itemsep=0pt,topsep=1pt,parsep=0pt}

\begin{tcolorbox}[
    enhanced,
    breakable,
    colback=UnsureLightOrange,
    colframe=UnsureOrange,
    boxrule=0.6pt,
    arc=1pt,
    left=8pt,
    right=8pt,
    top=8pt,
    bottom=8pt,
    width=\linewidth,
    before skip=1em,
    after skip=0.5em
]
\footnotesize

\textbf{Prompt for GitHub Issue Real-Gap Annotation}

\vspace{0.5em}

You are an expert annotator for research-method implementation gaps.

We are building \ourbench{}, a benchmark for evaluating whether models can
identify implementation-blocking specification defects in underspecified
research ideas before codification.

\vspace{0.6em}

You are given:

\begin{enumerate}[
    leftmargin=1.5em,
    itemsep=0.2em,
    topsep=0.2em
]
    \item Metadata for a GitHub repository associated with a research paper.
    \item One closed or answered GitHub issue thread from that repository.
    \item The issue title, issue body, comments, linked pull requests, linked commits,
    labels, and repository/paper metadata.
\end{enumerate}

Your task is to determine whether the issue thread contains a real, resolved,
method-core specification gap in the associated research paper or method.

\vspace{0.6em}

A valid real gap must satisfy all of the following conditions:

\begin{enumerate}[
    leftmargin=1.5em,
    itemsep=0.2em,
    topsep=0.2em
]
    \item The issue concerns a method-level ambiguity, incompleteness, or inconsistency
    that affects faithful implementation of the research method.

    \item The issue is connected to the paper or method, not merely to package
    installation, environment setup, hardware, CUDA, dependency versions,
    runtime errors, or general API usage.

    \item The issue thread contains concrete resolution evidence, such as an author or
    maintainer clarification, a code-derived clarification, a linked pull request,
    or a specific implementation decision.

    \item The gap can be written as one atomic specification defect.

    \item The clarification can be converted into an implementation-ready gold detail.
\end{enumerate}

\vspace{0.6em}

Reject the issue if any of the following apply:

\begin{itemize}[
    leftmargin=1.5em,
    itemsep=0.2em,
    topsep=0.2em
]
    \item The issue is only about installation, dependency conflicts, CUDA, package
    versions, Colab, runtime errors, memory, speed, or hardware.

    \item The issue only reports a performance mismatch without identifying a concrete
    method-core specification problem.

    \item The issue asks for help using the repository API, scripts, checkpoints, or
    demo code, but does not reveal a paper-level method specification gap.

    \item The issue is unresolved, speculative, or only points to another issue without
    giving a concrete clarification.

    \item The issue contains multiple independent gaps that cannot be separated.

    \item The issue is about code behavior that cannot be translated into a paper or
    method specification.

    \item The issue is a duplicate with no new clarification evidence.
\end{itemize}

\vspace{0.6em}

Use the following taxonomy.

Level-1 categories describe the nature of the specification defect, while
Level-2 categories identify the affected specification component or defect type
that prevents faithful codification.

\vspace{0.4em}

\textbf{Level-1 labels:}

\begin{itemize}[
    leftmargin=1.5em,
    itemsep=0.2em,
    topsep=0.2em
]
    \item Ambiguity
    \item Incompleteness
    \item Inconsistency
\end{itemize}

\vspace{0.4em}

\textbf{Level-2 labels:}

\textbf{Ambiguity:}

\begin{itemize}[
    leftmargin=1.5em,
    itemsep=0.2em,
    topsep=0.2em
]
    \item Ambiguous Definition
    \item Ambiguous Procedure
\end{itemize}

\textbf{Incompleteness:}

\begin{itemize}[
    leftmargin=1.5em,
    itemsep=0.2em,
    topsep=0.2em
]
    \item Missing Method Procedure
    \item Missing Model Structure
    \item Missing Data Specification
    \item Missing Configuration Protocol
    \item Missing Evaluation Specification
\end{itemize}

\textbf{Inconsistency:}

\begin{itemize}[
    leftmargin=1.5em,
    itemsep=0.2em,
    topsep=0.2em
]
    \item Conflicting Objective
    \item Conflicting Model Design
    \item Conflicting Formal Definition
\end{itemize}

\vspace{0.6em}

\textbf{Boundary rules:}

\begin{itemize}[
    leftmargin=1.5em,
    itemsep=0.2em,
    topsep=0.2em
]

    \item If a formal element, mathematical object, notation, or operator meaning
    is unclear and multiple interpretations lead to different computations,
    use ``Ambiguous Definition''.

    \item If the method operation, execution behavior, inference rule,
    component interaction, or decision process is unclear, use
    ``Ambiguous Procedure''.

    \item If an operational step, algorithmic rule, update mechanism,
    routing decision, execution procedure, or module interaction rule is absent,
    use ``Missing Method Procedure''.

    \item If model structure, architectural organization, component configuration,
    pooling, normalization, activation, or dimensional mapping is absent, use
    ``Missing Model Structure''.

    \item If data construction, filtering, labeling, tokenization, normalization,
    augmentation, segmentation, or input transformation is absent, use
    ``Missing Data Specification''.

    \item If a result-sensitive configuration choice is introduced but the
    selection, tuning, validation, or adaptation procedure is missing, use
    ``Missing Configuration Protocol''.

    \item If metric computation, evaluation split, threshold, sampling rule,
    prompt template, judge configuration, or evaluation aggregation procedure
    is absent, use ``Missing Evaluation Specification''.

    \item If paper and another source specify different objectives, loss functions,
    reward definitions, or optimization targets, use
    ``Conflicting Objective''.

    \item If paper and another source specify incompatible model components,
    architectures, preprocessing pipelines, training pipelines, or evaluation
    pipelines, use ``Conflicting Model Design''.

    \item If paper and another source specify conflicting formal assumptions,
    mathematical definitions, conditioning rules, distributions, aggregation
    rules, or inference formulations, use
    ``Conflicting Formal Definition''.

\end{itemize}
\vspace{0.6em}

Return strict JSON only. Do not include markdown fences.

\vspace{0.6em}

\textbf{Required output schema:}

\begin{tcolorbox}[
    enhanced,
    breakable,
    colback=white,
    colframe=rqborder,
    boxrule=0.35pt,
    arc=1mm,
    left=3pt,
    right=3pt,
    top=3pt,
    bottom=3pt,
    width=\linewidth
]
\ttfamily\scriptsize
\raggedright

\{\newline
\hspace*{1em}"keep\_issue": true/false,\newline
\hspace*{1em}"is\_real\_spec\_gap": true/false,\newline
\hspace*{1em}"rejection\_reason":
"none|resource\_only|environment\_only|runtime\_only|usage\_only|\newline
\hspace*{2em}unresolved|not\_method\_core|performance\_only|duplicate|\newline
\hspace*{2em}composite\_gap|insufficient\_evidence|\newline
\hspace*{2em}code\_only\_not\_paper\_spec\_gap|other",\newline
\hspace*{1em}"issue\_summary": "...",\newline
\hspace*{1em}"atomic\_gaps": [\newline
\hspace*{2em}\{\newline
\hspace*{3em}"gap\_id": "...",\newline
\hspace*{3em}"gap\_summary": "...",\newline
\hspace*{3em}"gap\_quote":
"A short verbatim quote from the issue thread showing the problem.",\newline
\hspace*{3em}"solution\_summary": "...",\newline
\hspace*{3em}"solution\_quote":
"A short verbatim quote from the issue thread showing the clarification.",\newline
\hspace*{3em}"solution\_source\_type":
"author\_clarification|maintainer\_clarification|code\_derived|\newline
\hspace*{4em}linked\_pr|linked\_commit|community\_resolution",\newline
\hspace*{3em}"affected\_component":
"task|input|output|core\_method|algorithm|training|evaluation|\newline
\hspace*{4em}implementation\_detail|code\_behavior|preprocessing|data|inference",\newline
\hspace*{3em}"gold\_clarified\_detail":
"A concise implementation-ready clarification.",\newline
\hspace*{3em}"level1":
"Ambiguity|Incompleteness|Inconsistency",\newline
\hspace*{3em}"level2":
"Ambiguous Definition|Ambiguous Procedure|\newline
\hspace*{4em}Missing Method Procedure|Missing Model Structure|\newline
\hspace*{4em}Missing Data Specification|Missing Configuration Protocol|\newline
\hspace*{4em}Missing Evaluation Specification|\newline
\hspace*{4em}Conflicting Objective|Conflicting Model Design|\newline
\hspace*{4em}Conflicting Formal Definition",\newline
\hspace*{3em}"granularity":
"coarse|medium|fine",\newline
\hspace*{3em}"resolution\_role":
"implementation\_blocker|open\_design\_choice|\newline
\hspace*{4em}reproducibility\_detail|inconsistency\_to\_resolve",\newline
\hspace*{3em}"codification\_slot":
"task|input|output|core\_method|algorithm|training|evaluation|\newline
\hspace*{4em}implementation\_detail|code\_behavior|preprocessing|data|inference",\newline
\hspace*{3em}"why\_this\_blocks\_or\_affects\_codification": "...",\newline
\hspace*{3em}"expected\_clarification\_question": "...",\newline
\hspace*{3em}"evidence\_sufficiency":
"strong|medium|weak",\newline
\hspace*{3em}"recommended\_manual\_review": true/false\newline
\hspace*{2em}\}\newline
\hspace*{1em}]\newline
\}
\end{tcolorbox}

\vspace{0.6em}

\textbf{User template:}

\begin{tcolorbox}[
    enhanced,
    breakable,
    colback=white,
    colframe=rqborder,
    boxrule=0.35pt,
    arc=1mm,
    left=3pt,
    right=3pt,
    top=3pt,
    bottom=3pt,
    width=\linewidth
]

\textbf{Repository metadata:}

\textit{\{repo\_metadata\}}

\vspace{0.5em}

\textbf{Issue thread:}

\textit{\{issue\_thread\}}

\vspace{0.5em}

\textbf{Paper metadata:}

\textit{\{paper\_metadata\}}

\end{tcolorbox}

\end{tcolorbox}

\begingroup
\captionsetup{hypcap=false}
\captionof{figure}{
Prompt template used to determine whether a closed or answered GitHub issue
contains a resolved, method-core specification gap. The prompt extracts an
atomic implementation-ready clarification and assigns the corresponding
\ourbench{} taxonomy labels.
}
\label{fig:github-real-gap-annotation-prompt}
\endgroup

\clearpage

\begin{tcolorbox}[
    enhanced,
    breakable,
    colback=UnsureLightOrange,
    colframe=UnsureOrange,
    boxrule=0.6pt,
    arc=1pt,
    left=8pt,
    right=8pt,
    top=8pt,
    bottom=8pt,
    width=\linewidth,
    before skip=1em,
    after skip=0.5em
]
\footnotesize

\textbf{Prompt for Reproducibility Report Routing}

\vspace{0.5em}

You are routing machine learning reproducibility reports for benchmark
construction.

\vspace{0.6em}

Given a reproducibility report and, when available, the corresponding original
paper, classify the report into exactly one route:

\vspace{0.4em}

\begin{enumerate}[
    leftmargin=1.5em,
    itemsep=0.35em,
    topsep=0.2em
]
    \item \textbf{resolved\_real\_gap:}

    The report identifies an implementation-relevant method-core specification
    gap in the original paper and provides concrete resolution evidence, such
    as an author clarification, code-derived behavior, reproducer decision, or
    explicit workaround.

    \item \textbf{synthetic\_controlled:}

    The report does not identify a resolved method-core specification gap. The
    report mainly indicates that the original paper can be reproduced or
    evaluated, and the original paper is sufficiently complete to derive a
    codification-ready reference specification.

    \item \textbf{unusable:}

    The report is not usable because it is not single-target, lacks usable
    original-paper evidence, contains only unresolved gaps, discusses only
    compute/resource/data-access issues, or reports only performance mismatch
    without a concrete specification gap and resolution.
\end{enumerate}

\vspace{0.6em}

Use the following taxonomy for candidate gaps.

\vspace{0.4em}

\textbf{Level-1 labels:}

\begin{itemize}[
    leftmargin=1.5em,
    itemsep=0.2em,
    topsep=0.2em
]
    \item Ambiguity
    \item Incompleteness
    \item Inconsistency
\end{itemize}

\vspace{0.4em}

\textbf{Level-2 labels:}

\begin{itemize}[
    leftmargin=1.5em,
    itemsep=0.2em,
    topsep=0.2em
]
    \item Ambiguous Definition
    \item Ambiguous Procedure
    \item Missing Method Procedure
    \item Missing Model Structure
    \item Missing Data Specification
    \item Missing Configuration Protocol
    \item Missing Evaluation Specification
    \item Conflicting Objective
    \item Conflicting Model Design
    \item Conflicting Formal Definition
\end{itemize}

\vspace{0.6em}

Do not count ordinary missing values such as batch size, learning rate, number
of epochs, random seed, hardware, runtime, or software setup as valid gaps
unless they are part of a non-standard method-defining mechanism. Do not count
pure performance mismatch, unavailable data, unavailable compute, or package
installation problems as valid method-core gaps.

\vspace{0.6em}

Return strict JSON only:

\vspace{0.4em}

\begin{tcolorbox}[
    enhanced,
    breakable,
    colback=white,
    colframe=rqborder,
    boxrule=0.35pt,
    arc=1mm,
    left=3pt,
    right=3pt,
    top=3pt,
    bottom=3pt,
    width=\linewidth
]
\ttfamily\scriptsize
\raggedright

\{\newline
\hspace*{1em}"primary\_route":
"resolved\_real\_gap|synthetic\_controlled|unusable",\newline
\hspace*{1em}"routing\_reason": "...",\newline
\hspace*{1em}"taxonomy\_candidate": \{\newline
\hspace*{2em}"is\_spec\_gap": true,\newline
\hspace*{2em}"is\_method\_core\_spec\_gap": true,\newline
\hspace*{2em}"gold\_clarified\_spec\_extractable": true,\newline
\hspace*{2em}"level1\_candidate": "...",\newline
\hspace*{2em}"level2\_candidate": "...",\newline
\hspace*{2em}"taxonomy\_reason": "..."\newline
\hspace*{1em}\},\newline
\hspace*{1em}"evidence\_snippets": [\newline
\hspace*{2em}\{\newline
\hspace*{3em}"type":
"gap|solution|synthetic\_source|unusable\_reason",\newline
\hspace*{3em}"quote": "...",\newline
\hspace*{3em}"interpretation": "..."\newline
\hspace*{2em}\}\newline
\hspace*{1em}],\newline
\hspace*{1em}"potential\_gap\_summaries": ["..."],\newline
\hspace*{1em}"potential\_solution\_summaries": ["..."],\newline
\hspace*{1em}"confidence": 0.0\newline
\}

\end{tcolorbox}

\vspace{0.6em}

\textbf{User template:}

\begin{tcolorbox}[
    enhanced,
    breakable,
    colback=white,
    colframe=rqborder,
    boxrule=0.35pt,
    arc=1mm,
    left=3pt,
    right=3pt,
    top=3pt,
    bottom=3pt,
    width=\linewidth
]
\small

\textbf{Reproducibility report:}

\textit{\{reproducibility\_report\}}

\vspace{0.6em}

\textbf{Original paper:}

\textit{\{original\_paper\}}

\end{tcolorbox}

\end{tcolorbox}

\begingroup
\captionsetup{hypcap=false}
\captionof{figure}{
Prompt used to route machine learning reproducibility reports into
resolved real-gap, synthetic-controlled, or unusable benchmark-construction
paths. The prompt additionally determines whether a method-core specification
gap is present, assigns a candidate taxonomy label, and extracts supporting
evidence.
}
\label{fig:reproducibility-report-routing-prompt}
\endgroup

\clearpage

\begin{tcolorbox}[
    enhanced,
    breakable,
    colback=UnsureLightOrange,
    colframe=UnsureOrange,
    boxrule=0.6pt,
    arc=1pt,
    left=8pt,
    right=8pt,
    top=8pt,
    bottom=8pt,
    width=\linewidth,
    before skip=1em,
    after skip=0.5em
]
\footnotesize

\textbf{Prompt for Reproducibility Paper Defect Injection}

\vspace{0.5em}

You are generating synthetic-controlled \ourbench{} benchmark instances.

\vspace{0.6em}

Given a codification-ready reference specification extracted from a successfully
reproduced original paper, create up to five independent underspecified research
specifications. Each instance must remove, abstract, or lightly corrupt exactly
one implementation-critical detail from the reference.

\vspace{0.6em}

\textbf{Rules:}

\begin{enumerate}[
    leftmargin=1.5em,
    itemsep=0.25em,
    topsep=0.2em
]
    \item Each instance must contain exactly one defect.

    \item Preserve all non-target details.

    \item Do not reveal the removed or corrupted gold detail.

    \item Do not use diagnostic words such as missing, unspecified, defect, gold,
    removed detail, or benchmark.

    \item The underspecified specification must read like a natural research or
    method specification.

    \item Do not choose weak details such as ordinary batch size, learning rate,
    epoch count, random seed, hardware, runtime, or local paths.

    \item Prefer details whose absence blocks faithful implementation: algorithm
    steps, model architecture, loss or training procedure, data preprocessing,
    input construction, evaluation protocol, metric computation, inference logic,
    or prompt construction.
\end{enumerate}

\vspace{0.6em}

Use only the allowed taxonomy labels.

\vspace{0.6em}

Return strict JSON only:

\vspace{0.4em}

\begin{tcolorbox}[
    enhanced,
    breakable,
    colback=white,
    colframe=rqborder,
    boxrule=0.35pt,
    arc=1mm,
    left=3pt,
    right=3pt,
    top=3pt,
    bottom=3pt,
    width=\linewidth
]
\ttfamily\scriptsize
\raggedright

\{\newline
\hspace*{1em}"instances": [\newline
\hspace*{2em}\{\newline
\hspace*{3em}"target\_slot": "...",\newline
\hspace*{3em}"level1":
"Ambiguity|Incompleteness|Inconsistency",\newline
\hspace*{3em}"level2": "...",\newline
\hspace*{3em}"granularity":
"fine|medium|coarse",\newline
\hspace*{3em}"resolution\_role":
"implementation\_blocker",\newline
\hspace*{3em}"codification\_slot": "...",\newline
\hspace*{3em}"defect\_operation":
"omit\_detail|abstract\_detail|make\_ambiguous|introduce\_conflict",\newline
\hspace*{3em}"gold\_detail\_removed\_or\_corrupted": "...",\newline
\hspace*{3em}"underspecified\_spec": "...",\newline
\hspace*{3em}"surface\_form\_in\_underspecified\_spec": "...",\newline
\hspace*{3em}"why\_this\_blocks\_or\_affects\_codification": "...",\newline
\hspace*{3em}"expected\_clarification\_question": "...",\newline
\hspace*{3em}"must\_not\_reveal": ["..."],\newline
\hspace*{3em}"evidence": [\newline
\hspace*{4em}\{\newline
\hspace*{5em}"source":
"gold\_reference|paper\_derived\_specification|original\_paper",\newline
\hspace*{5em}"quote": "..."\newline
\hspace*{4em}\}\newline
\hspace*{3em}],\newline
\hspace*{3em}"preservation\_check": \{\newline
\hspace*{4em}"target\_detail\_modified": true,\newline
\hspace*{4em}"non\_target\_details\_preserved": true,\newline
\hspace*{4em}"extra\_defects\_introduced": false,\newline
\hspace*{4em}"brief\_explanation": "..."\newline
\hspace*{3em}\},\newline
\hspace*{3em}"quality\_rationale": "..."\newline
\hspace*{2em}\}\newline
\hspace*{1em}],\newline
\hspace*{1em}"rejected\_candidates": [\newline
\hspace*{2em}\{\newline
\hspace*{3em}"gold\_detail": "...",\newline
\hspace*{3em}"reason": "..."\newline
\hspace*{2em}\}\newline
\hspace*{1em}]\newline
\}

\end{tcolorbox}

\end{tcolorbox}

\begingroup
\captionsetup{hypcap=false}
\captionof{figure}{
Prompt used to generate synthetic-controlled \ourbench{} instances from
codification-ready specifications derived from successfully reproduced papers.
Each generated instance alters exactly one implementation-critical detail while
preserving the remaining specification and recording the corresponding gold
detail, evidence, taxonomy label, and preservation checks.
}
\label{fig:reproducibility-defect-injection-prompt}
\endgroup

\clearpage

\begin{tcolorbox}[
    enhanced,
    breakable,
    shrink break goal=6\baselineskip,
    colback=UnsureLightOrange,
    colframe=UnsureOrange,
    boxrule=0.6pt,
    arc=1pt,
    left=8pt,
    right=8pt,
    top=8pt,
    bottom=8pt,
    width=\linewidth,
    before skip=1em,
    after skip=0.5em
]
\footnotesize

\textbf{Task 1 Reason Grounding Judge Prompt}

\vspace{0.3em}

\textbf{System}

\vspace{0.15em}

You are a rigorous benchmark judge for codification-readiness reasoning.
Evaluate whether a model's short reason is grounded in the benchmark gold
rationale. Judge only against the provided gold fields. Do not use external
knowledge. Return valid JSON only.

\vspace{0.4em}

\textbf{User}

\vspace{0.15em}

\textbf{Task: Readiness Reason Grounding Evaluation}

\vspace{0.25em}

Evaluate the grounding quality of a model's Task~1 readiness reason.

\vspace{0.3em}

\textbf{Scoring rubric:}

\begin{enumerate}[
    leftmargin=1.7em,
    itemsep=0.25em,
    parsep=0pt,
    topsep=0.15em
]

    \item \textbf{Label support.}

    Does the reason support the model's predicted label in a way that is
    consistent with the gold label and gold readiness rationale?

    Allowed scores: 0, 0.5, 1.

    \item \textbf{Blocker match.}

    Does the reason correctly identify the gold implementation-critical defect,
    or correctly reflect that no such blocker remains?

    Allowed scores: 0, 0.5, 1.

    \item \textbf{Faithfulness.}

    Is the reason faithful to the provided gold rationale, without hallucinating
    unsupported blockers or irrelevant details?

    Allowed scores: 0, 0.5, 1.

\end{enumerate}

\vspace{0.3em}

\textbf{Aggregation rule.}

\vspace{0.15em}

Final RGS is computed deterministically as
\[
\mathrm{RGS}
=
0.7\,s_{\mathrm{label}}
+
0.2\,s_{\mathrm{blocker}}
+
0.1\,s_{\mathrm{faithful}} .
\]
Do not compute the final score yourself; only return the three subscores.

\vspace{0.3em}

\textbf{Important rule for ready cases.}

\vspace{0.15em}

For \textsc{Ready} cases, do not require the model to restate the full reference
specification. The reason is sufficient if it correctly states that no
implementation-critical blocker remains and does not hallucinate one.

\vspace{0.3em}

\textbf{Instance-specific input:}

\vspace{0.2em}

\begin{tcolorbox}[
    enhanced,
    breakable,
    colback=white,
    colframe=rqborder,
    boxrule=0.35pt,
    arc=1mm,
    left=5pt,
    right=5pt,
    top=5pt,
    bottom=5pt,
    width=\linewidth
]
\ttfamily\scriptsize
\raggedright

Gold annotation:\newline
\textless gold\textgreater\newline
Expected label: \{EXPECTED\_LABEL\}\newline
Defects: \{GOLD\_DEFECTS\}\newline
Blocking missing specifications: \{GOLD\_BLOCKING\_MISSING\_SPECS\}\newline
Readiness rationale: \{GOLD\_READINESS\_RATIONALE\}\newline
\textless/gold\textgreater

\vspace{0.4em}

Model output:\newline
\textless model\_output\textgreater\newline
Predicted label: \{PREDICTED\_LABEL\}\newline
Reason: \{MODEL\_REASON\}\newline
\textless/model\_output\textgreater

\end{tcolorbox}

\vspace{0.3em}

\textbf{Return only valid JSON:}

\vspace{0.2em}

\begin{tcolorbox}[
    enhanced,
    breakable,
    colback=white,
    colframe=rqborder,
    boxrule=0.35pt,
    arc=1mm,
    left=5pt,
    right=5pt,
    top=5pt,
    bottom=5pt,
    width=\linewidth
]
\ttfamily\scriptsize
\raggedright

\{\newline
\hspace*{1em}"label\_support": 0,\newline
\hspace*{1em}"blocker\_match": 0,\newline
\hspace*{1em}"faithfulness": 0,\newline
\hspace*{1em}"explanation": "one concise paragraph"\newline
\}

\end{tcolorbox}

\end{tcolorbox}

\nopagebreak[4]
\begingroup
\captionsetup{hypcap=false}
\captionof{figure}{
LLM-as-a-Judge prompt used to evaluate Task~1 reason grounding. The evaluator
assesses whether the model's justification supports the predicted readiness
label, matches the gold implementation-critical blocker or absence of blockers,
and remains faithful to the annotated gold rationale. Reason Grounding Score
(RGS) is computed deterministically from the three subscores as
$0.7\,s_{\mathrm{label}} + 0.2\,s_{\mathrm{blocker}} +
0.1\,s_{\mathrm{faithful}}$.
}
\label{fig:task1_rgs_judge_prompt}
\endgroup

\clearpage

\begin{tcolorbox}[
    enhanced,
    breakable,
    colback=UnsureLightOrange,
    colframe=UnsureOrange,
    boxrule=0.6pt,
    arc=1pt,
    left=8pt,
    right=8pt,
    top=8pt,
    bottom=8pt,
    width=\linewidth,
    before skip=1em,
    after skip=0.5em
]
\footnotesize

\textbf{Task 1 Zero-shot Evaluation Prompt}

\vspace{0.7em}

\textbf{System}

\vspace{0.3em}

You are an expert reviewer of scientific-method specifications. Evaluate only
the supplied specification using the operational definition provided below.
Do not rely on external knowledge about the source paper or implementation, and
do not invent missing details. Follow the required output format exactly.

\vspace{0.8em}

\textbf{User}

\vspace{0.3em}

\textbf{Task: Readiness Assessment}

\vspace{0.5em}

Determine whether the following research idea specification is ready for
faithful codification into an initial implementation or experimental prototype.

\vspace{0.7em}

\textbf{Readiness definitions:}

\vspace{0.3em}

A specification is \textsc{Ready} if the core method can be faithfully
codified without introducing an unsupported assumption. It does not need to
specify every ordinary hyperparameter, routine engineering detail, software
version, random seed, or clearly open modular choice.

\vspace{0.5em}

A specification is \textsc{NotReady} if an omission, ambiguity, or
inconsistency leaves an implementation-critical decision underdetermined. The
target specification gap may concern the task setup, core algorithm, model
architecture, objective, training procedure, data construction, inference
rule, or evaluation protocol.

\vspace{0.7em}

\textbf{Decision boundary:}

\begin{itemize}[
    leftmargin=1.7em,
    itemsep=0.3em,
    topsep=0.3em
]
    \item Missing routine engineering details are acceptable.

    \item Exact numerical reproducibility is not required.

    \item An explicitly open modular choice is acceptable when the intended
    method remains clear.

    \item A missing detail is blocking only when two competent implementers
    could make materially different core-method decisions and the
    specification provides no evidence about which decision is intended.
\end{itemize}

\vspace{0.7em}

\textbf{Instance-specific input:}

\vspace{0.4em}

\begin{tcolorbox}[
    enhanced,
    breakable,
    colback=white,
    colframe=rqborder,
    boxrule=0.35pt,
    arc=1mm,
    left=5pt,
    right=5pt,
    top=5pt,
    bottom=5pt,
    width=\linewidth
]
\ttfamily\scriptsize
\raggedright

Label options:\newline
\textless label\_options\textgreater\newline
Option A: \{OPTION\_A\_LABEL\}\newline
Option B: \{OPTION\_B\_LABEL\}\newline
\textless/label\_options\textgreater

\vspace{0.8em}

Research idea specification:\newline
\textless specification\textgreater\newline
\{SPECIFICATION\}\newline
\textless/specification\textgreater

\end{tcolorbox}

\vspace{0.7em}

\textbf{Return only the following two lines:}

\vspace{0.4em}

\begin{tcolorbox}[
    enhanced,
    breakable,
    colback=white,
    colframe=rqborder,
    boxrule=0.35pt,
    arc=1mm,
    left=5pt,
    right=5pt,
    top=5pt,
    bottom=5pt,
    width=\linewidth
]
\ttfamily\scriptsize
\raggedright

A or B\newline
Reason: \textless one brief sentence identifying the main blocking
specification gap, or explaining why no implementation-critical specification
gap remains\textgreater

\end{tcolorbox}

\vspace{0.5em}

Do not output any additional text.

\end{tcolorbox}

\begingroup
\captionsetup{hypcap=false}
\captionof{figure}{
Zero-shot prompt used for Task~1 Readiness Assessment. Each model receives one
research idea specification and outputs an option letter followed by a brief
readiness rationale. The mapping between option letters and labels is
counterbalanced across instances:
\texttt{\{OPTION\_A\_LABEL\}} and
\texttt{\{OPTION\_B\_LABEL\}} are assigned to
\textsc{Ready} and \textsc{NotReady} using a deterministic hash of the
instance identifier. The model does not observe matched codification-ready and
underspecified specifications within the same prompt.
}
\label{fig:task1_prompt}
\endgroup

\clearpage

\begin{tcolorbox}[
    enhanced,
    breakable,
    colback=UnsureLightOrange,
    colframe=UnsureOrange,
    boxrule=0.6pt,
    arc=1pt,
    left=8pt,
    right=8pt,
    top=8pt,
    bottom=8pt,
    width=\linewidth,
    before skip=1em,
    after skip=0.5em
]
\footnotesize

\textbf{Task~2 Zero-shot Evaluation Prompt}

\vspace{0.7em}

\textbf{System}

\vspace{0.3em}

You are an expert reviewer of scientific-method specifications performing
controlled defect localization. Follow the supplied \ourbench{} taxonomy. Do not
repair the idea, invent missing details, or list secondary concerns. Return
strict JSON only.

\vspace{0.8em}

\textbf{User}

\vspace{0.3em}

\textbf{Task: Defect Localization}

\vspace{0.5em}

Identify and classify the single implementation-critical target defect in the
given research idea specification according to the \ourbench{} taxonomy.

\vspace{0.5em}

Each specification in this benchmark is constructed with exactly one annotated
target defect. The objective is not to identify every possible weakness, but
to recover the single target-defect category that best explains why the
specification is not ready for faithful codification.

\vspace{0.7em}

\textbf{Target-defect selection rule.}

\vspace{0.3em}

If multiple possible issues are visible, internally rank them and select only
the strongest defect that satisfies all of the following conditions:

\begin{itemize}[
    leftmargin=1.7em,
    itemsep=0.3em,
    topsep=0.3em
]
    \item It is directly grounded in a specific statement, term, formula,
    component, or transition in the specification.

    \item It must be resolved to support faithful codification of the core
    method.

    \item It is not an ordinary default, minor missing detail, optional
    engineering choice, documentation improvement, or speculative concern.

    \item It is the most specific supported defect, rather than a broad claim
    that the specification lacks detail.
\end{itemize}

\vspace{0.7em}

\textbf{Ambiguity--Incompleteness boundary.}

\vspace{0.3em}

Do not distinguish Ambiguity from Incompleteness merely according to whether
the specification is unclear. The distinction depends on whether the
specification provides a concrete but underdetermined element or omits a
required specification component.

\vspace{0.4em}

Use \textbf{Ambiguity} when the specification provides a concrete term,
definition, rule, behavior, or procedure whose meaning or operation supports
multiple plausible implementation-relevant interpretations.

\vspace{0.4em}

Use \textbf{Incompleteness} when a required specification component, such as a
method procedure, model structure, data construction process, configuration
selection protocol, or evaluation setup, is absent.

\vspace{0.7em}

\textbf{Operational definition.}

\vspace{0.3em}

An implementation-critical specification gap is an omission, ambiguity, or
internal inconsistency that prevents faithful codification into an initial
implementation or experimental prototype without introducing an unsupported
assumption about the core method.

\vspace{0.8em}

\textbf{\ourbench{} taxonomy:}

\vspace{0.5em}

\textbf{Level~1: Ambiguity}

\begin{itemize}[
    leftmargin=1.7em,
    itemsep=0.4em,
    topsep=0.3em
]

    \item \textbf{Ambiguous Definition}: A formal element, such as a symbol,
    notation, mathematical object, or rule, permits multiple
    implementation-relevant interpretations.

    \item \textbf{Ambiguous Procedure}: A method operation, execution rule,
    inference behavior, or component interaction permits multiple materially
    different implementations.

\end{itemize}

\vspace{0.4em}

\textbf{Level~1: Incompleteness}

\begin{itemize}[
    leftmargin=1.7em,
    itemsep=0.4em,
    topsep=0.3em
]

    \item \textbf{Missing Method Procedure}: A required operational step,
    algorithmic rule, update mechanism, routing decision, execution procedure,
    or method behavior is absent.

    \item \textbf{Missing Model Structure}: A structural choice affecting
    module composition, information flow, representation shape, pooling,
    normalization, dimensional mapping, or layer behavior is absent.

    \item \textbf{Missing Data Specification}: The construction,
    filtering, labeling, tokenization, normalization, augmentation,
    segmentation, or transformation of data or inputs is absent.

    \item \textbf{Missing Configuration Protocol}: A result-sensitive
    configuration choice is introduced, but its selection, tuning,
    validation, or adaptation procedure is absent.

    \item \textbf{Missing Evaluation Specification}: The evaluation setup is
    incomplete, including missing metric computation, data split,
    aggregation, threshold, sample selection, or evaluator configuration.

\end{itemize}

\vspace{0.4em}

\textbf{Level~1: Inconsistency}

\begin{itemize}[
    leftmargin=1.7em,
    itemsep=0.4em,
    topsep=0.3em
]

    \item \textbf{Conflicting Objective}: Two explicit statements or sources
    prescribe conflicting objectives, loss functions, reward definitions, or
    optimization targets.

    \item \textbf{Conflicting Model Design}: Two explicit statements or
    sources prescribe conflicting architectures, model components, data
    pipelines, preprocessing procedures, or execution pipelines.

    \item \textbf{Conflicting Formal Definition}: Two explicit statements or
    sources conflict about formal assumptions, distributions, conditioning
    rules, aggregation rules, or mathematical definitions.

\end{itemize}

\vspace{0.7em}

\textbf{Boundary rules:}

\begin{itemize}[
    leftmargin=1.7em,
    itemsep=0.35em,
    topsep=0.3em
]

    \item If an operational step, update mechanism, routing rule, inference
    procedure, decoding rule, or method execution logic is absent, use
    \textbf{Missing Method Procedure}.

    \item If a model component is mentioned but its architecture,
    organization, representation mapping, pooling, normalization, or structural
    design is absent, use \textbf{Missing Model Structure}.

    \item If data construction, filtering, labeling, tokenization,
    normalization, augmentation, segmentation, or input transformation is
    absent, use \textbf{Missing Data Specification}.

    \item If an important configuration choice is introduced but the procedure
    for selecting, tuning, validating, or adapting it is absent, use
    \textbf{Missing Configuration Protocol}.

    \item If metric computation, evaluation split, threshold, evaluation
    sample selection, aggregation, evaluator prompt, or evaluation model
    configuration is absent, use
    \textbf{Missing Evaluation Specification}.

    \item Use an Inconsistency label only when the specification contains
    explicit conflicting claims. A merely absent detail is an Incompleteness
    defect.

\end{itemize}
\vspace{0.7em}

\textbf{Additional ambiguity boundary:}

\begin{itemize}[
    leftmargin=1.7em,
    itemsep=0.35em,
    topsep=0.3em
]

    \item Use \textbf{Ambiguous Definition} when the uncertainty concerns the
    meaning, value, counting convention, mathematical interpretation, scope,
    or formal semantics of a term, symbol, quantity, equation, set, or
    formally defined object.

    \item Use \textbf{Ambiguous Procedure} when the uncertainty concerns what
    an algorithm, component, training stage, inference stage, or processing
    step operationally does.

\end{itemize}

\vspace{0.7em}

\textbf{Exclusions:}

\begin{itemize}[
    leftmargin=1.7em,
    itemsep=0.35em,
    topsep=0.3em
]

    \item Do not flag ordinary choices such as batch size, learning rate,
    epoch count, random seed, hardware, runtime, local paths, package setup,
    or credentials, unless they define a non-standard method-critical
    mechanism.

    \item Do not flag unavailable compute or data, a pure performance
    mismatch, stylistic weakness, or a legitimate open design choice as the
    target defect.

    \item Do not invent the missing value or gold resolution. Diagnose only
    what remains underdetermined in the specification.

\end{itemize}

\vspace{0.7em}

\textbf{Output rules:}

\begin{enumerate}[
    leftmargin=1.7em,
    itemsep=0.35em,
    topsep=0.3em
]

    \item Select exactly one Level~1 category and one Level~2 category.

    \item Do not output multiple defects or alternative labels.

    \item Ignore ordinary implementation choices, standard defaults, minor
    missing details, and speculative concerns.

    \item Prefer the most specific target defect directly supported by the
    specification.

    \item Do not repair the idea or propose a solution.

    \item The description must identify the concrete missing, ambiguous, or
    conflicting implementation detail. Repeating only the taxonomy label is
    insufficient.

\end{enumerate}

\vspace{0.7em}

\textbf{Instance-specific input:}

\vspace{0.4em}

\begin{tcolorbox}[
    enhanced,
    breakable,
    colback=white,
    colframe=rqborder,
    boxrule=0.35pt,
    arc=1mm,
    left=5pt,
    right=5pt,
    top=5pt,
    bottom=5pt,
    width=\linewidth
]
\ttfamily\scriptsize
\raggedright

Research idea specification:\newline
\textless specification\textgreater\newline
\{SPECIFICATION\}\newline
\textless/specification\textgreater

\end{tcolorbox}

\vspace{0.7em}

\textbf{Return only valid JSON:}

\vspace{0.4em}

\begin{tcolorbox}[
    enhanced,
    breakable,
    colback=white,
    colframe=rqborder,
    boxrule=0.35pt,
    arc=1mm,
    left=5pt,
    right=5pt,
    top=5pt,
    bottom=5pt,
    width=\linewidth
]
\ttfamily\scriptsize
\raggedright

\{\newline
\hspace*{1em}"description": "one concrete atomic target-defect diagnosis",\newline
\hspace*{1em}"level1": "Ambiguity|Incompleteness|Inconsistency",\newline
\hspace*{1em}"level2": "Ambiguous Definition|Ambiguous Procedure|\newline
\hspace*{2em}Missing Method Procedure|Missing Model Structure|\newline
\hspace*{2em}Missing Data Specification|Missing Configuration Protocol|\newline
\hspace*{2em}Missing Evaluation Specification|\newline
\hspace*{2em}Conflicting Objective|Conflicting Model Design|\newline
\hspace*{2em}Conflicting Formal Definition"\newline
\}

\end{tcolorbox}

\end{tcolorbox}

\begingroup
\captionsetup{hypcap=false}
\captionof{figure}{
Zero-shot prompt used for Task~2 Defect Localization. Each model receives one
underspecified research idea specification and returns a single atomic
target-defect diagnosis, together with one Level~1 category and one Level~2
category from the \ourbench{} taxonomy. The prompt instructs models to recover
the annotated target defect rather than enumerate all plausible specification
gaps or propose a resolution.
}
\label{fig:task2_prompt}
\endgroup

\clearpage

\begin{tcolorbox}[
    enhanced,
    breakable,
    colback=UnsureLightOrange,
    colframe=UnsureOrange,
    boxrule=0.6pt,
    arc=1pt,
    left=8pt,
    right=8pt,
    top=8pt,
    bottom=8pt,
    width=\linewidth,
    before skip=1em,
    after skip=0.5em
]
\footnotesize

\textbf{Task 2 LLM-as-a-Judge Evaluation Prompt}

\vspace{0.7em}

\textbf{System}

\vspace{0.3em}

You are a strict target-defect identity judge for a scientific-method
benchmark.

\vspace{0.5em}

Do NOT judge whether the predicted defect is valid, important, or present
somewhere in the specification. A specification may contain multiple genuine
defects. Your only task is to decide whether the prediction identifies the same
annotated implementation decision or specification slot as the gold target.

\vspace{0.5em}

Compare the affected component, the concrete implementation slot, and the
missing, ambiguous, or conflicting property at that slot. Shared terminology,
a shared broad module, or a shared downstream consequence is not sufficient.

\vspace{0.5em}

A prediction may still match when its taxonomy or defect-type characterization
is wrong, provided that it uniquely localizes the same target slot.

\vspace{0.5em}

Return strict JSON only.

\vspace{0.8em}

\textbf{User}

\vspace{0.3em}

Determine whether the prediction identifies the same concrete target defect as
the annotation.

\vspace{0.7em}

\textbf{IMPORTANT}

\vspace{0.3em}

The specification may contain multiple genuine defects. A prediction that finds
a valid but different defect must receive \texttt{match=false}.

\vspace{0.7em}

First normalize both defects into:

\begin{enumerate}[
    leftmargin=1.7em,
    itemsep=0.25em,
    topsep=0.3em
]
    \item affected component or entity;

    \item concrete implementation decision, rule, variable, or specification
    slot;

    \item property that is missing, ambiguous, or inconsistent.
\end{enumerate}

\vspace{0.7em}

\textbf{Apply these checks:}

\vspace{0.6em}

\textbf{DIRECT-FIX TEST}

\vspace{0.3em}

Would directly fixing the predicted issue also fix the annotated issue?

\vspace{0.6em}

\textbf{INDEPENDENT-COEXISTENCE TEST}

\vspace{0.3em}

Could both issues exist independently in the same specification? If yes, they
are normally different defects.

\vspace{0.6em}

\textbf{TARGET-SLOT TEST}

\vspace{0.3em}

Do both descriptions concern the same concrete rule, threshold, variable,
operation, module boundary, loss term, transformation, or stopping condition?

\vspace{0.7em}

\textbf{Decision rules:}

\begin{itemize}[
    leftmargin=1.7em,
    itemsep=0.25em,
    topsep=0.3em
]
    \item Paraphrases and different specificity may match.

    \item The prediction does not need to recover the correct hidden
    resolution.

    \item Wrong taxonomy may still match if the same target slot is localized.

    \item Same topic, component, algorithm, or downstream effect is not enough.

    \item A neighboring issue in the same component is \texttt{match=false}.

    \item A different valid defect is \texttt{match=false}.

    \item A generic statement that could refer to multiple defects is
    \texttt{match=false}.

    \item A repair without identifying the target slot is \texttt{match=false}.
\end{itemize}

\vspace{0.7em}

\textbf{Relationship labels:}

\begin{itemize}[
    leftmargin=1.7em,
    itemsep=0.25em,
    topsep=0.3em
]
    \item \texttt{exact\_same\_target}: same slot and substantially correct
    characterization.

    \item \texttt{same\_target\_wrong\_characterization}: same slot, wrong
    defect characterization.

    \item \texttt{broader\_but\_uniquely\_identifies\_target}: broader wording,
    but uniquely same slot.

    \item \texttt{neighboring\_defect}: related component, different
    implementation decision.

    \item \texttt{different\_target}: different component, rule, variable, or
    decision.

    \item \texttt{too\_vague\_to\_localize}: no unique concrete target slot.

    \item \texttt{no\_defect\_identified}: no actual defect is stated.
\end{itemize}

\vspace{0.7em}

Set \texttt{match=true} only for:

\begin{itemize}[
    leftmargin=1.7em,
    itemsep=0.25em,
    topsep=0.3em
]
    \item \texttt{exact\_same\_target}

    \item \texttt{same\_target\_wrong\_characterization}

    \item \texttt{broader\_but\_uniquely\_identifies\_target}
\end{itemize}

\vspace{0.7em}

\textbf{Boundary examples:}

\vspace{0.5em}

\textbf{Example 1}

\vspace{0.3em}

Gold: The tree-search stopping condition is inconsistent.

Prediction: The maximum tree depth is not specified.

Result: \texttt{neighboring\_defect}, \texttt{match=false}. Same search module,
different decision.

\vspace{0.5em}

\textbf{Example 2}

\vspace{0.3em}

Gold: Paper and code disagree on the event count defining a leaf.

Prediction: The leaf-node event threshold is underspecified.

Result: \texttt{same\_target\_wrong\_characterization}, \texttt{match=true}.
Same target slot, wrong defect type.

\vspace{0.5em}

\textbf{Example 3}

\vspace{0.3em}

Gold: The weight combining two loss terms is unspecified.

Prediction: The optimizer learning rate is unspecified.

Result: \texttt{different\_target}, \texttt{match=false}. Both affect training
but are independent.

\vspace{0.7em}

\textbf{Instance-specific input:}

\vspace{0.4em}

\begin{tcolorbox}[
    enhanced,
    breakable,
    colback=white,
    colframe=rqborder,
    boxrule=0.35pt,
    arc=1mm,
    left=5pt,
    right=5pt,
    top=5pt,
    bottom=5pt,
    width=\linewidth
]
\ttfamily\scriptsize
\raggedright

\textless gold\_target\textgreater\newline
Annotated defect:\newline
\{GOLD\_DESCRIPTION\}\newline

\vspace{0.8em}

Relevant source wording:\newline
\{SURFACE\_FORM\}\newline

\vspace{0.8em}

Canonical blocking specification / target slot:\newline
\{BLOCKING\_TEXT\}\newline

\vspace{0.8em}

Optional structured target signature:\newline
\{TARGET\_SIGNATURE\}\newline

\vspace{0.8em}

Resolution reference:\newline
\{RESOLUTION\_REFERENCE\}\newline
\textless/gold\_target\textgreater

\vspace{0.8em}

The resolution reference is only evidence for identifying the intended target
slot. The prediction does not need to recover that resolution. Do not establish
identity from shared consequences.

\vspace{0.8em}

\textless specification\_context\textgreater\newline
\{LOCAL\_CONTEXT\}\newline
\textless/specification\_context\textgreater

\vspace{0.8em}

Use the context only to resolve terminology. Do not validate an alternative
defect from the context and count it as a match.

\vspace{0.8em}

\textless prediction\textgreater\newline
\{PREDICTED\_DESCRIPTION\}\newline
\textless/prediction\textgreater

\end{tcolorbox}

\vspace{0.7em}

\textbf{Return one JSON object with:}

\begin{itemize}[
    leftmargin=1.7em,
    itemsep=0.25em,
    topsep=0.3em
]
    \item \texttt{gold\_target\_slot}: concise normalized gold slot;

    \item \texttt{predicted\_target\_slot}: concise normalized predicted slot;

    \item \texttt{relationship}: one allowed relationship label;

    \item \texttt{match}: boolean consistent with the relationship;

    \item \texttt{confidence}: high, medium, or low;

    \item \texttt{reason}: briefly compare the two concrete implementation
    decisions.
\end{itemize}

\vspace{0.5em}

The reason must state what implementation decision each defect refers to. Do
not merely say that they are similar or different.

\vspace{0.7em}

\textbf{Return only valid JSON:}

\vspace{0.4em}

\begin{tcolorbox}[
    enhanced,
    breakable,
    colback=white,
    colframe=rqborder,
    boxrule=0.35pt,
    arc=1mm,
    left=5pt,
    right=5pt,
    top=5pt,
    bottom=5pt,
    width=\linewidth
]
\ttfamily\scriptsize
\raggedright

\{\newline
\hspace*{1em}"gold\_target\_slot": "concise normalized gold slot",\newline
\hspace*{1em}"predicted\_target\_slot": "concise normalized predicted slot",\newline
\hspace*{1em}"relationship": one of [\newline
\hspace*{2em}"exact\_same\_target",\newline
\hspace*{2em}"same\_target\_wrong\_characterization",\newline
\hspace*{2em}"broader\_but\_uniquely\_identifies\_target",\newline
\hspace*{2em}"neighboring\_defect",\newline
\hspace*{2em}"different\_target",\newline
\hspace*{2em}"too\_vague\_to\_localize",\newline
\hspace*{2em}"no\_defect\_identified"\newline
\hspace*{1em}],\newline
\hspace*{1em}"match": true,\newline
\hspace*{1em}"confidence": "high|medium|low",\newline
\hspace*{1em}"reason": "brief comparison"\newline
\}

\end{tcolorbox}

\end{tcolorbox}

\begingroup
\captionsetup{hypcap=false}
\captionof{figure}{
LLM-as-a-Judge prompt used to evaluate Task~2 target-level defect localization.
The evaluator compares the model-predicted defect description against the
annotated gold target while hiding taxonomy labels. A prediction is counted as
a localization match only when it identifies the same concrete implementation
decision or specification slot as the gold target. Valid but different defects,
neighboring issues, and generic criticisms are counted as mismatches.
}
\label{fig:task2_judge_prompt}
\endgroup

\clearpage

\begin{tcolorbox}[
    enhanced,
    breakable,
    colback=UnsureLightOrange,
    colframe=UnsureOrange,
    boxrule=0.6pt,
    arc=1pt,
    left=8pt,
    right=8pt,
    top=8pt,
    bottom=8pt,
    width=\linewidth,
    before skip=1em,
    after skip=0.5em
]
\footnotesize

\textbf{Task 3 LLM-as-a-Judge Evaluation Prompt}

\vspace{0.7em}

\textbf{System}

\vspace{0.3em}

You are a strict evaluator of clarification actions for underspecified
scientific-method descriptions. Evaluate whether the proposed action would
obtain the information required to resolve the supplied annotated defect.

\vspace{0.5em}

Do not reward lexical similarity to the reference action. A differently worded
question or a different action type may be fully correct. Do not evaluate
whether the target defect itself is valid; it is an adjudicated benchmark input.
Return strict JSON only.

\vspace{0.8em}

\textbf{User}

\vspace{0.3em}

\textbf{Task: Clarification Action Evaluation}

\vspace{0.5em}

Evaluate the candidate clarification action using the following rubric.

\vspace{0.7em}

\textbf{Counterfactual sufficiency test.}

\vspace{0.3em}

Assume that a cooperative and knowledgeable respondent, or an artifact
inspector, answers only what the action explicitly requests. Determine whether
the resulting information would be sufficient to recover the hidden gold
resolution. A generic request for additional details is not sufficient merely
because a maximally helpful respondent could volunteer the complete answer.
A candidate is also not sufficient if it is not an actual clarification question
or concrete evidence-seeking action, such as a diagnosis, repair instruction,
implementation guess, or vague request to ``investigate'' without stating what
information should be obtained.

\vspace{0.7em}

\textbf{Evaluation dimensions:}

\begin{enumerate}[
    leftmargin=1.7em,
    itemsep=0.45em,
    topsep=0.3em
]

    \item \textbf{Target relevance.}

    The action specifically concerns the annotated target defect rather than
    another plausible issue in the specification.

    \item \textbf{Resolution sufficiency.}

    The candidate is a real clarification question or concrete evidence-seeking
    action, and a direct answer to the question, or the result of the proposed
    inspection, would determine every implementation-critical part of the hidden
    resolution. The candidate need not match the wording or action type of the
    reference action.

    \item \textbf{Unsupported assumption.}

    The candidate asserts, recommends, or presupposes an unsupported value for
    the missing detail. Merely asking whether one of several explicit
    alternatives applies does not constitute an unsupported assumption.

\end{enumerate}

\vspace{0.5em}

\textbf{Additional rules:}

\begin{itemize}[
    leftmargin=1.7em,
    itemsep=0.25em,
    topsep=0.3em
]
    \item The reference action is one valid realization, not the only valid
    wording.

    \item Disagreement with the reference action type is not itself an error.

    \item Fluency and verbosity are not evaluation criteria.

    \item Do not use the hidden resolution to excuse an unsupported assertion
    in the candidate, because the evaluated model did not observe that
    resolution.
\end{itemize}

\vspace{0.7em}

\textbf{Instance-specific input:}

\vspace{0.4em}

\begin{tcolorbox}[
    enhanced,
    breakable,
    colback=white,
    colframe=rqborder,
    boxrule=0.35pt,
    arc=1mm,
    left=5pt,
    right=5pt,
    top=5pt,
    bottom=5pt,
    width=\linewidth
]
\ttfamily\scriptsize
\raggedright

Underspecified specification:\newline
\textless specification\textgreater\newline
\{SPECIFICATION\}\newline
\textless/specification\textgreater

\vspace{0.8em}

Annotated target defect:\newline
\textless target\_defect\textgreater\newline
\{TARGET\_DEFECT\}\newline
\textless/target\_defect\textgreater

\vspace{0.8em}

Private hidden resolution:\newline
\textless hidden\_resolution\textgreater\newline
\{HIDDEN\_RESOLUTION\}\newline
\textless/hidden\_resolution\textgreater

\vspace{0.8em}

Private reference action:\newline
\textless reference\_action\textgreater\newline
Action type: \{GOLD\_ACTION\_TYPE\}\newline
Action: \{GOLD\_ACTION\}\newline
Evidence to seek: \{GOLD\_EVIDENCE\_TO\_SEEK\}\newline
\textless/reference\_action\textgreater

\vspace{0.8em}

Private solution evidence:\newline
\textless solution\_evidence\textgreater\newline
\{SOLUTION\_EVIDENCE\}\newline
\textless/solution\_evidence\textgreater

\vspace{0.8em}

Candidate clarification action:\newline
\textless candidate\_action\textgreater\newline
Action type: \{PREDICTED\_ACTION\_TYPE\}\newline
Action: \{PREDICTED\_ACTION\}\newline
Expected information: \{PREDICTED\_EXPECTED\_INFORMATION\}\newline
\textless/candidate\_action\textgreater

\end{tcolorbox}

\vspace{0.7em}

\textbf{Return only valid JSON:}

\vspace{0.4em}

\begin{tcolorbox}[
    enhanced,
    breakable,
    colback=white,
    colframe=rqborder,
    boxrule=0.35pt,
    arc=1mm,
    left=5pt,
    right=5pt,
    top=5pt,
    bottom=5pt,
    width=\linewidth
]
\ttfamily\scriptsize
\raggedright

\{\newline
\hspace*{1em}"target\_relevant": true,\newline
\hspace*{1em}"resolution\_sufficient": true,\newline
\hspace*{1em}"unsupported\_assumption": false,\newline
\hspace*{1em}"confidence": "high|medium|low",\newline
\hspace*{1em}"reason": "brief evidence-based explanation"\newline
\}

\end{tcolorbox}

\end{tcolorbox}

\begingroup
\captionsetup{hypcap=false}
\captionof{figure}{
LLM-as-a-Judge prompt used to evaluate Task~3 clarification actions.
The evaluator assesses target relevance, resolution sufficiency, and
unsupported assumptions using the private gold resolution and supporting
evidence. Resolution sufficiency requires the output to be a concrete
clarification question or evidence-seeking action whose answer or inspection
result would determine the missing implementation-critical detail.
Per-instance Clarification Action Success is deterministically derived from
these three judgments.
}
\label{fig:task3_judge_prompt}
\endgroup

\clearpage

\end{document}